\documentclass[10pt,twocolumn,letterpaper]{article}

\usepackage[pagenumbers]{cvpr} %

\usepackage{graphicx}
\usepackage{amsmath}
\usepackage{amssymb}
\usepackage{booktabs}
\usepackage{bbding}
\usepackage[ruled,linesnumbered]{algorithm2e}
\usepackage{subcaption}
\usepackage{multirow}

\usepackage[pagebackref,breaklinks,colorlinks]{hyperref}

\usepackage[capitalize]{cleveref}
\crefname{section}{Sec.}{Secs.}
\Crefname{section}{Section}{Sections}
\Crefname{table}{Table}{Tables}
\crefname{table}{Tab.}{Tabs.}

\begin{document}

\title{Learn to See Faster:
Pushing the Limits of High-Speed Camera with Deep Underexposed Image Denoising}

\author{Weihao Zhuang  \qquad Tristan Hascoet  
\qquad Ryoichi Takashima \qquad Tetsuya Takiguchi \\ \\Kobe University, Japan}

\maketitle

\begin{abstract}
The ability to record high-fidelity videos at high acquisition rates is central to the study of fast moving phenomena.
The difficulty of imaging fast moving scenes lies in a trade-off between motion blur and underexposure noise:
On the one hand, recordings with long exposure times suffer from motion blur effects caused by movements in the recorded scene.
On the other hand, the amount of light reaching camera photosensors decreases with exposure times so that short-exposure recordings suffer from underexposure noise.
In this paper, we propose to address this trade-off by treating the problem of high-speed imaging as an underexposed image denoising problem.
We combine recent advances on underexposed image denoising using deep learning and adapt these methods to the specificity of the high-speed imaging problem.
Leveraging large external datasets with a sensor-specific noise model, our method is able to speedup the acquisition rate of a High-Speed Camera over one order of magnitude while maintaining similar image quality.
\end{abstract}

\section{Introduction}
\label{sec_intro}

\begin{figure*}[t]
\centering
\begin{subfigure}[t]{0.75\textwidth}
\centering
    \includegraphics[width=0.24\textwidth]{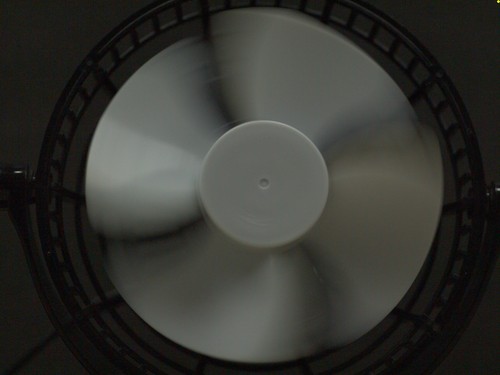}
    \includegraphics[width=0.24\textwidth]{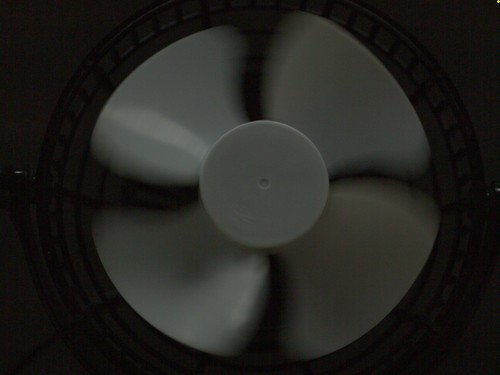}
    \includegraphics[width=0.24\textwidth]{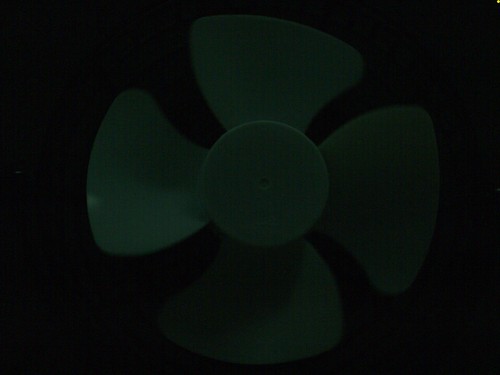}
    \includegraphics[width=0.24\textwidth]{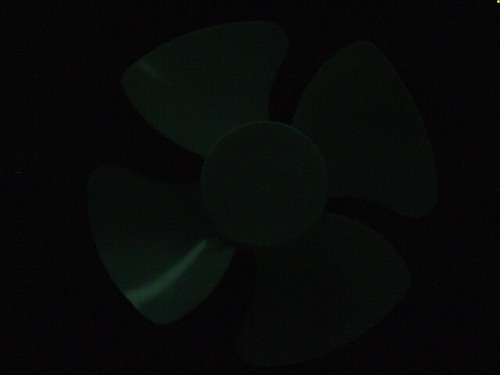}
\end{subfigure}
\\[2ex]

\begin{subfigure}[t]{0.75\textwidth}
\centering
    \includegraphics[width=0.24\textwidth]{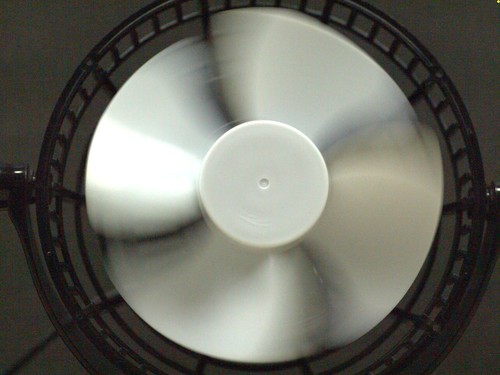}
    \includegraphics[width=0.24\textwidth]{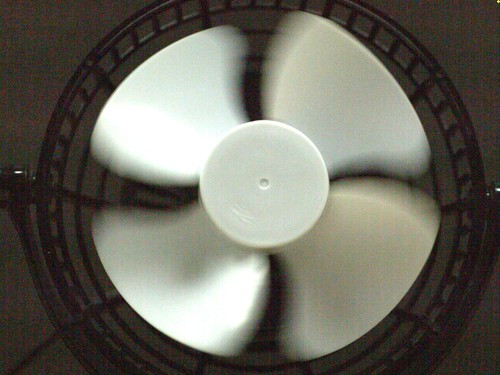}
    \includegraphics[width=0.24\textwidth]{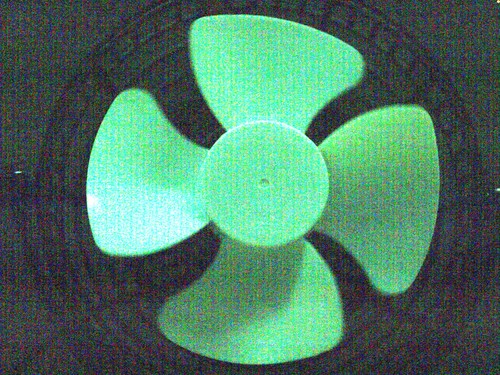}
    \includegraphics[width=0.24\textwidth]{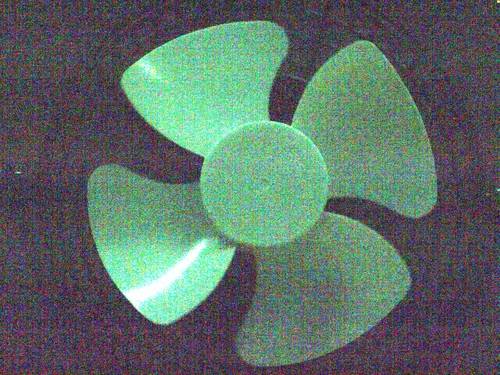}
\end{subfigure}
\\[2ex]

\begin{subfigure}[t]{0.168\textwidth}
\centering
    \includegraphics[width=1\textwidth]{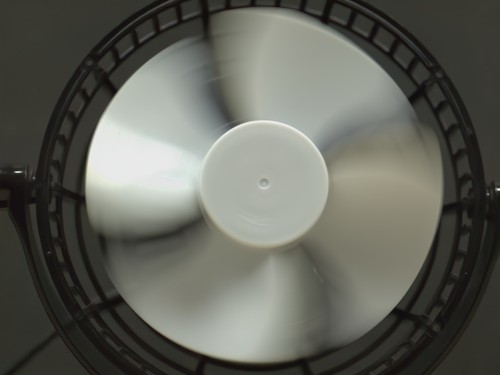}
$1/_{500} (s)$
\end{subfigure}
\begin{subfigure}[t]{0.18\textwidth}
\centering
\includegraphics[width=1\textwidth]{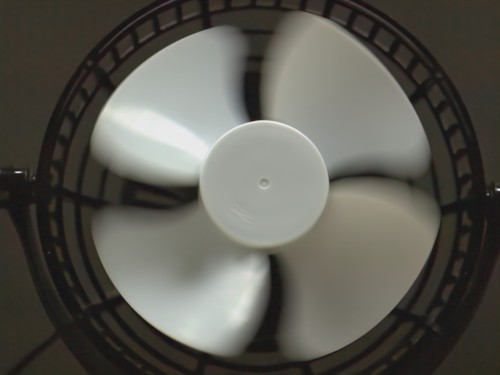}
$1/_{1k} (s)$
\end{subfigure}
\begin{subfigure}[t]{0.18\textwidth}
\centering
\includegraphics[width=1\textwidth]{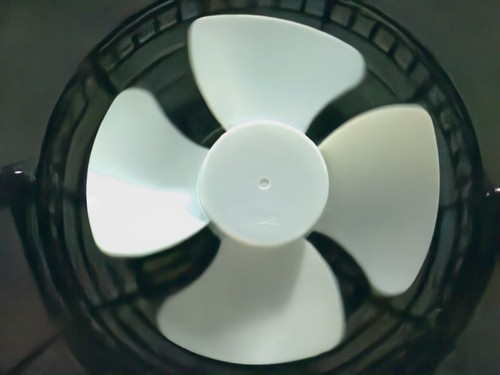}
$1/_{5k} (s)$
\end{subfigure}
\begin{subfigure}[t]{0.18\textwidth}
\centering
\includegraphics[width=1\textwidth]{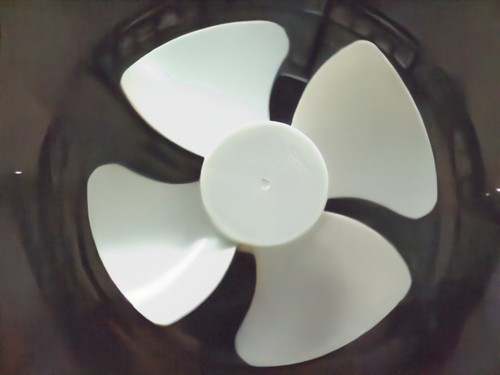}
$1/_{10k} (s)$
\end{subfigure}

\caption{Illustration of highly dynamic scenes recorded with high-speed camera at different shutter speed.
The three rows of images from top to bottom are: JPG images processed by the HSC proprietary ISP, 
JPG visualized with gain, and the ouput of our model.
Videos recorded with low shutter speed suffer motion blur, while videos recorded at high shutter speed suffer from underexposition noise.
Using carefully adapted underexposed image denoising, 
our method allows to push the shutter-speed limit at which we are able to record high-fidelity videos.
}
\label{fig_video}
\end{figure*}
How fast can we see? The human visual system is estimated to perceive the world at a rate of 30 to 60 Frames Per Second (FPS). 
Any movement or oscillation faster than this frequency is imperceptible to our naked eye. 
Augmented with cameras and computers, we are now able to record images of the world beyond hundreds of thousands of frames per second, allowing us to observe phenomena that cannot be captured by the naked eye. 

The ability to make near instantaneous observations of the world is key to many scientific research endeavours. 
For example, rocket combustion experiments \cite{marsh2021time, winter2017high, richecoeur2008experimental} in the aerospace research rely on high-speed imaging devices for data collection, as a minimum of 100,000 FPS is required to analyze the extremely fast movements resulting from combustion. 
Image cytometry in biology \cite{han2016imaging, mikami2018high} also relies on high-speed cameras to record the characteristics of each cell among tens of thousands of cells, making it convenient for researchers to filter out the cells useful for research.
Fluid dynamics \cite{song2019experimental, thoroddsen2008high}, structure mechanics \cite{pan2016full, liu2022research, wirth2018analysis}, and many other fields \cite{ficek2019influence, settles2006high, stevens2011rollover, wang2011high, liu2009novel, balsalobre2014concurrent, witte2008biomechanical} similarly rely on the ability to perceive the world at high frame rates. Beyond the sciences, industrial processes often rely on high-speed imaging for both quality control on production lines and development purposes. In the past decades, high-speed camera manufacturers have been pushing the limits of high-speed imaging, enabling the study of ever faster and higher frequency phenomenon.

Recording fast moving scenes is challenging due to a trade-off between underexposure noise and motion blur, as illustrated in Figure \ref{fig_video}. 
Dynamic scenes recorded with long exposure times suffer from motion blur caused by fast moving points reflecting light from multiple locations along their trajectory during each frame's recording.
Short exposure times alleviate motion blur by limiting the movement range occurring during each frame's recording.
However, as the camera sensor exposure time decreases, the amount of light reaching the sensor also decreases so that the strength of the signal recorded by the sensor decreases, while the sensor noise remains comparatively constant.
In other words, the Noise-to-Signal Ratio (SNR) of the recorded signal decreases with shutter speed, which results in problematic underexposure noise for very 
short exposure times. 
One solution to reduce the underexposure noise of short exposure recordings consists in increasing the amount of light in the scene using additional light sources. 
However, this solution requires a controlled recording environment, which is not always possible.

In this paper, we are interested in maximizing the acquisition rate at which we can get clear recordings from an industrial grade High-Speed Camera in controlled lighting conditions.
We propose to do so by treating the problem of high-speed imaging as an underexposed image denoising problem.

In recent years, several successful deep learning approaches to underexposed image denoising have been proposed \cite{chen2018learning, chen2019seeing, wei2020physics, lamba2021restoring, dong2022abandoning, zhang2021rethinking}.
These works formulate the problem as a supervised learning task in which a CNN is trained to regress clear long exposure ground-truth images from noisy short-exposure raw images, and have been applied to night scene recordings
with impressive image quality improvements over traditional Image Signal Processing (ISP) codecs.
Unfortunately, the most extreme low-light datasets publicly available (e.g; SID \cite{chen2018learning} or MCR \cite{dong2022abandoning}) show noise levels corresponding to video streams of only a few thousands FPS recorded with our HSC (High-Speed Camera) device in a standard lightning environment. 
For higher shutter speed, the denoising problem starts to differ for two reasons: First, the SNR becomes much lower, which makes the problem harder. Second, the noise stems from sensor-specific reactions to very short exposure times, which lends it characteristic patterns. 
Directly applying existing models to our data thus does not allow for clear denoising of high shutter speed videos.

One straightforward approach to build on these previous works and address this limitation would be to collect a large dataset of HSC images so as to train a denoising model tailored to our target noise distribution.
In practice, we found this approach to come with important data collection challenges:
Industrial-grade HSC are not portable devices: they are heavy, expensive and fragile equipments that require access to a strong power supply system. 
Collecting a diverse enough dataset to train a generic scene denoising model is thus significantly harder to do with a HSC than with portable devices like the ones used to collect existing datasets.
Furthermore, our HSC images display characteristic noise patterns that were not observed in existing datasets, and that existing models do not seem to accurately denoise. This suggests that sensor-specific noise patterns may be needed at training for efficient denoising
of very high shutter speed images.
Collecting a dedicated dataset of underexposed images for each type of HSC sensor 
would be a great practical hindrance. 
Instead, a general method that would allow to efficiently denoise HSC videos with minimum data collection efforts would be preferable.

Towards that goal, we draw inspiration from recent works on underexposure noise synthesis and propose a procedure to generate high shutter speed underexposed raw frames from large available datasets.
Our procedure combines high-bit raw reconstruction from JPG  images with a noise synthesis model using bias frames collected from the target HSC device. 
This allows us to leverage large and publicly available datasets to generate a training dataset of diverse scenes showcasing the specific noise distribution of the target HSC recording conditions.
We then leverage this synthetic dataset to train the latest and best performing underexposure denoising models.

To evaluate our method, we have collected a dataset of static and dynamic scences. 
Static scenes were used to quantitatively evaluate the accuracy of our denoising results using noisy frames aligned with their  ground-truth.
Dynamical scenes were used to qualitatively illustrate improvements in imaging fast moving phenomena.
To summarize, the contributions of our work are as follows:
\begin{itemize}
    \item We optimize the trade-off between underexposure noise and motion blur in high-speed imaging:
our method enables high-speed cameras to record clear images beyond one order of magnitude faster shutter speed, 
without relying on artificial light sources
\item We analyze the nature of the underexposure noise stemming from high shutter speed acquisition and propose a noise synthesis procedure that generates underexposed images with similar noise patterns.
\item Our method takes into account the practical difficulty of data collection with non-portable High-Speed Cameras: 
It only requires the collection of bias frames from the target device and leverages large public datasets to synthesize a diverse set of training data.

\end{itemize}

\section{Related works}
\paragraph{Underexposure Image Denoising.}
Many methods have been proposed for solving the well established problem of underexposure image denoising and enhancement.
Gamma correction and histogram equalization methods are  two classical methods for underexposure image enhancement.
Pre-deep learning approaches have proposed different methods relying on hand-crafted filters and feature representations,
including ealry works on adaptive histogram equalization (AHE) \cite{pizer1987adaptive}, the multi-scale Retinex model \cite{jobson1997multiscale}, based on the original Retinex model \cite{land1977retinex}, or wavelet transforms \cite{loza2013automatic}.
The NLM \cite{buades2008nonlocal} and BM3D \cite{dabov2007image} have shown powerful denoising abilities, but often require image prior Gaussian noise parameter.

In recent years, a number of works have proposed to address the underexposed image denoising problem with deep learning \cite{chen2018learning,chen2019seeing,wei2020physics,dong2022abandoning,lamba2021restoring,zhang2021rethinking}
.
In \cite{chen2018learning}, a UNet was first proposed
 to perform end-to-end denoising and processing of images from the camera sensor's raw output.
The rationale \cite{chen2018learning} for using RAW unprocessed image from camera sensors as input to their  model is to avoid the loss of information incurred by the traditional image processing pipeline so as to maximize the use of the optical information captured by the sensors. 

Despite impressive quantitative improvements over classical approaches one drawback of the deep learning methods \cite{chen2019seeing,dong2022abandoning,lamba2021restoring,chen2018learning} is that they require a large number of noisy/clear image pairs to train the denoising model.
Pairs of images need to be collected as a set of pixel-aligned long-exposure-short-exposure image pairs, and the collection effort is time-consuming.
This time-consuming image collection also contributes to the scarcity of datasets for studying underexposed RAW image denoising.
Current common datasets for underexposed RAW image denoising include SID \cite{chen2018learning}, DRV \cite{chen2019seeing},Canon \cite{Charles2013}, ELD \cite{wei2020physics} and MCR \cite{dong2022abandoning}.

These works mainly address the problem of night scene imaging, in which the low light exposure of the camera sensor is caused by weak illumination of the scene, rather than the very short exposure times we are considering in the setting of high-speed imaging.

\paragraph{Noise Synthesis.}
Noise synthesis models aim to generate noisy images from clear ground-truth in order to train denoising models without the need for expensive and time-consuming data collection.
Noise synthesis can be divided into three categories: DNN-based, physics-based, and the recently proposed real-noise-based.
DNN-based \cite{chang2020learning,chen2018image,kim2019grdn} leverage conditional generatve models, typically GANs, to synthesize noise image,
while physics-based models \cite{brooks2019unprocessing,wei2020physics,wang2019enhancing,foi2008practical} implement statistical models  of the process of noise formation from the photon hitting the optical sensor to the digital signal.

The real-noise-based approach was recently proposed by RNS \cite{zhang2021rethinking}.
RNS collects real bias frame as a database and synthesizes noise images by randomly sampling from the database.
Physics-based and real-noise-based approach will bring the trade-off between generalization and specialization.
Physics-based method can model most cases, but is not generalized to every camera.
Real-noise-based, on the other hand, can be specialized to a particular camera, but requires bias frames collected with the sensor.
The specialization of the real-noise-based approach is particularly suitable for our case, where we need to synthesize noisy images specifically for a particular camera.

We combine the UPI \cite{brooks2019unprocessing} and RNS \cite{zhang2021rethinking} methods to synthesize RAW training images using the COCO \cite{lin2014microsoft} dataset. Instead of relying on high-speed cameras to collect a large number of scene images.

\section{Method}

\begin{figure*}[t]
\centering
\begin{subfigure}[t]{1\textwidth}
\centering
    \includegraphics[width=1\textwidth]{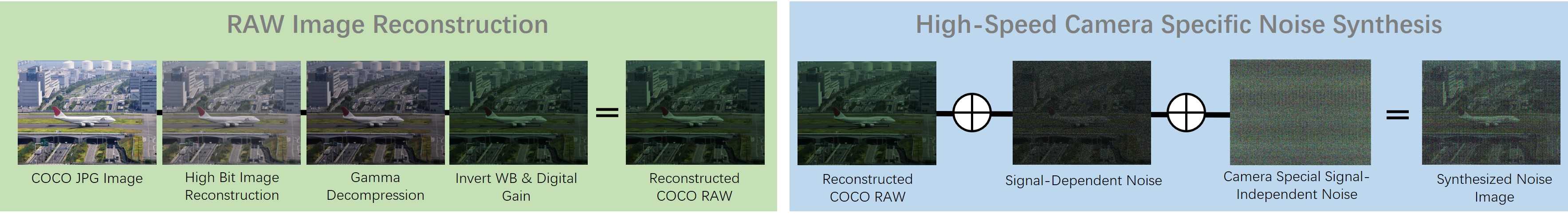}
    \caption{Noise Synthesis Pipeline}
    \label{fig_noise_synthesis}
\end{subfigure}

\begin{subfigure}[t]{1\textwidth}
\centering
    \includegraphics[width=1\textwidth]{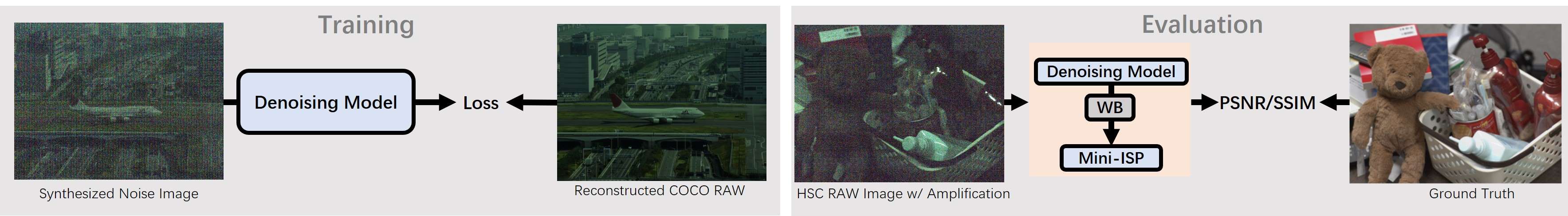}
    \caption{Training and Evaluation Pipeline}
    \label{fig_train_eval_pipeline}
\end{subfigure}%
\caption{Noise synthesis pipeline and denoising pipeline}
\label{fig_synthesis_pipeline}
\end{figure*}

Figure \ref{fig_synthesis_pipeline} illustrates the overall pipeline of our proposed approach. 
To synthesize a training dataset, 
we start by reconstructing long-exposure raw ground-truth frames from the JPG images provided by the COCO dataset.
We then simulate corresponding noisy short-exposure raw frames by scaling down the long-exposure frames' intensities and applying a sensor-specific noise model.
Using this dataset, we train the NAFNet \cite{chen2022simple} model to regress clear long-exposure raw frames from their corresponding noisy short-exposure frames. 
At inference time, we start by denoising short-exposure raw input frames using the trained NAFNet model.
We then perform the remaining image processing steps on the denoised raw frame using a small CNN module we refer to as the Mini-ISP. 
The Mini-ISP was trained to reproduce the processing of the HSC's proprietary ISP on a few training ground-truth frames collected from the camera.

We quantitatively evaluate our denoising results on static scenes collected from our HSC device and qualitatively evaluate our denoising results on highly dynamical scenes.
In its current version, our model is trained and evaluated separately for each exposure times: 
We have synthesized separate training datasets, with corresponding noise levels, 
for each of the exposure times we have collected HSC frames for.
We then train and evaluate one model per exposure time.

In the remainder of this section, 
we describe each of the above steps in more details:
We start by presenting the data we collected using a high-speed camera in Section \ref{sec_dataset}.
In Section \ref{sec_noise_ana}, we analyse the noise present in our collected frames, 
and compare it to currently available underexposed image datasets to situate our work.
We present the noisy frame synthesis procedure we used to generate our synthetic dataset in Section \ref{sec_noise}.
Lastly, we give further details on our training and evaluation procedure in Section \ref{sec_denoise}.

\subsection{Data Collection}
\label{sec_dataset}

We used the HX-7S manufactured by Nac Image Technology Inc. as our target HSC device.
We recorded videos for three kind of scenes from this camera, as described below.
Unless stated otherwise, we collected videos of each scene using five different shutter speeds: $1/_{10k}$, $1/_{5k}$, $1/_{1k}$, $1/_{500}$ and $1/_{100}$ seconds. 
Each video is made of 10 consecutive frames, recorded at a resolution of $2560\times1920$ pixels.

\textbf{Static Scenes}:
For the quantitative evaluation experiment, we collected 40 static scene images using a high-speed camera.
We saved the images in two formats: RAW and JPG.
We use the images with shutter speed $1/_{100}$ second as the correct exposure images ground truth.
That is, we do not feed the denoising model with $1/_{100}$ images and perform quantitative metrics evaluation.
Our collection of static images is strictly pixel-aligned.
The manufacturer provides a software interface 
that can remotely operate the high-speed camera,
so that in each capture we do not need to touch the high-speed camera.
This ensures that the pixels of each frame are strictly aligned in the same scene.

\textbf{Dynamic Scenes}:
In addition, we recorded videos of fast moving scenes to illustrate the improvements in the motion blur and underexposure noise trade-off.
These dynamic scenes were record in a studio with fluorescent lamp light source illumination.
We recorded 4 such videos illustrating the application and difficulty of high-speed imaging, including acoustic and fluid dynamics.
Figure \ref{fig_video} illustrates frames from one such dynamic video, and more illustrations are given in the Appendix.

\textbf{Bias Frames}:
As further detailed in the following Sections, HSC video frames feature a strong signal-independent noise component with characteristic spatial patterns.
In order to model this characteristic noise component in our synthesized images, 
we collected bias frames in the darkroom using the five different shutter speeds described above.
Each shutter speed was collected for 1000 frames, and a total of 5000 images were collected as a database for noise analysis and noise image synthesis.

\subsection{Noise Analysis}
\label{sec_noise_ana}

\begin{figure}[t]
  \centering
  \includegraphics[width=.8\linewidth]{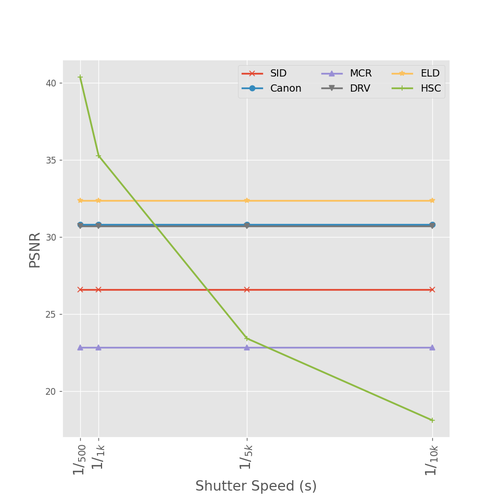}
  \caption{Noise intensity of different dataset. The HSC dataset is noisier than the public datasets (SID \cite{chen2018learning}, DRV \cite{chen2019seeing},Canon \cite{Charles2013}, ELD \cite{wei2020physics}, MCR \cite{dong2022abandoning})at $1/_{10k} (s)$ shutter speed.}
\label{fig_noise}
\end{figure}

In Figure \ref{fig_noise}, we show the evolution of the average PSNR of HSC video frames with shutter speed and contrast it to the average PSNR of publicly available datasets including SID \cite{chen2018learning}, DRV \cite{chen2019seeing}, Canon \cite{Charles2013}, ELD \cite{wei2020physics} and  MCR \cite{dong2022abandoning}).
A sharp decline in PSNR can be seen in the range of a few thousandth shutter speed and, 
at $1/_{10k} (s)$, our video frames show PSNR quite lower than other existing datasets.
As shown in the middle row of Figure \ref{fig_video}, a distinctive streak noise pattern becomes increasingly visible with higher shutter speeds, which was not observed in previous datasets. %
Interestingly, the bias frames collected from the darkroom show similar streak noise patterns, which suggest that the noise in high shutter speed may predominantly stem from a signal independant component.

Accurately isolating and modeling different noise components from observations is a challenging task.
In \cite{zhang2021rethinking}, the authors divide the noise of underexposed images into two components: 
a signal-dependent and a signal-independent noise.
In order to investigate the extent of these two noise components on our HSC recordings and their evolution with shutter speed,
we analyze random variations of raw sensor values using our static scene videos.
Given a video of $T=10$ frames, and a spatial location $i,j$ on the sensor grid, we denote by $X_{t,i,j}$ the raw sensor value returned at location $i,j$ for the frame $t\in[1,T]$.
Denoting by $\bar{X}_{i,j}$ the raw sensor value averaged in time,
we define the noise energy $E_{i,j}$ for this sensor response as:

\begin{equation}
\begin{split}
\bar{X}_{ij} &= \frac{1}{T} \sum_{t=1}^{T} X_{ijt} \\
E_{ij} &= \frac{1}{T} \sum_{t=1}^{T} (X_{ijt}-\bar{X}_{ij})^2
\end{split}
\end{equation}

\begin{figure}[t]
\centering
\begin{subfigure}[t]{0.23\textwidth}
\centering
    \includegraphics[width=1\textwidth]{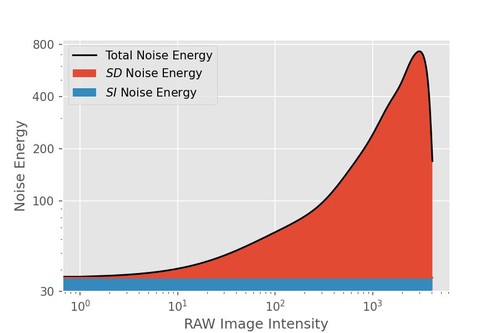}
    \caption{Noise Energy Function}
    \label{fig_noise_energy}
\end{subfigure}
\begin{subfigure}[t]{0.23\textwidth}
\centering
    \includegraphics[width=1\textwidth]{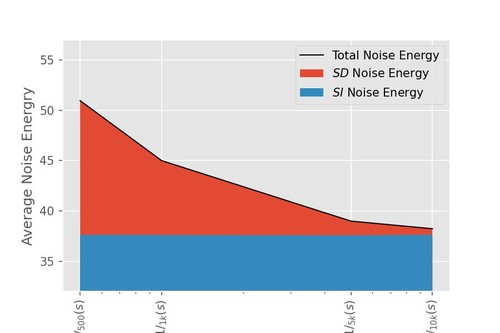}
    \caption{Average Noise Energy}
    \label{fig_average_noise}
\end{subfigure}

\begin{subfigure}[t]{0.23\textwidth}
\centering
    \includegraphics[width=1\textwidth]{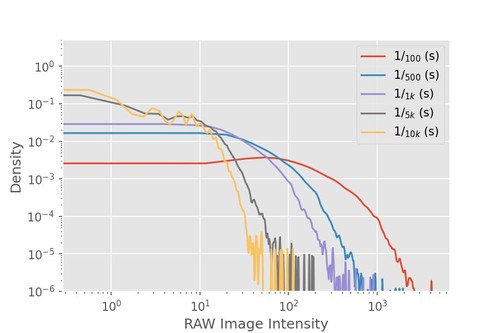}
    \caption{Raw sensor value distributions}
    \label{fig_pixel_intensity}
\end{subfigure}
\begin{subfigure}[t]{0.23\textwidth}
\centering
    \includegraphics[width=1\textwidth]{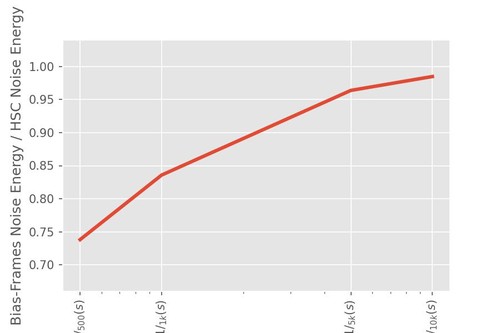}
    \caption{Relative contribution of the $SI$ noise to the total noise energy}
    \label{fig_bias_noise_energy}
\end{subfigure}%
\caption{Figure \ref{fig_noise_energy} shows the change of noise energy of the HSC dataset as the optical information increases ($x$-axis). 
As the signal-dependant noise remains constant, 
the increase in noise energy with raw values 
is attributed to the signal-dependant noise component. 
Figure \ref{fig_average_noise} shows the trend of the average noise energy at different shutter speed. The noise energy of the Bias-Frames collected by the high-speed camera remains stable, while the noise energy of the HSC real scene decreases. Figure \ref{fig_bias_noise_energy} shows that as the shutter speed increases, the noise is dominated by Bias-Fames (signal-independent noise).}
\label{fig_noise_analysis}
\end{figure}

Figure \ref{fig_noise_energy} shows the evolution of the noise energy $E=f(\bar{X})$ as a function of raw sensor values $\bar{X}$ averaged across all static scene video frames of our dataset.
We consider the energy of the signal-independent component of the noise to be a constant function of raw sensor values equal to the noise energy estimated %
from our bias frames
and the signal-dependent component to account for the remaining noise energy. 
This allows us to estimate the noise energy of both components as a function of raw sensor values, as illustrated by the area curves in Figure \ref{fig_average_noise}.

\begin{equation}
\begin{split}
E &= E_{SD} + E_{SI} \\
E &= f_{SD}(\bar{X}) + f_{SI}(\bar{X}) %
\end{split}
\end{equation}

where the $SD$ and $SI$ subscripts refer to the signal-dependant and signal-independent noise components respectively.
We then compute the raw sensor value distribution $p_S(\bar{X})$ for each shutter speed $S$ from the static video frames and derive the noise energy of both noise components for each shutter speed $S$ as:

\begin{equation}
\begin{split}
E_{SD}(s) = \mathbb{E}_{\bar{X}} p_S(\bar{X}) \times f_{SD}(\bar{X})  \\
E_{SI}(s) = \mathbb{E}_{\bar{X}} p_S(\bar{X}) \times f_{SI}(\bar{X})  \\
\end{split}
\end{equation}

\begin{figure*}[t]
\centering
\begin{subfigure}[b]{0.3\textwidth}
\centering
    \includegraphics[width=0.23\linewidth]{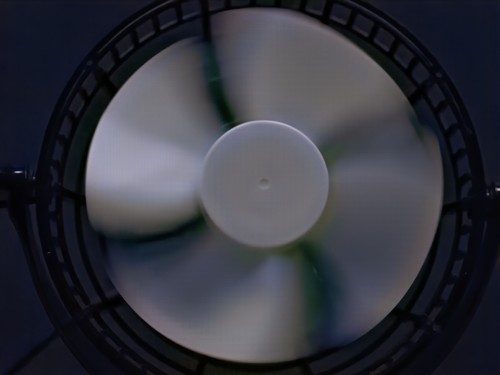}
    \includegraphics[width=0.23\linewidth]{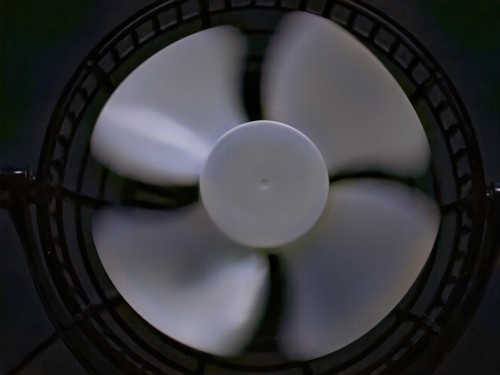}
    \includegraphics[width=0.23\linewidth]{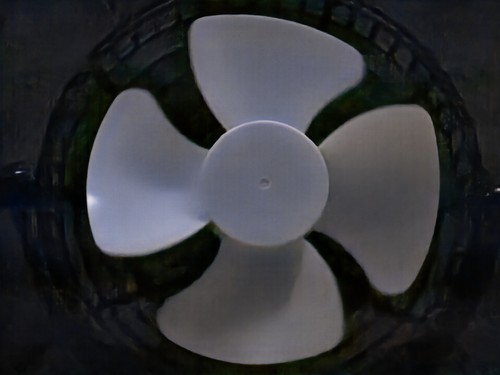}
    \includegraphics[width=0.23\linewidth]{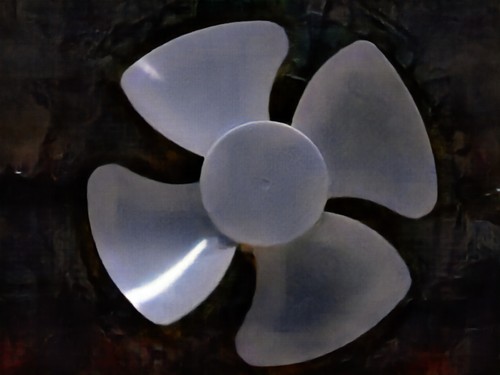}
    \caption{SID \cite{chen2018learning}}
    \label{fig_adapt_sid}
\end{subfigure}
\begin{subfigure}[b]{0.3\textwidth}
\centering
    \includegraphics[width=0.23\linewidth]{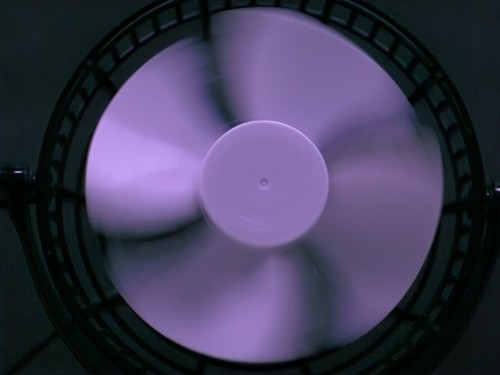}
    \includegraphics[width=0.23\linewidth]{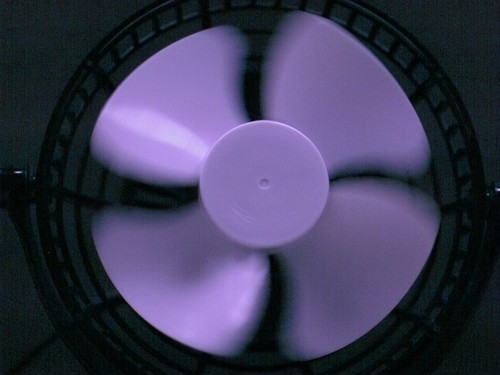}
    \includegraphics[width=0.23\linewidth]{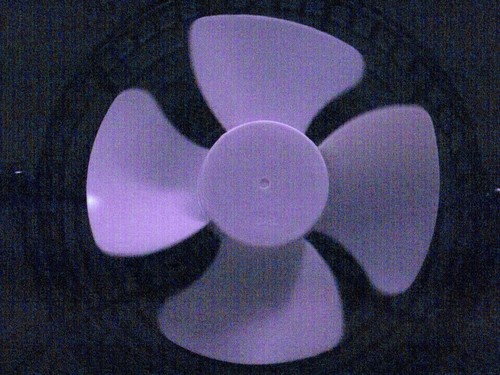}
    \includegraphics[width=0.23\linewidth]{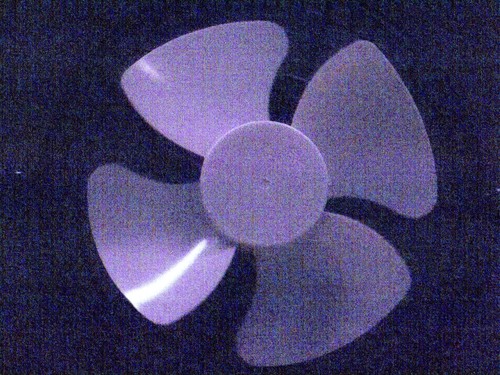}
    \caption{SMID \cite{chen2019seeing}}
    \label{fig_adapt_smid}
\end{subfigure}
\begin{subfigure}[b]{0.3\textwidth}
\centering
    \includegraphics[width=0.23\linewidth]{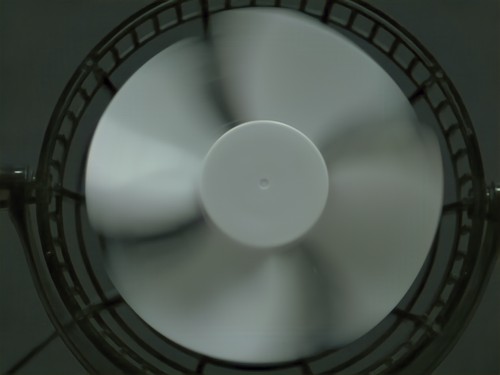}
    \includegraphics[width=0.23\linewidth]{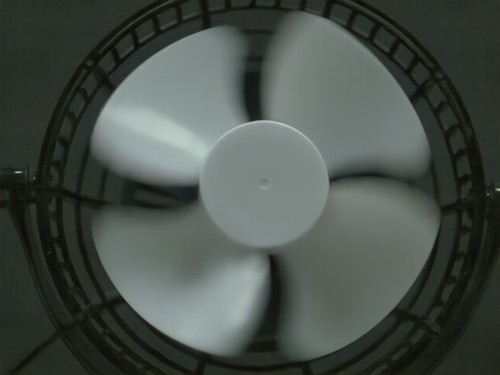}
    \includegraphics[width=0.23\linewidth]{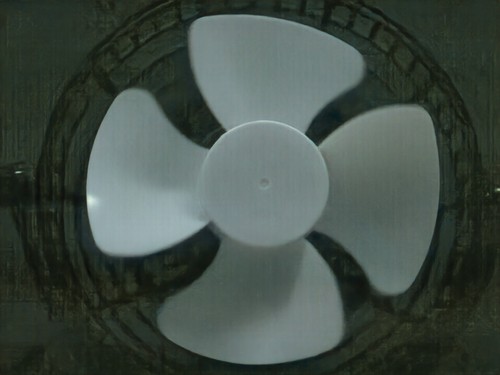}
    \includegraphics[width=0.23\linewidth]{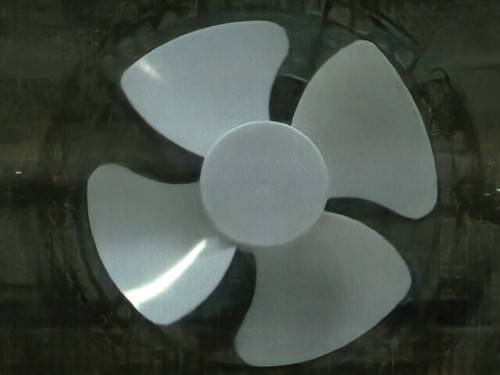}
    \caption{ELD \cite{wei2020physics}}
    \label{fig_adapt_eld}
\end{subfigure}

\begin{subfigure}[b]{0.3\textwidth}
\centering
    \includegraphics[width=0.23\linewidth]{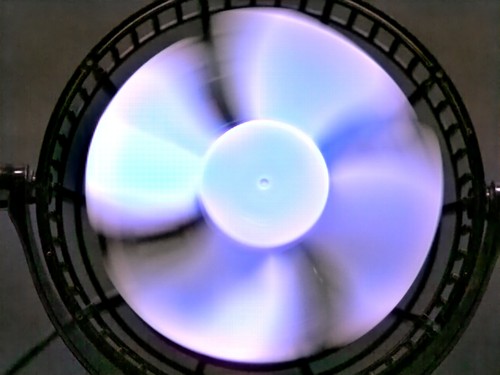}
    \includegraphics[width=0.23\linewidth]{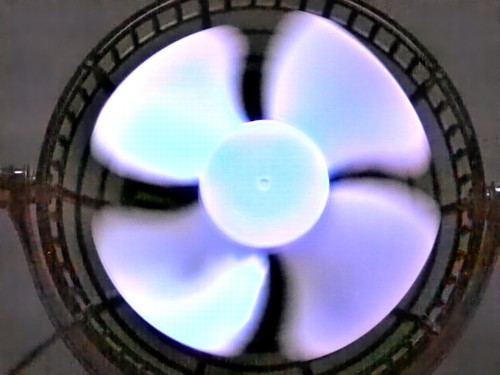}
    \includegraphics[width=0.23\linewidth]{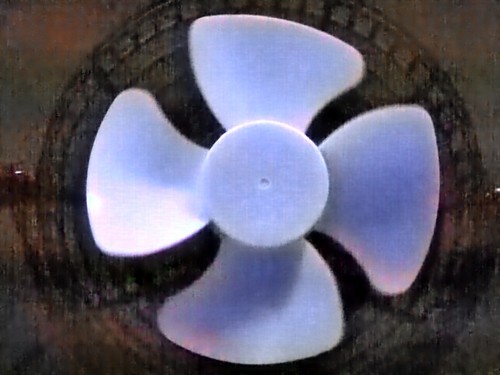}
    \includegraphics[width=0.23\linewidth]{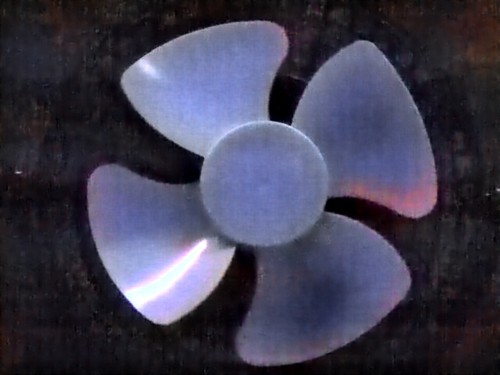}
    \caption{RED \cite{lamba2021restoring}}
    \label{fig_adapt_red}
\end{subfigure}
\begin{subfigure}[b]{0.3\textwidth}
\centering
    \includegraphics[width=0.23\linewidth]{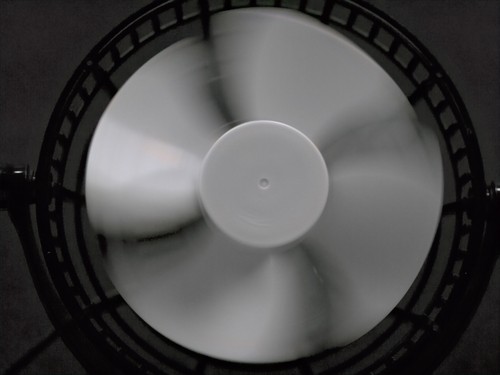}
    \includegraphics[width=0.23\linewidth]{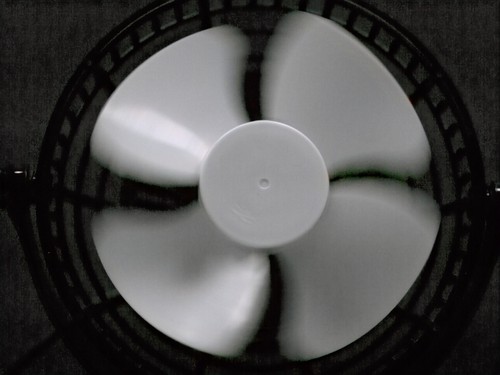}
    \includegraphics[width=0.23\linewidth]{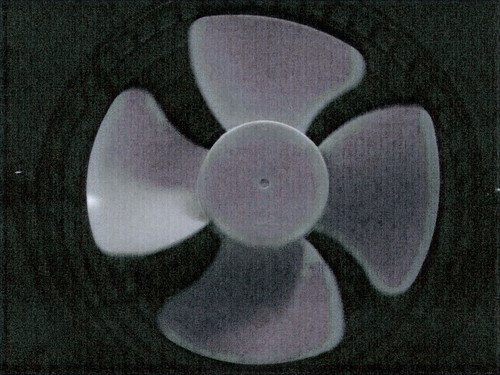}
    \includegraphics[width=0.23\linewidth]{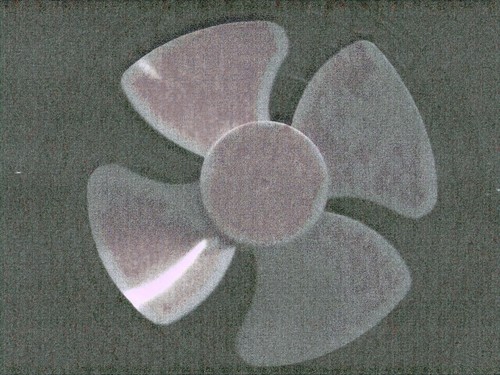}
    \caption{ABFS \cite{dong2022abandoning}}
    \label{fig_adapt_abfsd}
\end{subfigure}
\begin{subfigure}[b]{0.3\textwidth}
\flushright
    \includegraphics[width=0.23\linewidth]{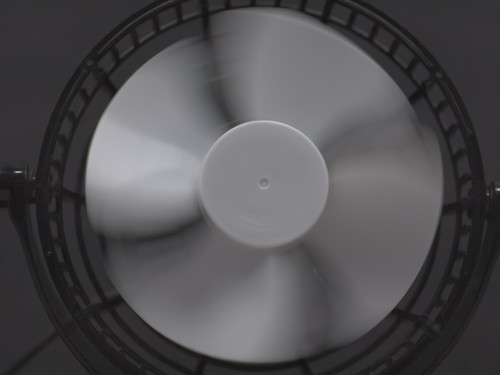}
    \includegraphics[width=0.23\linewidth]{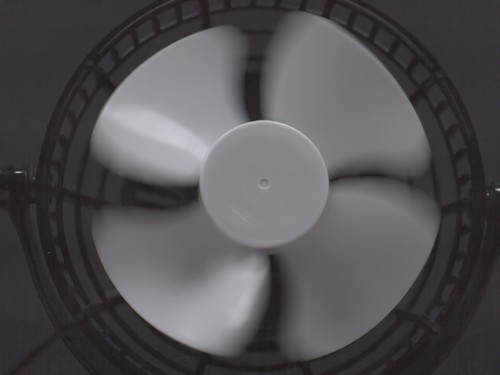}
    \includegraphics[width=0.23\linewidth]{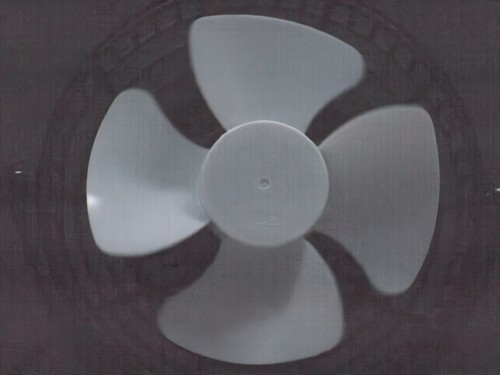}
    \includegraphics[width=0.23\linewidth]{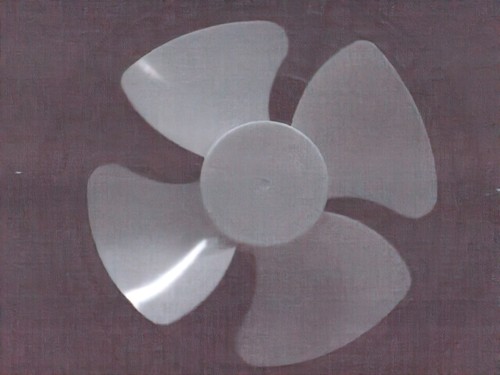}
    \caption{RNS \cite{zhang2021rethinking}}
    \label{fig_adapt_rns}
\end{subfigure}

\begin{subfigure}[b]{0.3\textwidth}
\centering
    \includegraphics[width=0.23\linewidth]{fig/hscVideo_jpggain_10_500.jpg}
    \includegraphics[width=0.23\linewidth]{fig/hscVideo_jpggain_10_1000.jpg}
    \includegraphics[width=0.23\linewidth]{fig/hscVideo_jpggain_10_5000.jpg}
    \includegraphics[width=0.23\linewidth]{fig/hscVideo_jpggain_10_10000.jpg}
    \caption{Camera JPG}
    \label{fig_adapt_jpg}
\end{subfigure}
\begin{subfigure}[b]{0.3\textwidth}
\centering
    \includegraphics[width=0.23\linewidth]{fig/hscVideo_ours_10_500.jpg}
    \includegraphics[width=0.23\linewidth]{fig/hscVideo_ours_10_1000.jpg}
    \includegraphics[width=0.23\linewidth]{fig/hscVideo_ours_10_5000.jpg}
    \includegraphics[width=0.23\linewidth]{fig/hscVideo_ours_10_10000.jpg}
    \caption{Ours}
    \label{fig_adapt_ours}
\end{subfigure}
\caption{HSC dataset qualitative denoising results. \ref{fig_adapt_sid} to \ref{fig_adapt_rns} are the outputs of some representative denoising models. \ref{fig_adapt_jpg} is the JPG image exported from the high-speed camera. \ref{fig_adapt_ours} is the our denoising results. Every four images as a group, the shutter speed from left to the right are: $1/_{500} (s)$, $1/_{1k} (s)$, $1/_{5k} (s)$, $1/_{10k} (s)$.}.
\label{fig_adapt}
\end{figure*}

Figure \ref{fig_pixel_intensity} shows the raw sensor value distributions $p_S(\bar{X})$ for the different shutter speed $S$ of our dataset. 
The raw intensity values returned by the sensor decrease with higher shutter speeds due to lesser light exposure. 
Figure \ref{fig_average_noise} shows the decomposition of the noise energy into both components as a function of shutter speed, and Figure \ref{fig_bias_noise_energy} shows the relative contribution of signal-independent noise to the total noise in our static video frames. This ratio is defines as $\frac{E_{SI}}{E}$.
As the shutter speed increases, 
the signal independent noise component becomes increasingly preponderant:
For the very noisy frames recorded at $1/_{10k} (s)$ shutter speed, the signal-independant noise component accounts for 98\% of the total noise energy, which explains the characteristic streak noise patterns observed in Figure \ref{fig_video}.

\subsection{Raw Frame Synthesis}
\label{sec_noise}

The procedure used to generate our training dataset proceeds in two steps, as illustrated in Figure \ref{fig_noise_synthesis}. 
In a first step, we synthesize ground-truth long-exposure raw frames from the JPG images of the COCO dataset by combining and adapting the methods of UPI \cite{brooks2019unprocessing} %
and InvISP \cite{xing2021invertible}. %
In a second step, we combine the generated long-exposure raw with our collected bias frames to synthesize noisy short-exposure frames 
following the procedure of the RNS noise model \cite{zhang2021rethinking}. 

\paragraph{Long-Exposure Raw Frame Reconstruction.}
In UPI \cite{brooks2019unprocessing}, %
the authors propose a generic inverse procedure to reconstruct RAW images from JPG images.
The UPI procedure consists in sequentially applying the 5 steps of inverse tone mapping, 
gamma decompression, 
reverse color correction, 
reverse white balance 
and reverse digital gain.
However, directly applying UPI to the 8-bit JPG images result in sparse raw value histograms, which we suspect to not capture the full extent of the visual information contained in dense 12-bit RAW images from a high-speed camera.

Recently, the InvISP \cite{xing2021invertible} method has been proposed to reconstruct
high-bit RAW images using a reversible CNN.
Unfortunately, InvISP is only meant to reconstruct RAW images from the given camera from which training images have been recorded so that it cannot directly be applied to extract RAW images from the COCO dataset.
Instead, we propose to use the pre-trained InvISP as a reconstructor for high bit data, before applying the UPI method to the reconstructed high bit RAW images.
This procedure is described in further details in the Appendix.

\paragraph{Noisy Short-Exposure Frame Synthesis.}
Given a clear long-exposure raw, we now aim to generate noisy short-exposure to be used as input images to our training procedure.
In RNS \cite{zhang2021rethinking}, %
the authors divide raw images noise into two categories: 
signal-dependent noise and signal-independent noise:

\begin{equation}
X = (X_{S} + N_{SD}) + N_{SI}
\label{equ_si}
\end{equation}

where $X_{S}$ represents the signal component of an observed raw intensity $X$.
As discussed in Section \ref{sec_noise_ana}, %
high shutter speed recordings we are interested in denoising are dominated by the signal-independent noise component.
To model signal-independent noise, the authors of RNS have proposed to collect device-specific bias frames.
These frames are collected in a darkroom so as to prevent any light from hitting the camera sensor.
This way, only the raw sensor response is contained in the collected bias frames. 
Following the RNS noise synthesis model, 
we thus use our collected bias frames to model the signal-independent noise $N_b$.
For each training image, 
we randomly sample one of our bias frames and apply it to the image.

To model the signal-dependent noise component, 
the RNS model considers photon shot noise as the main component of signal-dependent noise, and neglect other known sources of signal-dependant noise. We follow their model and simulate photon shot noise $N_p$ following a Poisson distribution:

\begin{equation}
(X_{S} + N_{SD}) = \mathcal{P}\Bigg( \frac{X_h}{K} \Bigg)K
\label{equ_sd}
\end{equation}

where the system gain parameter $K$ of the camera was computed from field flat frames \cite{janesick1987charge} collected from our device.

In order to simulate different noise distributions for different exposure times, we simply applied Equation \ref{equ_si} to the long-exposure ground truth raw frames $X$ 
scaled down by the shutter speed amplification ratio $R$:

\begin{equation}
X = (\frac{X_{S}}{R} + N_{SD}) + N_{SI}
\label{equ_si}
\end{equation}

where $R$ represents the ratio of shutter speed increase between the simulated noise and the ground-truth.

\subsection{Denoising Pipeline}
\label{sec_denoise}

Our denoising pipeline features two learnable modules: the denoising model and the Mini-ISP model.
The denoising model is used to regress clear long-exposure raw images from the noisy short-exposure frames.
The Mini-ISP model is used to reproduce the processing from RAW to JPG of the camera's proprietary ISP instead of the libraw library used in previous studies \cite{wei2020physics, zhang2021rethinking}. %
As denoising model, we used the NAFNet model \cite{chen2022simple} in its RAW-to-RAW configuration. 
The Mini-ISP model is a small convolutional neural network
consisting of 5 layers of 3x3 convolution.

We start by splitting our set of static scene videos into a set of 6 videos used for testing, 
and a set of 34 videos used for training and validation.
This latter set is used for two purposes:
As a validation set to control overfitting of the denoising model, and as a training set for the Mini-ISP module.

\paragraph{Training Procedure.}
We train the NAFNet model to regress the long-exposure RAW images reconstructed from the COCO dataset from their corresponding noisy short-exposure images.
We train the model to minimize an L1 loss function for $300,000$ iterations with the Adam optimizer, using an initial learning rate of $2\times 10^{-4}$ and a cosine learning rate schedule.
At each iteration, we randomly sample a batch of 1 image pairs and bias frames from the training. 
We resize the image pairs to a resolution of $640\times640$ pixels and randomly crop similar resolution patches from the sampled bias frames. We apply random horizontal and vertical flipping data augmentation.
We control overfitting using the validation subset of 34 static scenes collected with the HSC device.

The Mini-ISP module is introduced as a replacement for the proprietary ISP of our HSC device, to which we do not have access.
We trained the Mini-ISP to regress the RGB values of JPG images from their corresponding raw images
using the training static scenes recorded at $1/_{100} (s)$ shutter speed.
For space constraint, we have included the details of the Mini-ISP training to the Appendix.

\begin{figure}[t]
\centering
\begin{subfigure}{.5\textwidth}
	\includegraphics[width=0.45\textwidth]{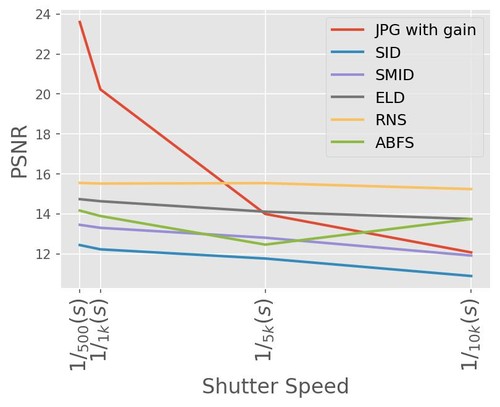}
	\includegraphics[width=0.45\textwidth]{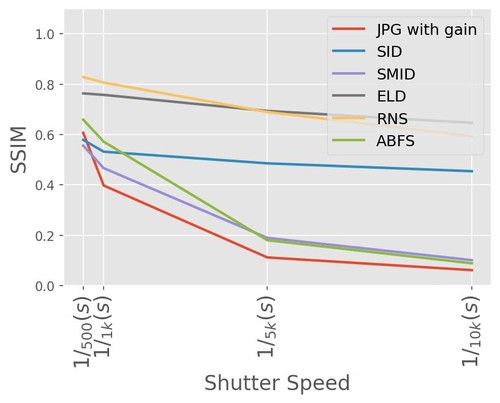}
\end{subfigure}
\caption{HSC dataset RGB image quantitative denoising results on existing denoising model.}
\label{fig_exist_res}
\end{figure}

\paragraph{Evaluation.}
To quantitatively evaluate our model, 
we consider the long-exposure static videos recorded at $1/_{100} (s)$ shutter speed 
to be our ground-truth.
Given a noisy raw frame  recorded at high shutter speed,
we start by denoising the noisy frame with the trained NAFNet, 
apply the white balance adjustment, and then process the denoised 
raw frame using the Mini-ISP to compute the final RGB output.
We use the PSNR and SSIM metrics to evaluate their similarity to 
the long-exposure RGB image exported in JPG.
In addition, we isolate the impact of the denoising model 
by evaluating the PSNR and SSIM of denmoised raw frames 
to the long exposure ground-truth raw frame.
This allows us to compare our denoising results to other existing methods.

\section{Experiments}
\subsection{Evaluation of Existing Denoising Model}
\label{sec_eval_exist}

We start by evaluating the performance of current state-of-the-art underexposed image denoising models on our HSC dataset.
Figure \ref{fig_adapt} illustrates the denoising results of RAW image denoising models \cite{chen2018learning, chen2019seeing, wei2020physics, dong2022abandoning, lamba2021restoring, zhang2021rethinking}.
Higher resolution visualizations of theses models outputs are shown for diverse scenes in the Appendix.
Despite a notable color bias, 
most models are able to denoise clear images for the lower shutter speed range, 
from $1/_{500} (s)$ to $1/_{1k} (s)$, 
which correspond to the noise levels of most existing datasets (cf. Figure \ref{fig_noise}). %
However, for higher shutter speed (i.e.; $1/_{5k} (s)$ and $1/_{10k} (s)$), 
the output images tend to 
display either blurring effects or streak noise patterns similar to that of our bias frames.
These results suggest that existing models trained on publicly available datasets do not efficiently address the underexposure noise regime of high shutter speed recordings in which the noise is dominated by a device-specific signal-independent component.
Figure \ref{fig_exist_res} shows the quantitative evaluation of these results on our static scene videos.
As a baseline for comparison, we gained the JPG frames exported for each shutter speed to compensate for the loss of brightness incurred by shorter exposure times. 
The PSNR metric is sensitive to color bias which explains most of the poor performance of these models.
Looking at the SSIM metric, we found that models trained using noise synthesis approaches (ELD and RNS) tend to outperform other methods trained using observed noise-clean image pairs for all shutter speed.
These results, combined with the noise analysis performed in Section \ref{sec_noise_ana}, have motivated our approach to train existing models on noisy frames synthesized with device-specific bias frames.

\begin{figure*}[htpb]
\centering
\begin{subfigure}[t]{0.49\textwidth}
\centering
	\includegraphics[width=0.49\textwidth]{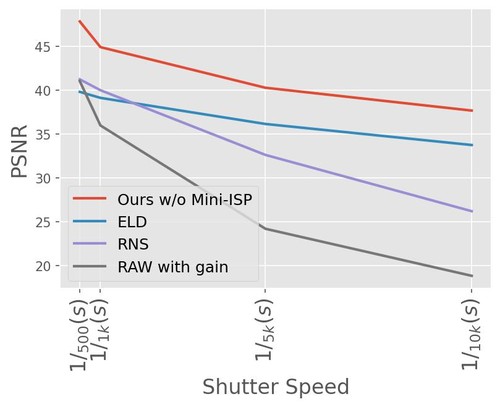}
	\includegraphics[width=0.49\textwidth]{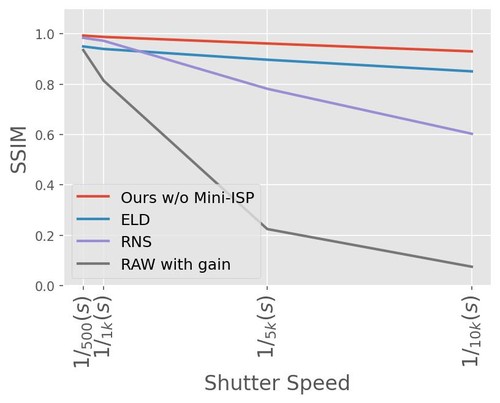}
	\caption{Denoising results for RAW images of HSC dataset. The results of training with our method w/o Mini-ISP on COCO dataset, the results of using pre-trained ELD, the results of using pre-trained RNS and the results of RAW images exported from high-speed cameras with gain are compared. }
	\label{fig_main_raw}
\end{subfigure}\hfill
\begin{subfigure}[t]{0.49\textwidth}
\centering
	\includegraphics[width=0.49\textwidth]{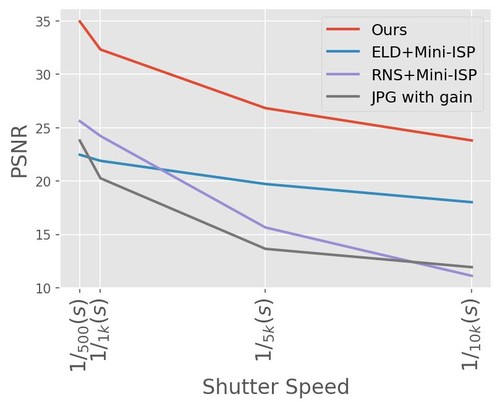}
	\includegraphics[width=0.49\textwidth]{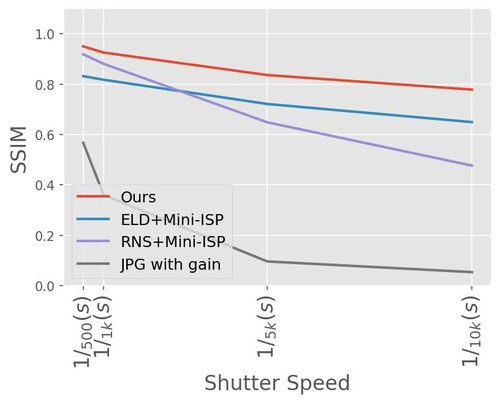}
	\caption{Denoising results for RGB images of HSC dataset. The results of training with our method on COCO dataset, pre-trained ELD+Mini-ISP, pre-trained RNS+Mini-ISP and JPG images exported from high-speed camera with gain are compared.}
	\label{fig_main_rgb}
\end{subfigure}
\caption{HSC dataset main denoising results.}
\label{fig_main_denoising}
\end{figure*}

\subsection{Main Denoising Results}
\label{sec_main_denoise}

Figure \ref{fig_main_raw} and Figure \ref{fig_main_rgb} summarize our main results.
In Figure \ref{fig_main_raw}, we aime to isolate the results of the denoising step of our pipeline: 
we evaluate the PSNR and SSIM of the raw frames outputted by the NAFNet model using the raw frames captured at $1/100 (s)$ shutter speed as ground-truth. %
In Figure \ref{fig_main_rgb}, we evaluate the PSNR and SSIM in RGB space after the full  pipeline including the processing of the Mini-ISP. We compare our results to those of the gained JPG images and denoising results of the two best-performing pretrained models identified in the previous section (ELD and RNS).

Our method achieves higher denoising accuracy than other methods accross the range of shutter speed. 
Across all metrics, the results of our method at $1/_{10k} (s)$ shutter speed are similar to those of other methods at $1/_{500} (s)$ shutter speed.

Of particular interest to us are the results of the RNS model. The results of the pre-trained RNS model we used were computed using the U-Net architecture proposed in \cite{chen2018learning} trained with synthetic noise generated from the black frames of a different camera device. 
To shed some light into the impact of the different components on our results, we conduct an ablation study in the following section. %

\subsection{Ablation Experiments}
\label{sec_abl}

We focus on the $1/_{10k} (s)$ shutter speed for our ablation study, which represents the hardest denoising case, and summarize our results in Table \ref{tab_ablation}.
We quantify the improvements brought by different modeling components:
(1) Using NAFNet as the denoising model instead of the standard baseline UNet model used in most early studies \cite{zhang2021rethinking, chen2018learning, wei2020physics}.
(2) Using InvISP to extract high-bit reconstruction from the JPG images instead of directly applying the UPI procedure to the original JPG images.
(3) Integrating bias frames collected from our target device as the $SI$ noise component of the RNS noise synthesis procedure.
(4) Integrating the statistical generation of $SD$ noise to the RNS noise synthesis procedure.

The modeling of $SI$ noise component is by far the most impacting factor.
This concurs with the different observations made in the previous sections and the noise analysis conducted in Section \ref{sec_noise_ana}.
The second most impacting improvement came from using the more powerful NAFNet for denoising, instead of the baseline UNet.
Leveraging the high-bit reconstruction provded by InvISP to process the JPG files of the COCO dataset also further improves the results of our experiments.
Lastly, althoug integrating the $SD$ noise component did bring a slight numerical improvement, the impact of this noise component has indeed been shown to be much lesser than the impact of the $SI$ component.

\begin{table}[t]
\centering
\begin{tabular}{cccccc}
\toprule
\# & Model  & InvISP & SI & SD & PSNR / SSIM  \\
\midrule
1 & Unet   & \checkmark      & \checkmark  & \checkmark  & 22.34 / 0.7527 \\
2 &NAFNet & -      & \checkmark  & \checkmark  & 23.46 / 0.7770 \\
3 &NAFNet & \checkmark      & \checkmark  & -  & 23.69 / 0.7739 \\
4 &NAFNet & \checkmark      & -  & \checkmark  & 11.22 / 0.1946 \\
5 &NAFNet & \checkmark      & \checkmark  & \checkmark  & 23.80 / 0.7787
\\
\bottomrule
\end{tabular}
\caption{Ablation Experiment Results}
\label{tab_ablation}
\end{table}

\section{Conclusions}
The ability to record high-fidelity videos at high acquisition rates is central to the study of fast moving phenomena.
As high shutter recordings 
suffer from underexposure of the camera sensor to light,
this paper has proposed to treat the problem of high speed imaging as an underexposed image denoising problem.
Combining recent methods on underexposed image denoising,
we were able to generate video frames of similar clarity 
for shutter speed up to 20 times that of available methods,
without requiring expensive HSC data collection.

{\small
\bibliographystyle{plain}
\bibliography{egbib}

\begin{thebibliography}{10}

\bibitem{balsalobre2014concurrent}
Carlos Balsalobre-Fern{\'a}ndez, Carlos~M Tejero-Gonz{\'a}lez, Juan del
  Campo-Vecino, and Nicol{\'a}s Bavaresco.
\newblock The concurrent validity and reliability of a low-cost, high-speed
  camera-based method for measuring the flight time of vertical jumps.
\newblock {\em The Journal of Strength \& Conditioning Research},
  28(2):528--533, 2014.

\bibitem{brooks2019unprocessing}
Tim Brooks, Ben Mildenhall, Tianfan Xue, Jiawen Chen, Dillon Sharlet, and
  Jonathan~T Barron.
\newblock Unprocessing images for learned raw denoising.
\newblock In {\em Proceedings of the IEEE/CVF Conference on Computer Vision and
  Pattern Recognition}, pages 11036--11045, 2019.

\bibitem{buades2008nonlocal}
Antoni Buades, Bartomeu Coll, and Jean-Michel Morel.
\newblock Nonlocal image and movie denoising.
\newblock {\em International journal of computer vision}, 76(2):123--139, 2008.

\bibitem{chang2020learning}
Ke-Chi Chang, Ren Wang, Hung-Jin Lin, Yu-Lun Liu, Chia-Ping Chen, Yu-Lin Chang,
  and Hwann-Tzong Chen.
\newblock Learning camera-aware noise models.
\newblock In {\em European Conference on Computer Vision}, pages 343--358.
  Springer, 2020.

\bibitem{chen2019seeing}
Chen Chen, Qifeng Chen, Minh~N Do, and Vladlen Koltun.
\newblock Seeing motion in the dark.
\newblock In {\em Proceedings of the IEEE/CVF International Conference on
  Computer Vision}, pages 3185--3194, 2019.

\bibitem{chen2018learning}
Chen Chen, Qifeng Chen, Jia Xu, and Vladlen Koltun.
\newblock Learning to see in the dark.
\newblock In {\em Proceedings of the IEEE conference on computer vision and
  pattern recognition}, pages 3291--3300, 2018.

\bibitem{chen2018image}
Jingwen Chen, Jiawei Chen, Hongyang Chao, and Ming Yang.
\newblock Image blind denoising with generative adversarial network based noise
  modeling.
\newblock In {\em Proceedings of the IEEE conference on computer vision and
  pattern recognition}, pages 3155--3164, 2018.

\bibitem{chen2022simple}
Liangyu Chen, Xiaojie Chu, Xiangyu Zhang, and Jian Sun.
\newblock Simple baselines for image restoration.
\newblock {\em arXiv preprint arXiv:2204.04676}, 2022.

\bibitem{Charles2013}
Joey Conenna.
\newblock Canon-6d-datasets-for-learning-to-see-in-the-dark.
\newblock GitHub, 2018.

\bibitem{dabov2007image}
Kostadin Dabov, Alessandro Foi, Vladimir Katkovnik, and Karen Egiazarian.
\newblock Image denoising by sparse 3-d transform-domain collaborative
  filtering.
\newblock {\em IEEE Transactions on image processing}, 16(8):2080--2095, 2007.

\bibitem{dong2022abandoning}
Xingbo Dong, Wanyan Xu, Zhihui Miao, Lan Ma, Chao Zhang, Jiewen Yang, Zhe Jin,
  Andrew Beng~Jin Teoh, and Jiajun Shen.
\newblock Abandoning the bayer-filter to see in the dark.
\newblock In {\em Proceedings of the IEEE/CVF Conference on Computer Vision and
  Pattern Recognition}, pages 17431--17440, 2022.

\bibitem{ficek2019influence}
Martin Ficek, Zden{\v{e}}k Mal{\'a}n{\'\i}k, Michaela Mikuli{\v{c}}ov{\'a}, and
  Michal Gracla.
\newblock Influence of the shooting distance on the depth of penetration of the
  bullet into the replacement material for air gun weapons.
\newblock In {\em Annals of DAAAM and Proceedings of the International DAAAM
  Symposium}. Danube Adria Association for Automation and Manufacturing, DAAAM,
  2019.

\bibitem{foi2008practical}
Alessandro Foi, Mejdi Trimeche, Vladimir Katkovnik, and Karen Egiazarian.
\newblock Practical poissonian-gaussian noise modeling and fitting for
  single-image raw-data.
\newblock {\em IEEE Transactions on Image Processing}, 17(10):1737--1754, 2008.

\bibitem{han2016imaging}
Yuanyuan Han, Yi~Gu, Alex~Ce Zhang, and Yu-Hwa Lo.
\newblock Imaging technologies for flow cytometry.
\newblock {\em Lab on a Chip}, 16(24):4639--4647, 2016.

\bibitem{janesick1987charge}
James~R Janesick, Kenneth~P Klaasen, and Tom Elliott.
\newblock Charge-coupled-device charge-collection efficiency and the
  photon-transfer technique.
\newblock {\em Optical engineering}, 26(10):972--980, 1987.

\bibitem{jobson1997multiscale}
Daniel~J Jobson, Zia-ur Rahman, and Glenn~A Woodell.
\newblock A multiscale retinex for bridging the gap between color images and
  the human observation of scenes.
\newblock {\em IEEE Transactions on Image processing}, 6(7):965--976, 1997.

\bibitem{kim2019grdn}
Dong-Wook Kim, Jae Ryun~Chung, and Seung-Won Jung.
\newblock Grdn: Grouped residual dense network for real image denoising and
  gan-based real-world noise modeling.
\newblock In {\em Proceedings of the IEEE/CVF Conference on Computer Vision and
  Pattern Recognition Workshops}, pages 0--0, 2019.

\bibitem{lamba2021restoring}
Mohit Lamba and Kaushik Mitra.
\newblock Restoring extremely dark images in real time.
\newblock In {\em Proceedings of the IEEE/CVF Conference on Computer Vision and
  Pattern Recognition}, pages 3487--3497, 2021.

\bibitem{land1977retinex}
Edwin~H Land.
\newblock The retinex theory of color vision.
\newblock {\em Scientific american}, 237(6):108--129, 1977.

\bibitem{lin2014microsoft}
Tsung-Yi Lin, Michael Maire, Serge Belongie, James Hays, Pietro Perona, Deva
  Ramanan, Piotr Doll{\'a}r, and C~Lawrence Zitnick.
\newblock Microsoft coco: Common objects in context.
\newblock In {\em European conference on computer vision}, pages 740--755.
  Springer, 2014.

\bibitem{liu2022research}
Bowen Liu, Chunming Wang, Gaoyang Mi, and Wei Zhang.
\newblock Research on the mechanism of laser removal for the oxide layer in
  ta15 titanium alloy by analyzing in-situ high-speed camera and product
  characteristics.
\newblock {\em Materials Chemistry and Physics}, 276:125325, 2022.

\bibitem{liu2009novel}
GuoJun Liu, XiangLong Tang, Heng-Da Cheng, JianHua Huang, and JiaFeng Liu.
\newblock A novel approach for tracking high speed skaters in sports using a
  panning camera.
\newblock {\em Pattern recognition}, 42(11):2922--2935, 2009.

\bibitem{loza2013automatic}
Artur {\L}oza, David~R Bull, Paul~R Hill, and Alin~M Achim.
\newblock Automatic contrast enhancement of low-light images based on local
  statistics of wavelet coefficients.
\newblock {\em Digital Signal Processing}, 23(6):1856--1866, 2013.

\bibitem{marsh2021time}
Andrew~W Marsh, Gwendolyn~T Wang, Jeffery~D Heyborne, Daniel~R Guildenbecher,
  and Yi~Chen Mazumdar.
\newblock Time-resolved size, velocity, and temperature statistics of aluminum
  combustion in solid rocket propellants.
\newblock {\em Proceedings of the Combustion Institute}, 38(3):4417--4424,
  2021.

\bibitem{mikami2018high}
Hideharu Mikami, Cheng Lei, Nao Nitta, Takeaki Sugimura, Takuro Ito, Yasuyuki
  Ozeki, and Keisuke Goda.
\newblock High-speed imaging meets single-cell analysis.
\newblock {\em Chem}, 4(10):2278--2300, 2018.

\bibitem{pan2016full}
Bing Pan, Liping Yu, Yongqi Yang, Weidong Song, and Licheng Guo.
\newblock Full-field transient 3d deformation measurement of 3d braided
  composite panels during ballistic impact using single-camera high-speed
  stereo-digital image correlation.
\newblock {\em Composite Structures}, 157:25--32, 2016.

\bibitem{pizer1987adaptive}
Stephen~M Pizer, E~Philip Amburn, John~D Austin, Robert Cromartie, Ari
  Geselowitz, Trey Greer, Bart ter Haar~Romeny, John~B Zimmerman, and Karel
  Zuiderveld.
\newblock Adaptive histogram equalization and its variations.
\newblock {\em Computer vision, graphics, and image processing},
  39(3):355--368, 1987.

\bibitem{richecoeur2008experimental}
F~Richecoeur, S~Ducruix, P~Scouflaire, and S~Candel.
\newblock Experimental investigation of high-frequency combustion instabilities
  in liquid rocket engine.
\newblock {\em Acta Astronautica}, 62(1):18--27, 2008.

\bibitem{settles2006high}
Gary~S Settles.
\newblock High-speed imaging of shock waves, explosions and gunshots: new
  digital video technology, combined with some classic imaging techniques,
  reveals shock waves as never before.
\newblock {\em American Scientist}, 94(1):22--31, 2006.

\bibitem{song2019experimental}
GuangChun Song, YuXing Li, WuChang Wang, Shuai Liu, XiaoYu Wang, ZhengZhuo Shi,
  and Shupeng Yao.
\newblock Experimental investigation on the microprocess of hydrate particle
  agglomeration using a high-speed camera.
\newblock {\em Fuel}, 237:475--485, 2019.

\bibitem{stevens2011rollover}
Don~C Stevens, Stephen Arndt, Leda Wayne, Mark Arndt, Robert Anderson, Joseph
  Manning, and Russell Anderson.
\newblock Rollover crash test results: steer-induced rollovers.
\newblock Technical report, SAE Technical Paper, 2011.

\bibitem{thoroddsen2008high}
Sigurdur~T Thoroddsen, Takeharu~Goji Etoh, and Kohsei Takehara.
\newblock High-speed imaging of drops and bubbles.
\newblock {\em Annu. Rev. Fluid Mech.}, 40:257--285, 2008.

\bibitem{wang2019enhancing}
Wei Wang, Xin Chen, Cheng Yang, Xiang Li, Xuemei Hu, and Tao Yue.
\newblock Enhancing low light videos by exploring high sensitivity camera
  noise.
\newblock In {\em Proceedings of the IEEE/CVF International Conference on
  Computer Vision}, pages 4111--4119, 2019.

\bibitem{wang2011high}
Yan~Zhao Wang, Mu~Xi Lei, Zheng~Bao Lei, and Jian~Jun Ling.
\newblock High speed photography system of vehicle/barrier crash testing
  laboratory.
\newblock In {\em Advanced Materials Research}, volume 255, pages 1745--1749.
  Trans Tech Publ, 2011.

\bibitem{wei2020physics}
Kaixuan Wei, Ying Fu, Jiaolong Yang, and Hua Huang.
\newblock A physics-based noise formation model for extreme low-light raw
  denoising.
\newblock In {\em Proceedings of the IEEE/CVF Conference on Computer Vision and
  Pattern Recognition}, pages 2758--2767, 2020.

\bibitem{winter2017high}
F~Winter, Simona Silvestri, Maria~Palma Celano, Gregor Schlieben, and O~Haidn.
\newblock High-speed and emission imaging of a coaxial single element gox/gch4
  rocket combustion chamber.
\newblock In {\em European Conference for Aeronautics and Space Sciences},
  2017.

\bibitem{wirth2018analysis}
Florian Wirth, Samuel Arpagaus, and Konrad Wegener.
\newblock Analysis of melt pool dynamics in laser cladding and direct metal
  deposition by automated high-speed camera image evaluation.
\newblock {\em Additive Manufacturing}, 21:369--382, 2018.

\bibitem{witte2008biomechanical}
Kerstin Witte, Peter Emmermacher, and Marion Lessau.
\newblock Biomechanical measuring stations to solve practical problems in
  karate sport.
\newblock In {\em ISBS-Conference Proceedings Archive}, 2008.

\bibitem{xing2021invertible}
Yazhou Xing, Zian Qian, and Qifeng Chen.
\newblock Invertible image signal processing.
\newblock In {\em Proceedings of the IEEE/CVF Conference on Computer Vision and
  Pattern Recognition}, pages 6287--6296, 2021.

\bibitem{zhang2021rethinking}
Yi~Zhang, Hongwei Qin, Xiaogang Wang, and Hongsheng Li.
\newblock Rethinking noise synthesis and modeling in raw denoising.
\newblock In {\em Proceedings of the IEEE/CVF International Conference on
  Computer Vision}, pages 4593--4601, 2021.

\end{thebibliography}
}

\begin{figure*}[t]
  \centering
  \includegraphics[width=1\linewidth]{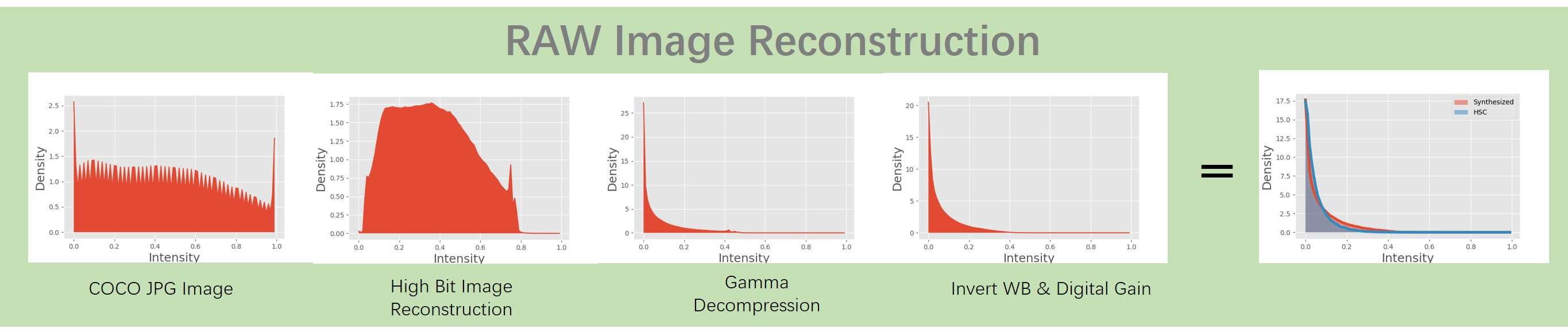}
  \caption{Evolution of the pixel intensity distribution through the reconstruction procedure.}
\label{fig_raw_process}
\end{figure*}

\begin{figure*}[t]
\centering

\begin{subfigure}[t]{0.24\textwidth}
\centering
    \includegraphics[width=1\textwidth]{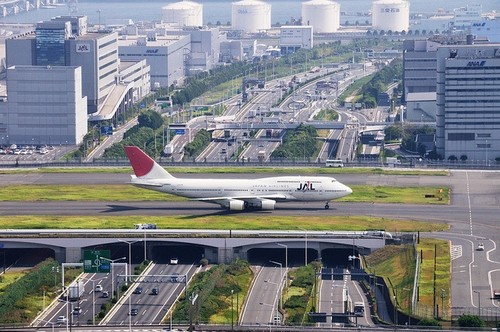}
    \caption{COCO JPG}
    \label{coco_jpg_gt}
\end{subfigure}
\begin{subfigure}[t]{0.24\textwidth}
\centering
    \includegraphics[width=1\textwidth]{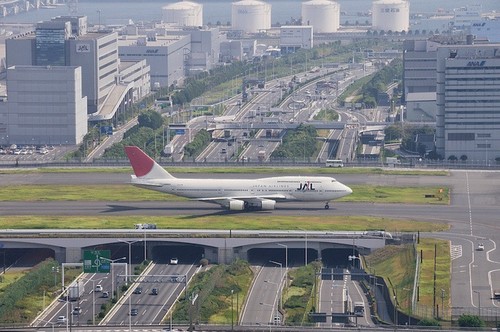}
    \caption{UPI invert tone mapping output}
    \label{coco_upi}
\end{subfigure}
\begin{subfigure}[t]{0.24\textwidth}
\centering
    \includegraphics[width=1\textwidth]{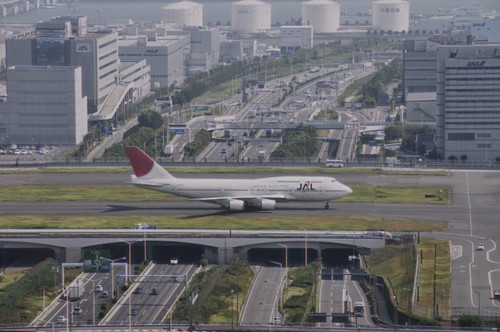}
    \caption{Pre-trained InvISP output}
    \label{coco_invisp}
\end{subfigure}
\begin{subfigure}[t]{0.24\textwidth}
\centering
    \includegraphics[width=1\textwidth]{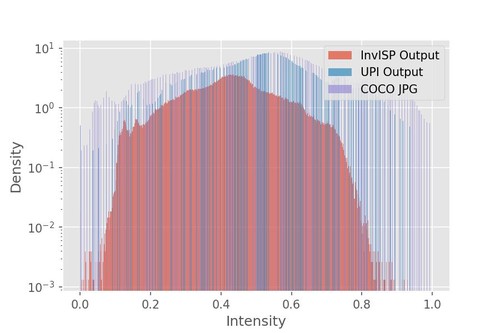}
    \caption{Image Histogram}
    \label{invisp_hist}
\end{subfigure}
\caption{Figure \ref{coco_jpg_gt} is the COCO JPG image. Figure \ref{coco_upi} is the output after processing the COCO JPG using the invert tone mapping proposed by UPI. Figure \ref{coco_invisp} is the output after processing the COCO JPG using the pre-trained InvISP. Figure \ref{invisp_hist} is the image histogram of the Figure \ref{coco_jpg_gt} to Figure \ref{coco_invisp}.}
\label{coco_img}
\end{figure*}

\begin{table*}[b]
\centering
\begin{tabular}{ccccccccc}
\toprule
\multirow{2}{*}{\#}  & \multirow{2}{*}{Model} & \multirow{2}{*}{InvISP} & \multirow{2}{*}{SI} & \multirow{2}{*}{SD} & \multicolumn{4}{c}{PSNR / SSIM}                              \\
        &               &                         &                     &                     & $1/_{500} (s)$           & $1/_{1k} (s)$         & $1/_{5k} (s)$            & $1/_{10k} (s)$          \\
\midrule
1 & Unet                   & \checkmark                       & \checkmark                   & \checkmark                   & 34.09 / 0.9396 & 31.37 / 0.9127  & 25.11 / 0.8162  & 22.34 / 0.7527 \\
2 & NAFNet                 & -                       & \checkmark                   & \checkmark                   & 34.44 / 0.9429  & 31.83 / 0.9219  & 25.23 / 0.8237  & 23.46 / 0.7770 \\
3 & NAFNet                 & \checkmark                       & \checkmark                   & -                   & 34.56 / 0.9447  & 32.14 / 0.9249  & 25.36 / 0.8329  & 23.69 / 0.7739  \\
4 & NAFNet                 & \checkmark                       & -                   & \checkmark                   & 29.17 / 0.7267  & 24.08 / 0.5219 & 15.19 / 0.2489 & 11.22 / 0.1946 \\
5 & NAFNet                 & \checkmark                       & \checkmark                   & \checkmark                   & \textbf{34.95 / 0.9503}  & \textbf{32.33 / 0.9256}  & \textbf{26.83 / 0.8364}  & \textbf{23.80 / 0.7787} 
\\
\bottomrule
\end{tabular}
\caption{Ablation Experiment Results}
\label{tab_ablation}
\end{table*}

\newpage
\appendix
\section{RAW Image Reconstruction Procedure}

We combined the UPI \cite{brooks2019unprocessing} and InvISP \cite{zhang2021rethinking} 
methods to reconstruct long-exposure RAW frames from the JPG images of the COCO dataset.
Figure \ref{fig_raw_process} illustrates the evolution of 
the pixel intensity distribution along each step of our procedure.
We start by first applying the InvISP model to reconstruct high-bit encoding of the low-bit quantized JPG images, and then proceed with the steps of UPI.
We slightly modify some of the original UPI procedure steps 
in order to adapt to the InvISP-processed output:

\paragraph{Tone mapping and Gamma Decompression}
We did not apply the first step (invert tone mapping) of the UPI procedure.
Figure \ref{coco_img} compares the intensity distribution after processing a COCO JPG image 
using the invert tone mapping proposed by UPI and using a pre-trained InvISP.
Not only does the InvISP output reconstructs a dense distribution, 
it also shows a reduced dynamic range compared to the UPI invert tone mapping output.
We thus skip the tone mapping step of the UPI procedure 
and perform gamma decompression on the pre-trained InvISP output images.
The parameters of gamma decompression was empirically set to 3 so as to better 
match the target HSC distribution.

\paragraph{Invert WB \& Gain.}
Finally, we reconstruct the raw using the same method proposed by UPI, by applying the inverted gain, and the inverted white balance.
Note that, in the same way as UPI, the red gain and blue gain used for the inverted white balance were randomly sampled from a value range inferred from our target HSC dataset.

Applying these steps, we were able to reconstruct long-exposure RAW frames from COCO's JPG with an intensity distribution similar to our HSC dataset, as illustrated in the right-most plot of FIgure 1.

\section{Noisy Short-Exposure Frame Synthesis.}
\begin{figure*}[t]
\centering
\begin{subfigure}[t]{0.24\textwidth}
\centering
    \includegraphics[width=1\textwidth]{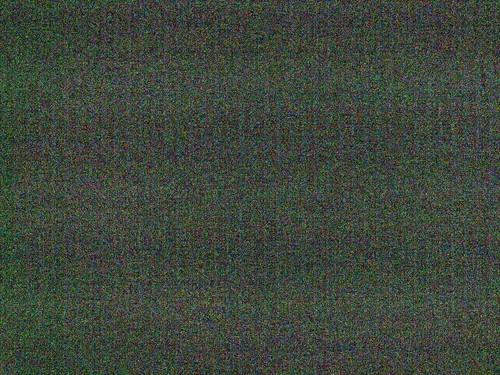}
    \caption{Bias Frame}
    \label{fig_bias_frame}
\end{subfigure}
\begin{subfigure}[t]{0.24\textwidth}
\centering
    \includegraphics[width=1\textwidth]{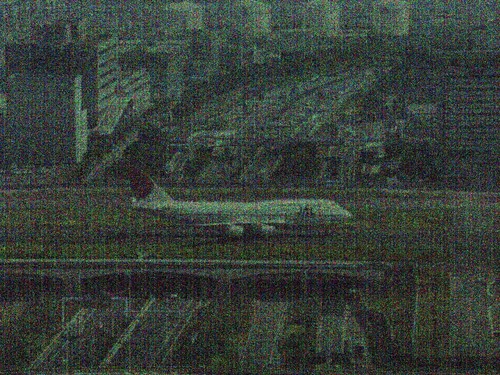}
    \caption{Synthesized Noise COCO Image}
    \label{fig_coco_sys}
\end{subfigure}
\begin{subfigure}[t]{0.24\textwidth}
\centering
    \includegraphics[width=1\textwidth]{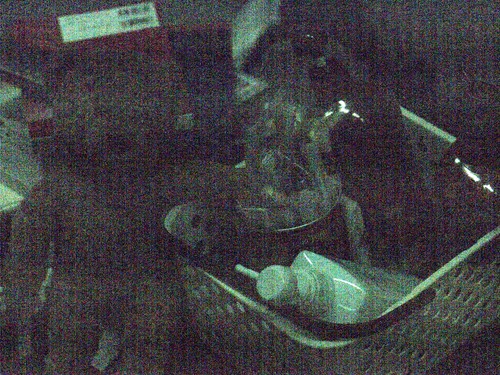}
    \caption{HSC Noise Image ($1/_{10k} (s)$)}
    \label{fig_hsc_noise}
\end{subfigure}

\begin{subfigure}[t]{0.24\textwidth}
\centering
    \includegraphics[width=1\textwidth]{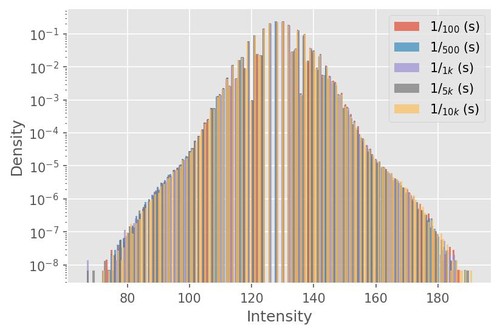}
    \caption{Bias Frames Distribution}
\end{subfigure}
\begin{subfigure}[t]{0.24\textwidth}
\centering
    \includegraphics[width=1\textwidth]{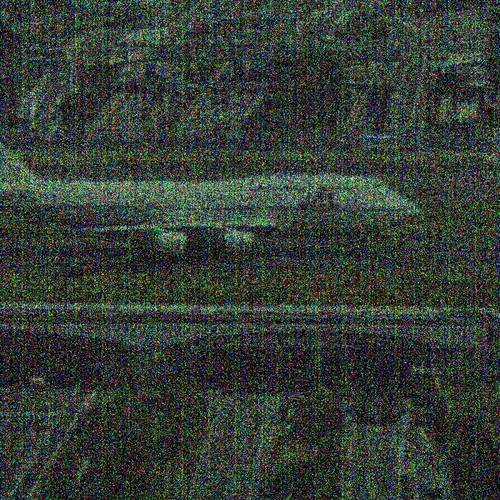}
    \caption{Cropping of Synthesized COCO Image.}
\end{subfigure}
\begin{subfigure}[t]{0.24\textwidth}
\centering
    \includegraphics[width=1\textwidth]{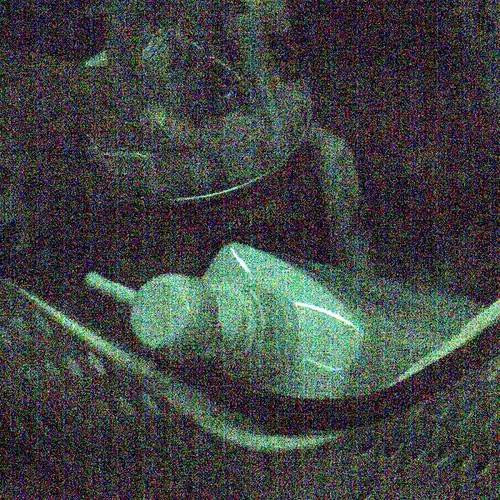}
    \caption{Cropping of HSC Noise Image ($1/_{10k} (s)$)}
    \label{fig_bias_hist}
\end{subfigure}
\caption{The high shutter speed(e.g., $1/_{10k}$) Raw image \ref{fig_hsc_noise} of the HSC dataset has obvious pattern noise, as shown in \ref{fig_bias_frame}.
We directly use bias frame as the signal-independent noise synthesized noise images can present the pattern noise well, as shown in Figure \ref{fig_coco_sys}. Figure \ref{fig_bias_hist} shows the distribution of bias frames collected using different shutter speeds. The different shutter speeds do not affect the distribution of bias frames, and the top of the distribution is concentrated at 129.}
\label{fig_miniisp_output}
\end{figure*}

\begin{figure*}[t]
\centering

\begin{subfigure}[t]{1\textwidth}
\centering
	\includegraphics[width=0.245\textwidth]{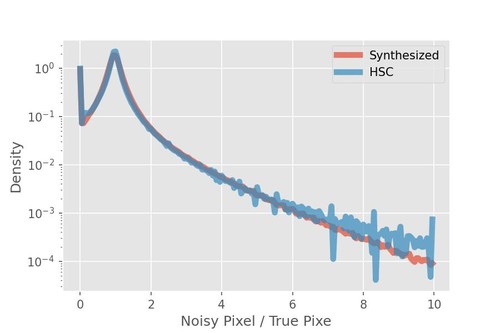}
	\includegraphics[width=0.245\textwidth]{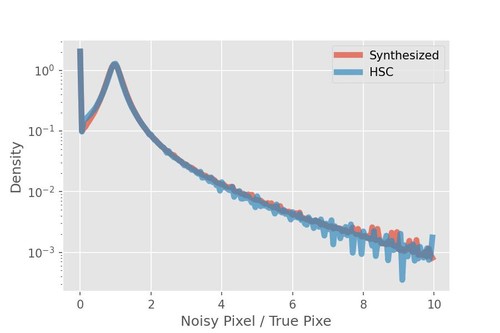}
	\includegraphics[width=0.245\textwidth]{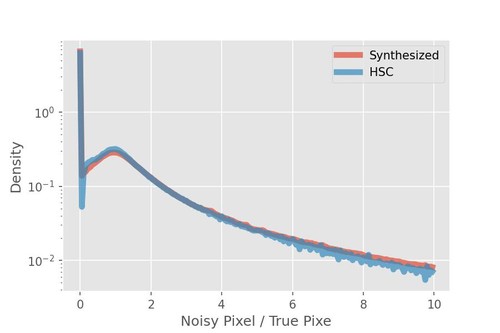}
    \includegraphics[width=0.245\textwidth]{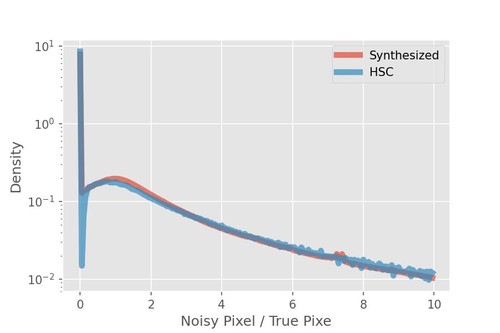}
\end{subfigure}
\begin{subfigure}[t]{0.245\textwidth}
\centering
$1/_{500} (s)$
\end{subfigure}
\begin{subfigure}[t]{0.245\textwidth}
\centering
$1/_{1k} (s)$
\end{subfigure}
\begin{subfigure}[t]{0.245\textwidth}
\centering
$1/_{5k} (s)$
\end{subfigure}
\begin{subfigure}[t]{0.245\textwidth}
\centering
$1/_{10k} (s)$
\end{subfigure}

\caption{The real HSC noise distribution and the synthetic COCO noise distribution are compared using the method proposed in \cite{chen2019seeing}. The shutter speeds from left to right are $1/_{500} (s)$, $1/_{1k} (s)$, $1/_{5k} (s)$, and $1/_{10k} (s)$, respectively.}
\label{fig_noise_hist}
\end{figure*}

Figure \ref{fig_hsc_noise} illustrates a noisy image recorded with our HSC at high shutter speeds.
This image shows a streak-like pattern noise very similar to the collected bias frame, as shown in Figure \ref{fig_bias_frame}.
In order to accurately synthesize similar noisy images, 
we have recorded bias frames from our target HSC device
and applied this noise to the long-exposure RAW frames 
following the procedure proposed in \cite{zhang2021rethinking}.
Figure \ref{fig_coco_sys} illustrates a noisy image synthesized from the COCO dataset.
The pattern noise of the camera sensor is successfully synthesized,
yielding images visually similar to the HSC dataset.
In Figure \ref{fig_noise_hist}, we show the evolution of the noise distribution of both
the HSC dataset and the synthesized dataset.
Our noise synthesis method yields a noise distribution similar to the target HSC data.

Figure \ref{fig_bias_hist} illustrates the distribution of bias frames obtained at different shutter speeds. 
No significant difference in the bias frames collected at different shutter speeds was found.

\section{Mini-ISP}

\begin{figure*}[b]
\centering
\begin{subfigure}[t]{0.2\textwidth}
\centering
    \includegraphics[width=1\textwidth]{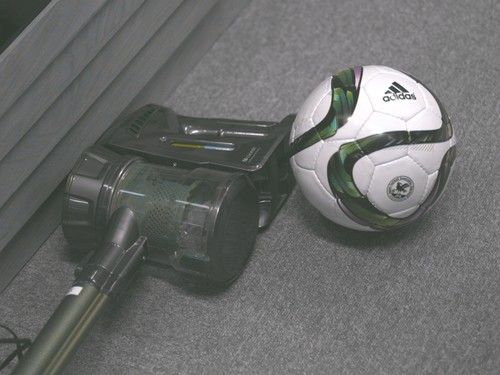}
    \caption{LibRaw default setting output}
    \label{fig_libraw_default}
\end{subfigure}
\begin{subfigure}[t]{0.2\textwidth}
\centering
    \includegraphics[width=1\textwidth]{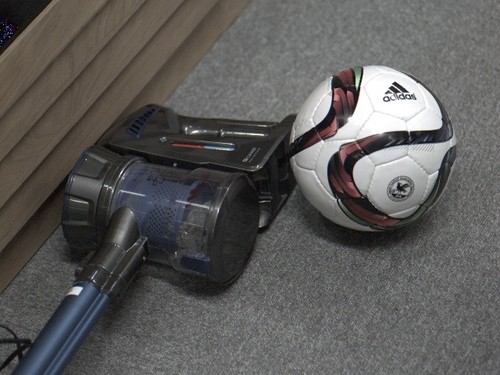}
    \caption{LibRaw output with correct debayering, black level, white balance}
    \label{fig_libraw}
\end{subfigure}
\begin{subfigure}[t]{0.2\textwidth}
\centering
    \includegraphics[width=1\textwidth]{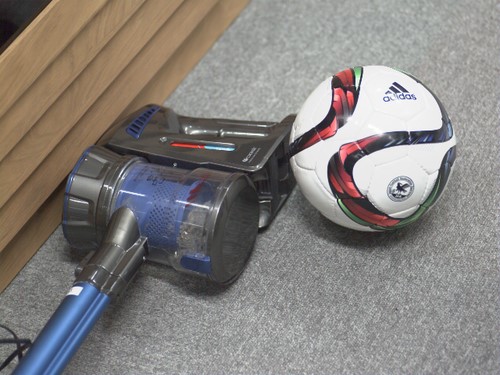}
    \caption{Mini-ISP output}
    \label{fig_mini}
\end{subfigure}
\begin{subfigure}[t]{0.2\textwidth}
\centering
    \includegraphics[width=1\textwidth]{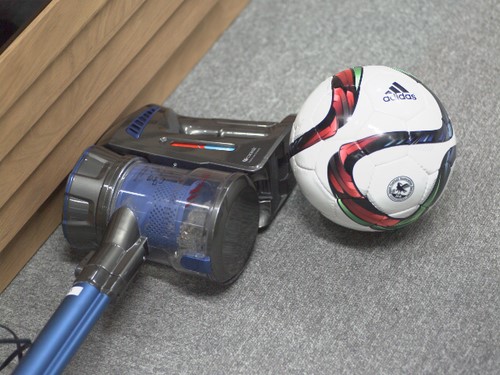}
    \caption{High-Speed Camera JPG}
    \label{fig_gt}
\end{subfigure}

\caption{Figure \ref{fig_libraw_default} shows the RAW image using the default libRAW to process the HSC dataset. Due to the inaccurate metadata of the HSC dataset, we cannot use the default libRAW to output the same image as Figure \ref{fig_gt}. Even though the bayer pattern and black level are set correctly, libRAW still does not output the correct brightness in Figure \ref{fig_libraw}. We use a Mini-ISP to fit the non-linear processing of the high-speed camera and are able to output images very similar to Figure \ref{fig_gt}.}
\label{fig_miniisp_output}
\end{figure*}

\begin{figure*}[htpb]
  \centering
  \includegraphics[width=0.5\linewidth]{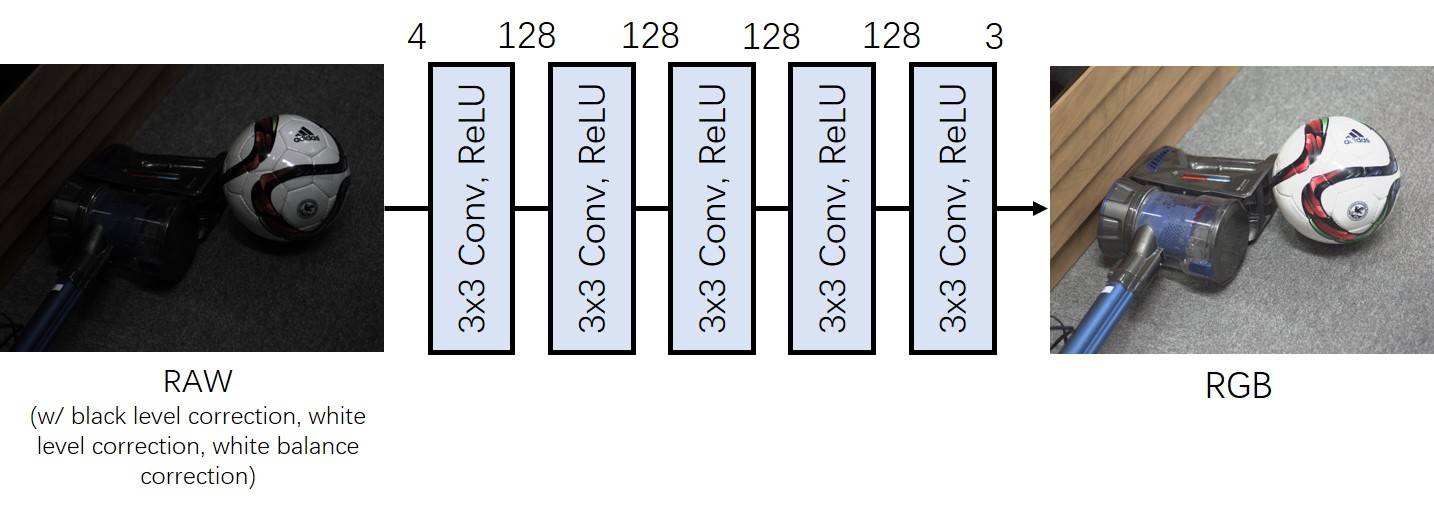}
  \caption{Mini-ISP Architecture.}
  
\label{fig_miniisp_arch}
\end{figure*}

\subsection{Motivation}
Following previous works, we started by using the Libraw library to process long-exposure RAW frames into RGB space.
However, directly applyin libraw to our RAW frames yielded sub-optimal results, as shown in Figure \ref{fig_libraw_default}.

Upon closer inspection, we found most of our problems were caused by the inaccurate metadata of the RAW images returned by our camera: the Bayer pattern and CCM matrix were missing, 
and we found the black levels were incorrect.

In Figure \ref{fig_libraw}, we show the results after using libRAW with correct debayering, black level and white balance processing.
Unfortunately, there is still a significant difference between Figure \ref{fig_libraw} and Figure \ref{fig_gt}, which we believe to be due to the different processing used by the high-speed camera proprietary ISP and libRAW.
In particular, the libRAW processed images show less crisp contrast and less varied colors.
Instead of using libRAW, we thus opted from learning the RGB space conversion procedure from the camera data.

\subsection{Mini-ISP Training}

To do so, we used a five-layer CNN, we refer to as the Mini-ISP.
Each layer is made of a $3 \times 3$ convolution with a ReLU activation function.
We used 128 channels at each layer and no pooling.
The architecture of Mini-ISP is illustrated in Figure \ref{fig_miniisp_arch}.

We trained this module to regress the JPG images recorded at $1/_{100} (s)$ shutter speed from their RAW frames.
The RAW frames were first processed by applying black level correction, white level correction and white balance correction.
We used the Adam optimizer with a cosine learning rate schedule using an initial learning rate set to $1\times10^{-2}$ for a total of $240,000$ training iterations.
The Mini-ISP was trained on the 34 static scenes of the training set and we controlled overfitting using the testing set of 6 static scenes.

\section{Additional HSC Dataset Denoising Visualizations}

\paragraph{Existing Model Evaluation}
Figure \ref{fig_existing} shows denoising results of publicly available 
raw denoising models and contrast it to our results, as discussed in the main paper.

\paragraph{Dynamic Video Denoising}
Figure \ref{fig_video} shows sample frames of our collected dynamical scenes:
(1) poking balloons with a razor blade, 
(2) hitting mineral water with high-speed gas, 
(3) plucking guitar strings and 
(4) high-speed rotating fan.
We can see that the JPG files exported directly from the camera gradually become underexposed as the shutter speed increases. 
After applying our proposed method to images taken at high shutter speeds
for denoising tasks, we can have clear and distinguishable results at $1/_{10k} (s)$ 
shutter speeds as well, 
although our model was not able to recover some fine-grained details
such as the thinnest cords of the guitar or parts of the fan frame.
The dynamic scene videos can be viewed at the following link: \textcolor[rgb]{1,0,0}{https://youtu.be/41QHzK2jCFo}

\section{Additional Ablation Experiments}
In Table \ref{tab_ablation}, we show the results of our ablation experiments for different shutter speeds.
Our proposed method (i.e., \# 5) has the best metrics at all shutter speeds.
For all shutter speed, the Signal-independent (SI) noise is the most impacting factor, 
although the relative improvement SI brings is most impressive for higher shutter speeds.
Surprisingly, the SD noise component is not as important as SI noise even for relatively low shutter speed,
in which SD noise was estimated to account for 25\% of the noise energy.
This suggests that a more efficient SD noise model may provide further improvements on the low shutter speed range.
The more powerful noise reduction model NAFNet \cite{chen2022simple} provides a small but constant improvement across the shutter speed range compared to the baseline architecture of many previous works \cite{zhang2021rethinking, chen2018learning, wei2020physics}, as does the InvISP \cite{xing2021invertible} high-bit reconstruction trick.

\begin{figure*}[t]
\begin{subfigure}[t]{1\textwidth}
\centering
    \includegraphics[width=0.24\textwidth]{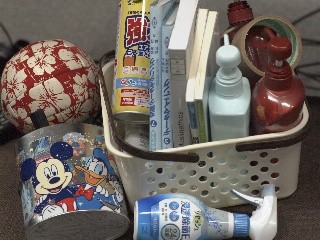}
    \includegraphics[width=0.24\textwidth]{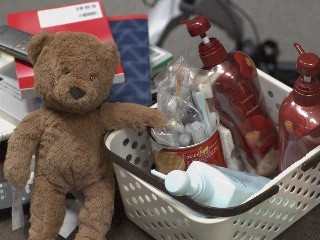}
\end{subfigure}
\\[1ex]
\centering
\begin{subfigure}[t]{0.24\textwidth}
\centering
    \includegraphics[width=0.49\textwidth]{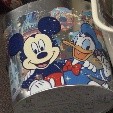}
    \includegraphics[width=0.49\textwidth]{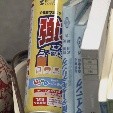}
\end{subfigure}
\begin{subfigure}[t]{0.24\textwidth}
\centering
    \includegraphics[width=0.49\textwidth]{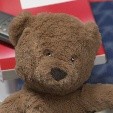}
    \includegraphics[width=0.49\textwidth]{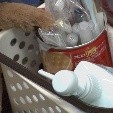}
\end{subfigure}

\begin{subfigure}[t]{1\textwidth}
\centering
\caption{Camera JPG Ground Truth}
\end{subfigure}
\end{figure*}

\begin{figure*}[t]\ContinuedFloat
\centering
\begin{subfigure}[t]{0.8\textwidth}
\centering
    \includegraphics[width=0.24\textwidth]{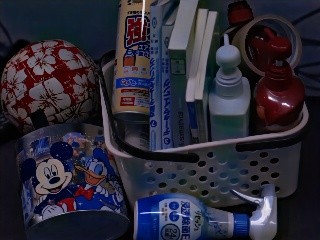}
    \includegraphics[width=0.24\textwidth]{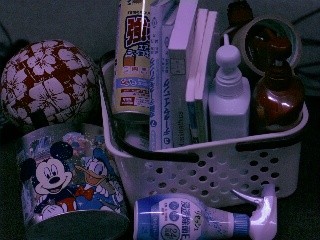}
    \includegraphics[width=0.24\textwidth]{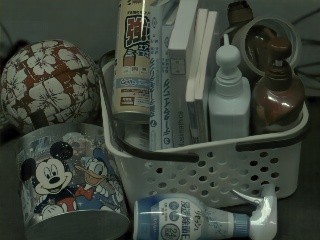}
    \includegraphics[width=0.24\textwidth]{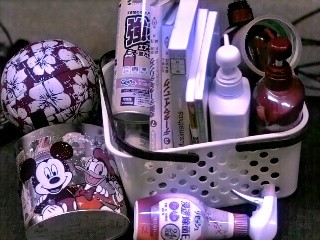}
\end{subfigure}
\\[1ex]
\begin{subfigure}[t]{0.192\textwidth}
\centering
    \includegraphics[width=0.48\textwidth]{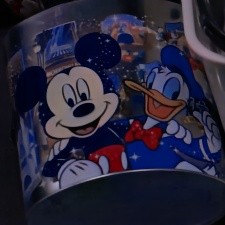}
    \includegraphics[width=0.48\textwidth]{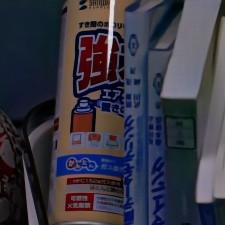}
\end{subfigure}
\begin{subfigure}[t]{0.192\textwidth}
\centering
    \includegraphics[width=0.48\textwidth]{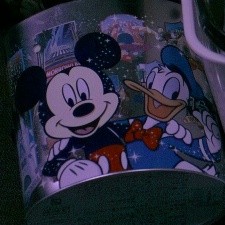}
    \includegraphics[width=0.48\textwidth]{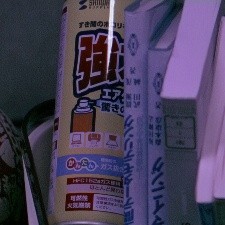}
\end{subfigure}
\begin{subfigure}[t]{0.192\textwidth}
\centering
    \includegraphics[width=0.48\textwidth]{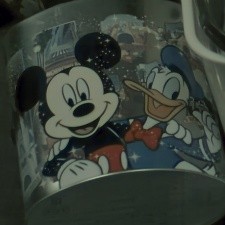}
    \includegraphics[width=0.48\textwidth]{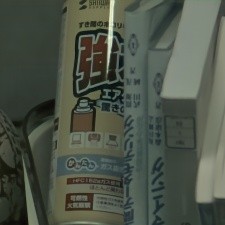}
\end{subfigure}
\begin{subfigure}[t]{0.192\textwidth}
\centering
    \includegraphics[width=0.48\textwidth]{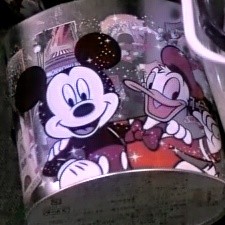}
    \includegraphics[width=0.48\textwidth]{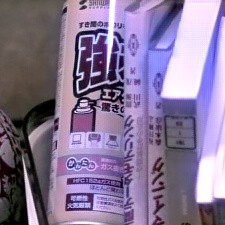}
\end{subfigure}
\\[3ex]

\begin{subfigure}[t]{0.8\textwidth}
\centering
    \includegraphics[width=0.24\textwidth]{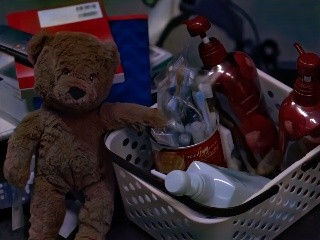}
    \includegraphics[width=0.24\textwidth]{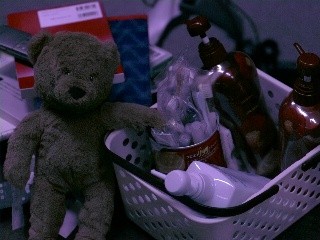}
    \includegraphics[width=0.24\textwidth]{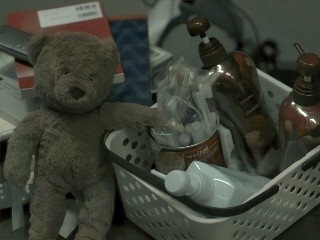}
    \includegraphics[width=0.24\textwidth]{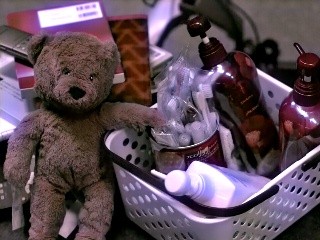}
\end{subfigure}
\\[1ex]
\begin{subfigure}[t]{0.192\textwidth}
\centering
    \includegraphics[width=0.48\textwidth]{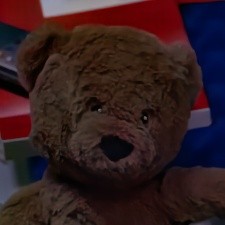}
    \includegraphics[width=0.48\textwidth]{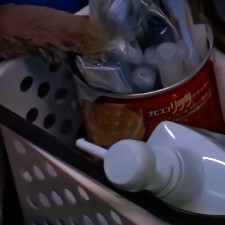}
\end{subfigure}
\begin{subfigure}[t]{0.192\textwidth}
\centering
    \includegraphics[width=0.48\textwidth]{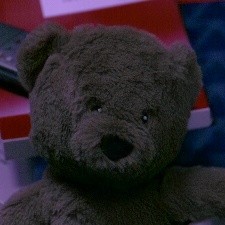}
    \includegraphics[width=0.48\textwidth]{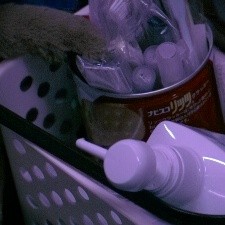}
\end{subfigure}
\begin{subfigure}[t]{0.192\textwidth}
\centering
    \includegraphics[width=0.48\textwidth]{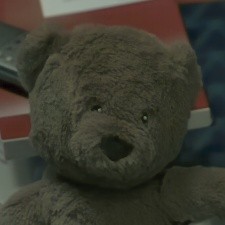}
    \includegraphics[width=0.48\textwidth]{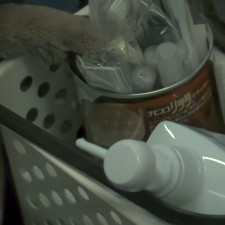}
\end{subfigure}
\begin{subfigure}[t]{0.192\textwidth}
\centering
    \includegraphics[width=0.48\textwidth]{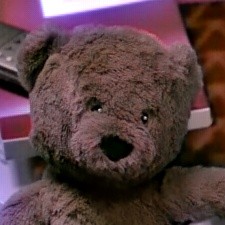}
    \includegraphics[width=0.48\textwidth]{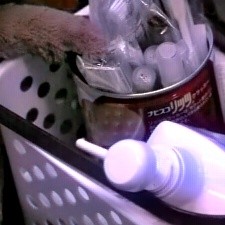}
\end{subfigure}

\begin{subfigure}[t]{0.192\textwidth}
\centering
SID \cite{chen2018learning}
\end{subfigure}
\begin{subfigure}[t]{0.192\textwidth}
\centering
SMID \cite{chen2019seeing}
\end{subfigure}
\begin{subfigure}[t]{0.192\textwidth}
\centering
ELD \cite{wei2020physics}
\end{subfigure}
\begin{subfigure}[t]{0.192\textwidth}
\centering
RED \cite{lamba2021restoring}
\end{subfigure}
\\[3ex]

\begin{subfigure}[t]{0.8\textwidth}
\centering
    \includegraphics[width=0.24\textwidth]{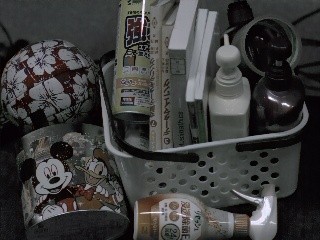}
    \includegraphics[width=0.24\textwidth]{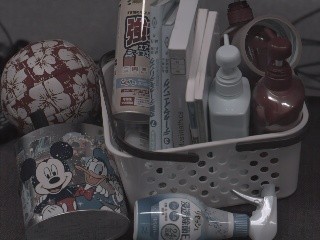}
    \includegraphics[width=0.24\textwidth]{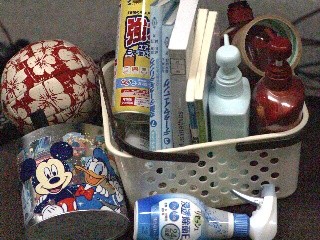}
    \includegraphics[width=0.24\textwidth]{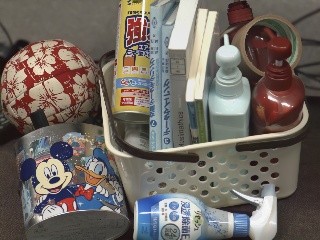}
\end{subfigure}
\\[1ex]
\begin{subfigure}[t]{0.192\textwidth}
\centering
    \includegraphics[width=0.48\textwidth]{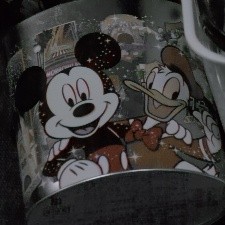}
    \includegraphics[width=0.48\textwidth]{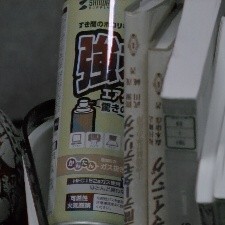}
\end{subfigure}
\begin{subfigure}[t]{0.192\textwidth}
\centering
    \includegraphics[width=0.48\textwidth]{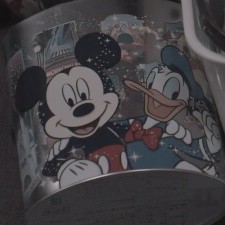}
    \includegraphics[width=0.48\textwidth]{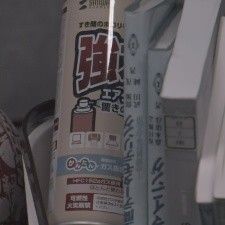}
\end{subfigure}
\begin{subfigure}[t]{0.192\textwidth}
\centering
    \includegraphics[width=0.48\textwidth]{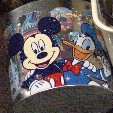}
    \includegraphics[width=0.48\textwidth]{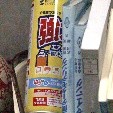}
\end{subfigure}
\begin{subfigure}[t]{0.192\textwidth}
\centering
    \includegraphics[width=0.48\textwidth]{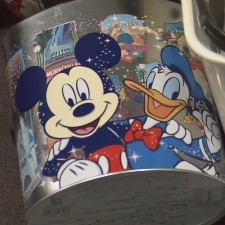}
    \includegraphics[width=0.48\textwidth]{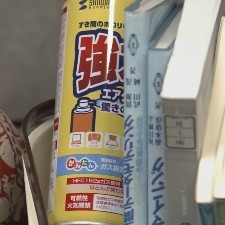}
\end{subfigure}
\\[3ex]

\begin{subfigure}[t]{0.8\textwidth}
\centering
    \includegraphics[width=0.24\textwidth]{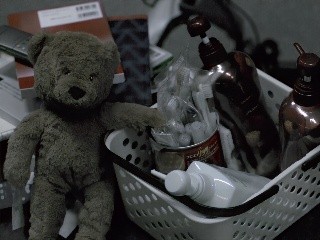}
    \includegraphics[width=0.24\textwidth]{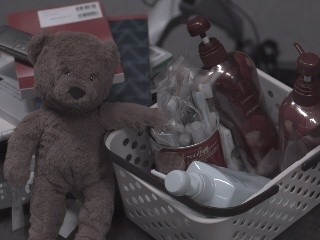}
    \includegraphics[width=0.24\textwidth]{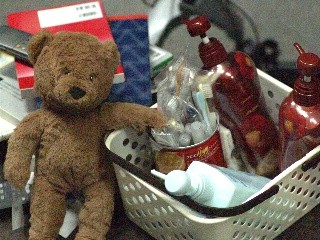}
    \includegraphics[width=0.24\textwidth]{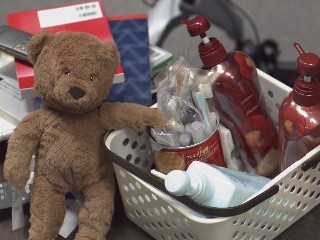}
\end{subfigure}
\\[1ex]
\begin{subfigure}[t]{0.192\textwidth}
\centering
    \includegraphics[width=0.48\textwidth]{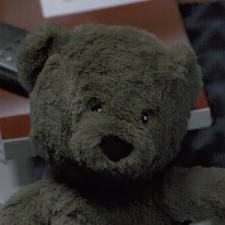}
    \includegraphics[width=0.48\textwidth]{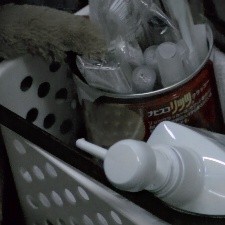}
\end{subfigure}
\begin{subfigure}[t]{0.192\textwidth}
\centering
    \includegraphics[width=0.48\textwidth]{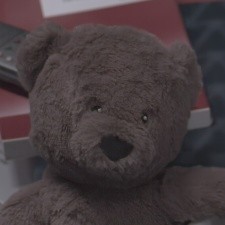}
    \includegraphics[width=0.48\textwidth]{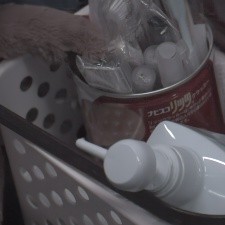}
\end{subfigure}
\begin{subfigure}[t]{0.192\textwidth}
\centering
    \includegraphics[width=0.48\textwidth]{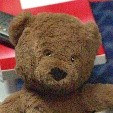}
    \includegraphics[width=0.48\textwidth]{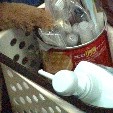}
\end{subfigure}
\begin{subfigure}[t]{0.192\textwidth}
\centering
    \includegraphics[width=0.48\textwidth]{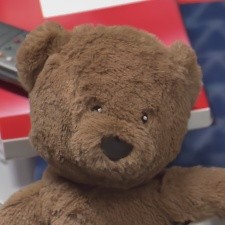}
    \includegraphics[width=0.48\textwidth]{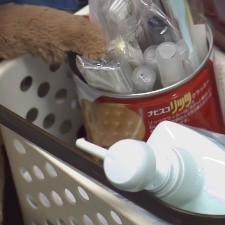}
\end{subfigure}

\begin{subfigure}[t]{0.192\textwidth}
\centering
ABFS \cite{dong2022abandoning}
\end{subfigure}
\begin{subfigure}[t]{0.192\textwidth}
\centering
RNS \cite{zhang2021rethinking}
\end{subfigure}
\begin{subfigure}[t]{0.192\textwidth}
\centering
Camera JPG
\end{subfigure}
\begin{subfigure}[t]{0.192\textwidth}
\centering
Ours
\end{subfigure}

\begin{subfigure}[t]{1\textwidth}
\centering
\caption{Denoising results for images with $1/_{500} (s)$ shutter speed.}
\end{subfigure}

\end{figure*}

\begin{figure*}[t]\ContinuedFloat
\centering
\begin{subfigure}[t]{0.8\textwidth}
\centering
    \includegraphics[width=0.24\textwidth]{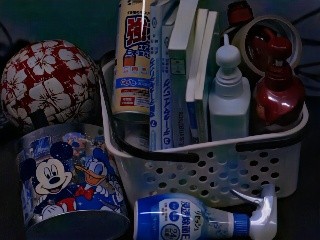}
    \includegraphics[width=0.24\textwidth]{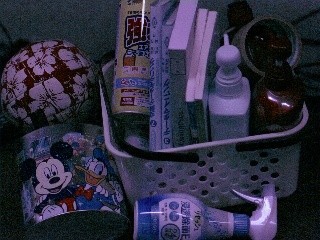}
    \includegraphics[width=0.24\textwidth]{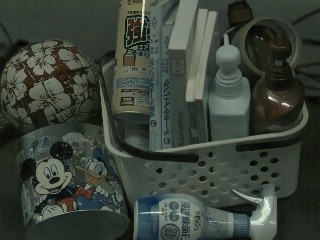}
    \includegraphics[width=0.24\textwidth]{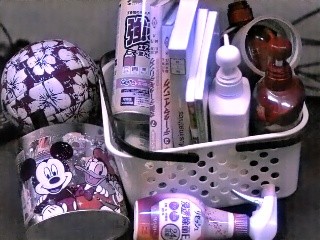}
\end{subfigure}
\\[1ex]
\begin{subfigure}[t]{0.192\textwidth}
\centering
    \includegraphics[width=0.48\textwidth]{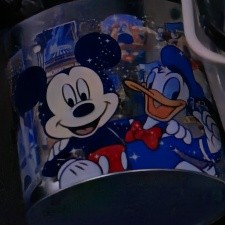}
    \includegraphics[width=0.48\textwidth]{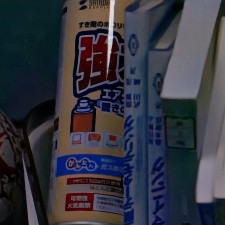}
\end{subfigure}
\begin{subfigure}[t]{0.192\textwidth}
\centering
    \includegraphics[width=0.48\textwidth]{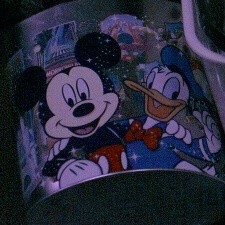}
    \includegraphics[width=0.48\textwidth]{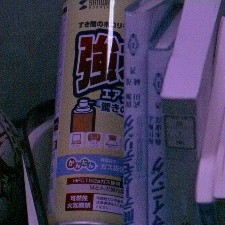}
\end{subfigure}
\begin{subfigure}[t]{0.192\textwidth}
\centering
    \includegraphics[width=0.48\textwidth]{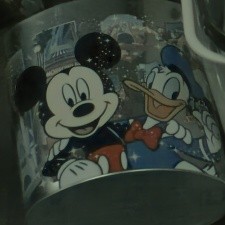}
    \includegraphics[width=0.48\textwidth]{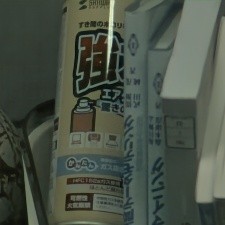}
\end{subfigure}
\begin{subfigure}[t]{0.192\textwidth}
\centering
    \includegraphics[width=0.48\textwidth]{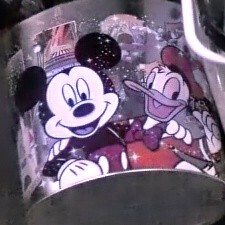}
    \includegraphics[width=0.48\textwidth]{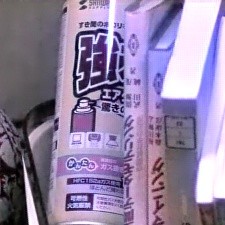}
\end{subfigure}
\\[3ex]

\begin{subfigure}[t]{0.8\textwidth}
\centering
    \includegraphics[width=0.24\textwidth]{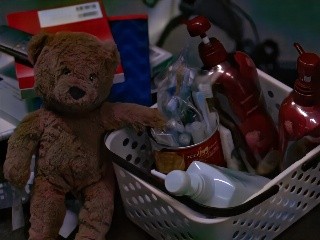}
    \includegraphics[width=0.24\textwidth]{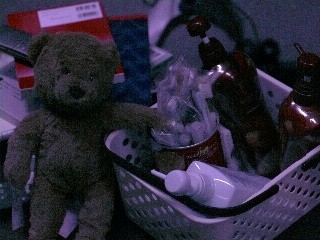}
    \includegraphics[width=0.24\textwidth]{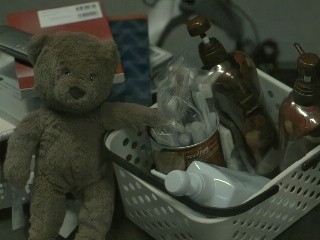}
    \includegraphics[width=0.24\textwidth]{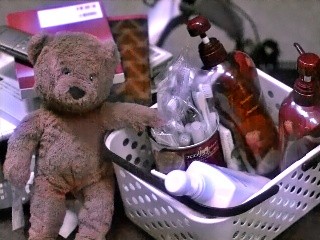}
\end{subfigure}
\\[1ex]
\begin{subfigure}[t]{0.192\textwidth}
\centering
    \includegraphics[width=0.48\textwidth]{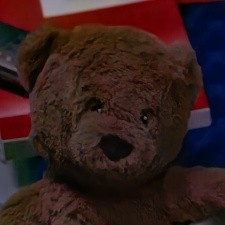}
    \includegraphics[width=0.48\textwidth]{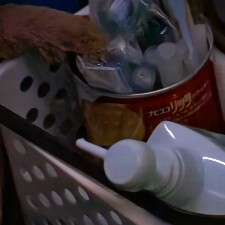}
\end{subfigure}
\begin{subfigure}[t]{0.192\textwidth}
\centering
    \includegraphics[width=0.48\textwidth]{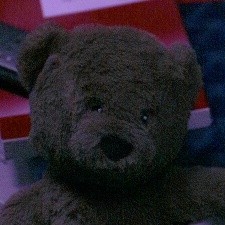}
    \includegraphics[width=0.48\textwidth]{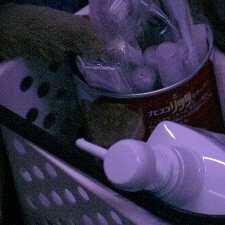}
\end{subfigure}
\begin{subfigure}[t]{0.192\textwidth}
\centering
    \includegraphics[width=0.48\textwidth]{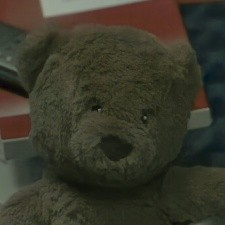}
    \includegraphics[width=0.48\textwidth]{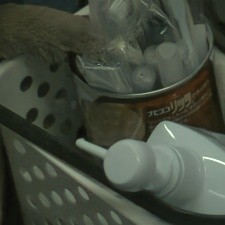}
\end{subfigure}
\begin{subfigure}[t]{0.192\textwidth}
\centering
    \includegraphics[width=0.48\textwidth]{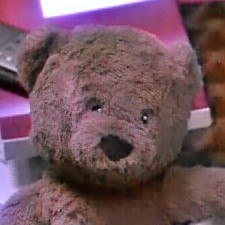}
    \includegraphics[width=0.48\textwidth]{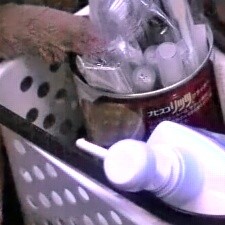}
\end{subfigure}

\begin{subfigure}[t]{0.192\textwidth}
\centering
SID \cite{chen2018learning}
\end{subfigure}
\begin{subfigure}[t]{0.192\textwidth}
\centering
SMID \cite{chen2019seeing}
\end{subfigure}
\begin{subfigure}[t]{0.192\textwidth}
\centering
ELD \cite{wei2020physics}
\end{subfigure}
\begin{subfigure}[t]{0.192\textwidth}
\centering
RED \cite{lamba2021restoring}
\end{subfigure}
\\[3ex]

\begin{subfigure}[t]{0.8\textwidth}
\centering
    \includegraphics[width=0.24\textwidth]{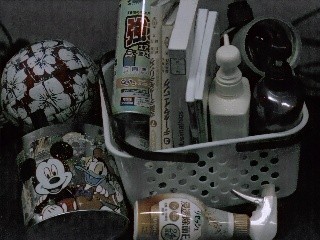}
    \includegraphics[width=0.24\textwidth]{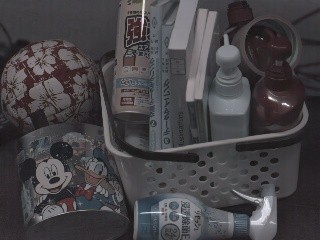}
    \includegraphics[width=0.24\textwidth]{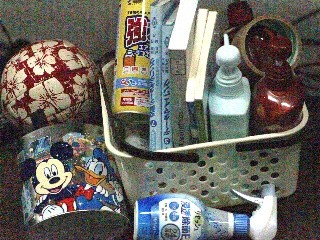}
    \includegraphics[width=0.24\textwidth]{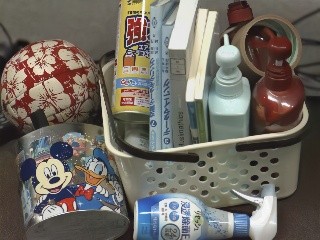}
\end{subfigure}
\\[1ex]
\begin{subfigure}[t]{0.192\textwidth}
\centering
    \includegraphics[width=0.48\textwidth]{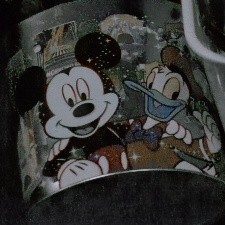}
    \includegraphics[width=0.48\textwidth]{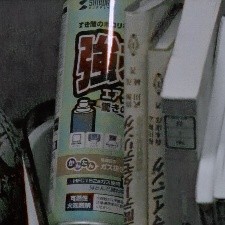}
\end{subfigure}
\begin{subfigure}[t]{0.192\textwidth}
\centering
    \includegraphics[width=0.48\textwidth]{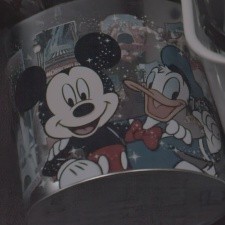}
    \includegraphics[width=0.48\textwidth]{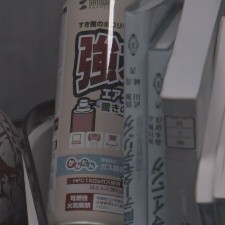}
\end{subfigure}
\begin{subfigure}[t]{0.192\textwidth}
\centering
    \includegraphics[width=0.48\textwidth]{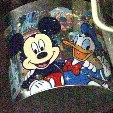}
    \includegraphics[width=0.48\textwidth]{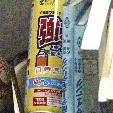}
\end{subfigure}
\begin{subfigure}[t]{0.192\textwidth}
\centering
    \includegraphics[width=0.48\textwidth]{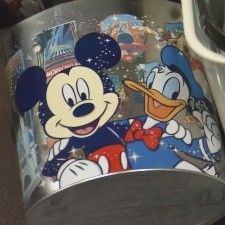}
    \includegraphics[width=0.48\textwidth]{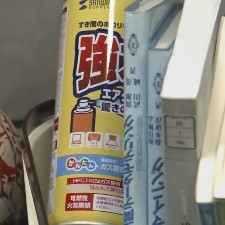}
\end{subfigure}
\\[3ex]

\begin{subfigure}[t]{0.8\textwidth}
\centering
    \includegraphics[width=0.24\textwidth]{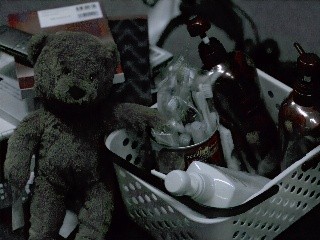}
    \includegraphics[width=0.24\textwidth]{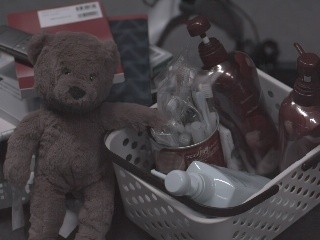}
    \includegraphics[width=0.24\textwidth]{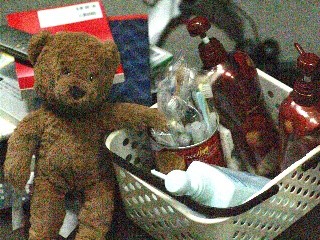}
    \includegraphics[width=0.24\textwidth]{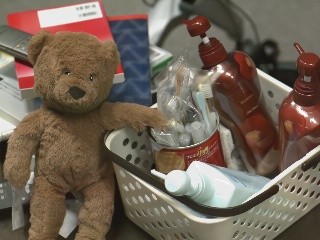}
\end{subfigure}
\\[1ex]
\begin{subfigure}[t]{0.192\textwidth}
\centering
    \includegraphics[width=0.48\textwidth]{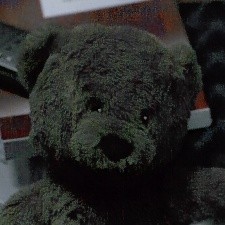}
    \includegraphics[width=0.48\textwidth]{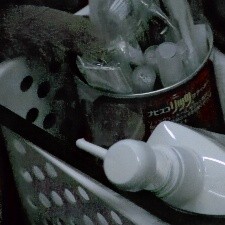}
\end{subfigure}
\begin{subfigure}[t]{0.192\textwidth}
\centering
    \includegraphics[width=0.48\textwidth]{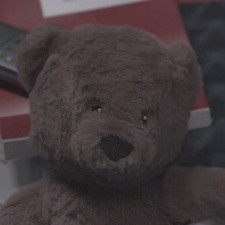}
    \includegraphics[width=0.48\textwidth]{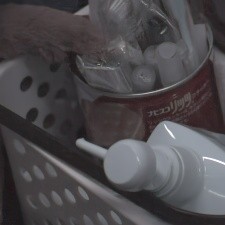}
\end{subfigure}
\begin{subfigure}[t]{0.192\textwidth}
\centering
    \includegraphics[width=0.48\textwidth]{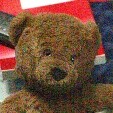}
    \includegraphics[width=0.48\textwidth]{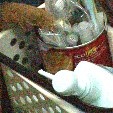}
\end{subfigure}
\begin{subfigure}[t]{0.192\textwidth}
\centering
    \includegraphics[width=0.48\textwidth]{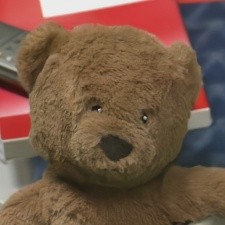}
    \includegraphics[width=0.48\textwidth]{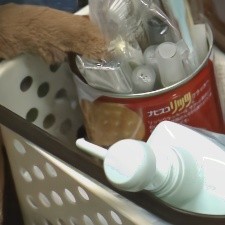}
\end{subfigure}

\begin{subfigure}[t]{0.192\textwidth}
\centering
ABFS \cite{dong2022abandoning}
\end{subfigure}
\begin{subfigure}[t]{0.192\textwidth}
\centering
RNS \cite{zhang2021rethinking}
\end{subfigure}
\begin{subfigure}[t]{0.192\textwidth}
\centering
Camera JPG
\end{subfigure}
\begin{subfigure}[t]{0.192\textwidth}
\centering
Ours
\end{subfigure}

\begin{subfigure}[t]{1\textwidth}
\centering
\caption{Denoising results for images with $1/_{1k} (s)$ shutter speed.}
\end{subfigure}
\end{figure*}

\begin{figure*}[t]\ContinuedFloat
\centering
\begin{subfigure}[t]{0.8\textwidth}
\centering
    \includegraphics[width=0.24\textwidth]{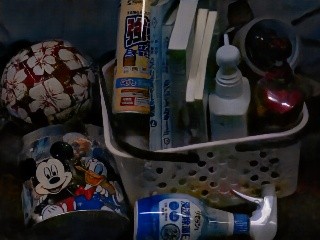}
    \includegraphics[width=0.24\textwidth]{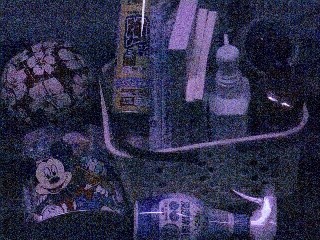}
    \includegraphics[width=0.24\textwidth]{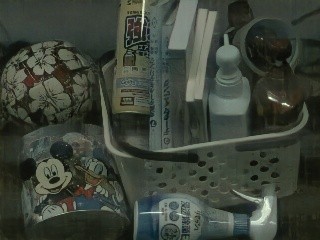}
    \includegraphics[width=0.24\textwidth]{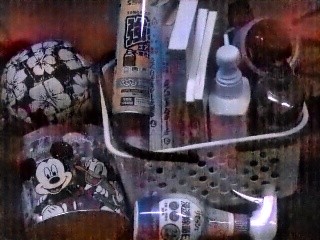}
\end{subfigure}
\\[1ex]
\begin{subfigure}[t]{0.192\textwidth}
\centering
    \includegraphics[width=0.48\textwidth]{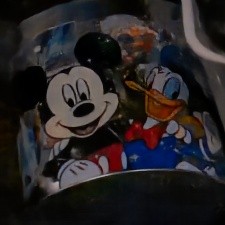}
    \includegraphics[width=0.48\textwidth]{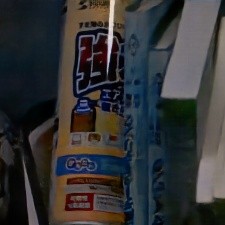}
\end{subfigure}
\begin{subfigure}[t]{0.192\textwidth}
\centering
    \includegraphics[width=0.48\textwidth]{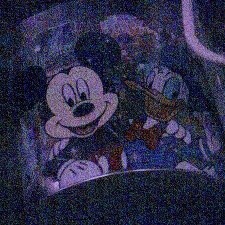}
    \includegraphics[width=0.48\textwidth]{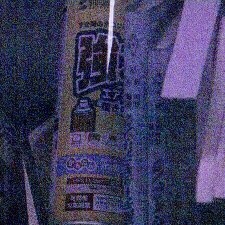}
\end{subfigure}
\begin{subfigure}[t]{0.192\textwidth}
\centering
    \includegraphics[width=0.48\textwidth]{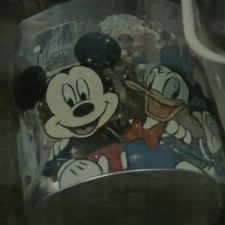}
    \includegraphics[width=0.48\textwidth]{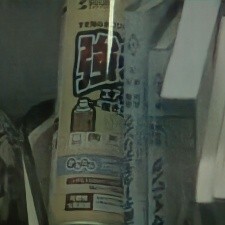}
\end{subfigure}
\begin{subfigure}[t]{0.192\textwidth}
\centering
    \includegraphics[width=0.48\textwidth]{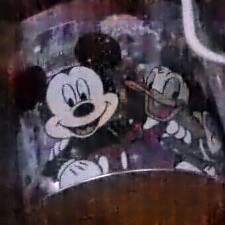}
    \includegraphics[width=0.48\textwidth]{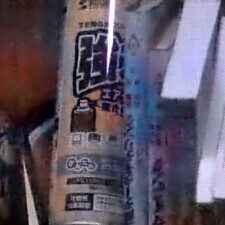}
\end{subfigure}
\\[3ex]

\begin{subfigure}[t]{0.8\textwidth}
\centering
    \includegraphics[width=0.24\textwidth]{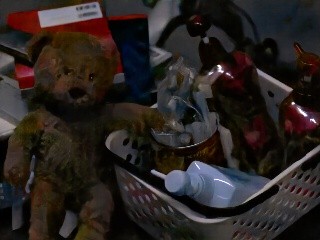}
    \includegraphics[width=0.24\textwidth]{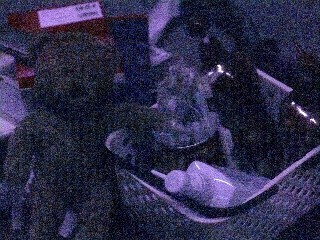}
    \includegraphics[width=0.24\textwidth]{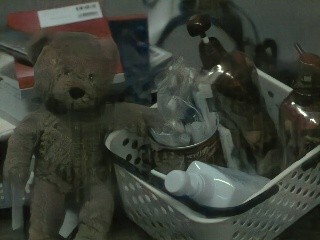}
    \includegraphics[width=0.24\textwidth]{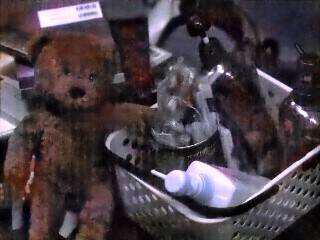}
\end{subfigure}
\\[1ex]
\begin{subfigure}[t]{0.192\textwidth}
\centering
    \includegraphics[width=0.48\textwidth]{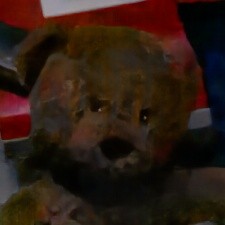}
    \includegraphics[width=0.48\textwidth]{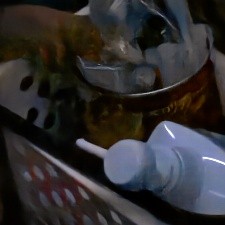}
\end{subfigure}
\begin{subfigure}[t]{0.192\textwidth}
\centering
    \includegraphics[width=0.48\textwidth]{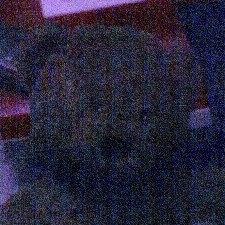}
    \includegraphics[width=0.48\textwidth]{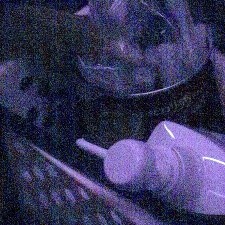}
\end{subfigure}
\begin{subfigure}[t]{0.192\textwidth}
\centering
    \includegraphics[width=0.48\textwidth]{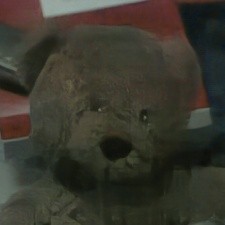}
    \includegraphics[width=0.48\textwidth]{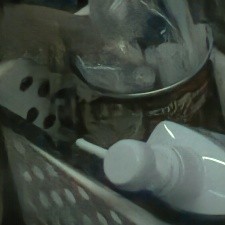}
\end{subfigure}
\begin{subfigure}[t]{0.192\textwidth}
\centering
    \includegraphics[width=0.48\textwidth]{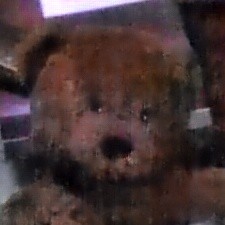}
    \includegraphics[width=0.48\textwidth]{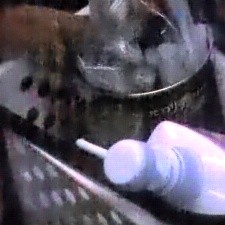}
\end{subfigure}

\begin{subfigure}[t]{0.192\textwidth}
\centering
SID \cite{chen2018learning}
\end{subfigure}
\begin{subfigure}[t]{0.192\textwidth}
\centering
SMID \cite{chen2019seeing}
\end{subfigure}
\begin{subfigure}[t]{0.192\textwidth}
\centering
ELD \cite{wei2020physics}
\end{subfigure}
\begin{subfigure}[t]{0.192\textwidth}
\centering
RED \cite{lamba2021restoring}
\end{subfigure}
\\[3ex]

\begin{subfigure}[t]{0.8\textwidth}
\centering
    \includegraphics[width=0.24\textwidth]{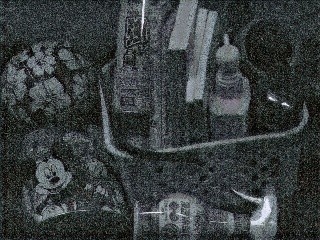}
    \includegraphics[width=0.24\textwidth]{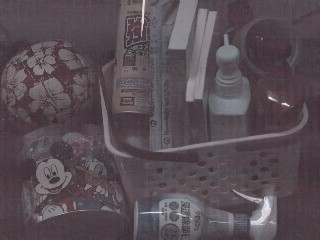}
    \includegraphics[width=0.24\textwidth]{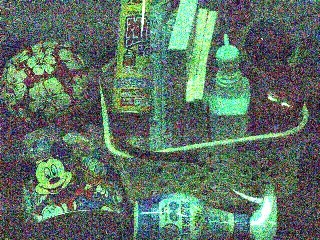}
    \includegraphics[width=0.24\textwidth]{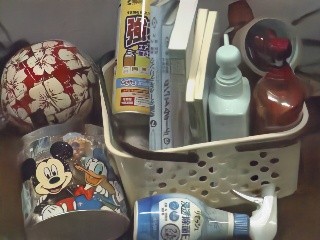}
\end{subfigure}
\\[1ex]
\begin{subfigure}[t]{0.192\textwidth}
\centering
    \includegraphics[width=0.48\textwidth]{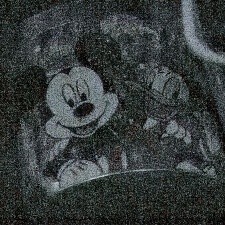}
    \includegraphics[width=0.48\textwidth]{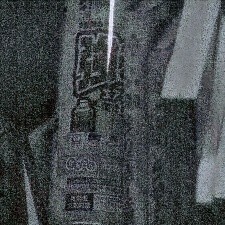}
\end{subfigure}
\begin{subfigure}[t]{0.192\textwidth}
\centering
    \includegraphics[width=0.48\textwidth]{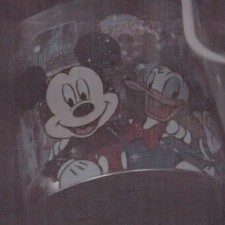}
    \includegraphics[width=0.48\textwidth]{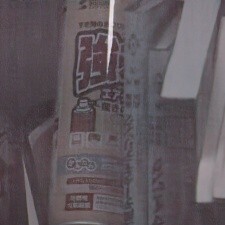}
\end{subfigure}
\begin{subfigure}[t]{0.192\textwidth}
\centering
    \includegraphics[width=0.48\textwidth]{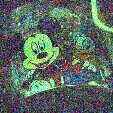}
    \includegraphics[width=0.48\textwidth]{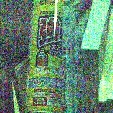}
\end{subfigure}
\begin{subfigure}[t]{0.192\textwidth}
\centering
    \includegraphics[width=0.48\textwidth]{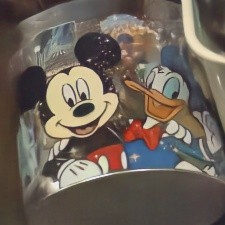}
    \includegraphics[width=0.48\textwidth]{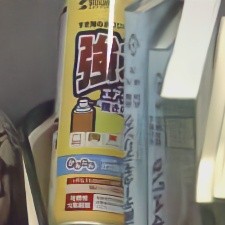}
\end{subfigure}
\\[3ex]

\begin{subfigure}[t]{0.8\textwidth}
\centering
    \includegraphics[width=0.24\textwidth]{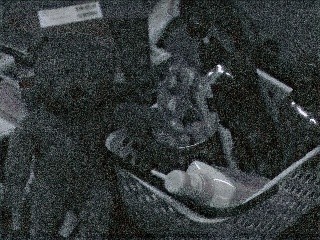}
    \includegraphics[width=0.24\textwidth]{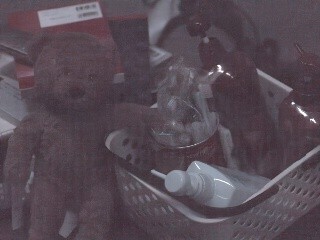}
    \includegraphics[width=0.24\textwidth]{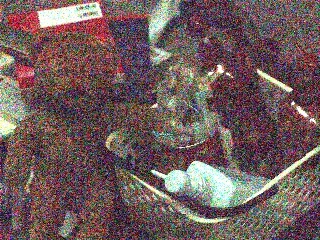}
    \includegraphics[width=0.24\textwidth]{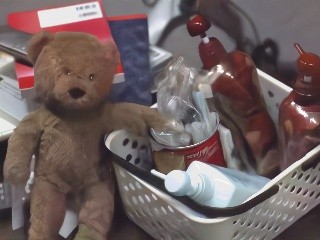}
\end{subfigure}
\\[1ex]
\begin{subfigure}[t]{0.192\textwidth}
\centering
    \includegraphics[width=0.48\textwidth]{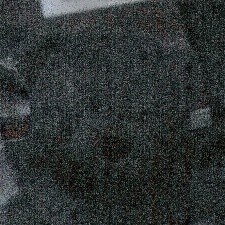}
    \includegraphics[width=0.48\textwidth]{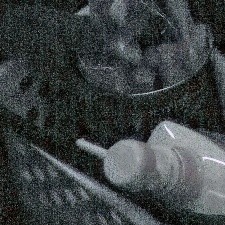}
\end{subfigure}
\begin{subfigure}[t]{0.192\textwidth}
\centering
    \includegraphics[width=0.48\textwidth]{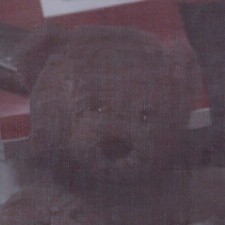}
    \includegraphics[width=0.48\textwidth]{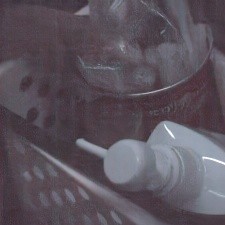}
\end{subfigure}
\begin{subfigure}[t]{0.192\textwidth}
\centering
    \includegraphics[width=0.48\textwidth]{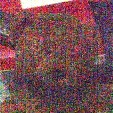}
    \includegraphics[width=0.48\textwidth]{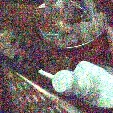}
\end{subfigure}
\begin{subfigure}[t]{0.192\textwidth}
\centering
    \includegraphics[width=0.48\textwidth]{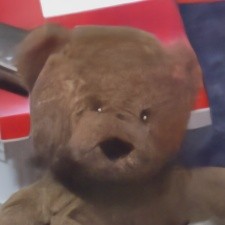}
    \includegraphics[width=0.48\textwidth]{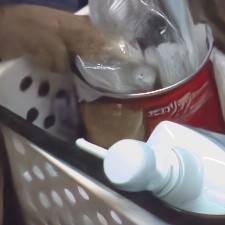}
\end{subfigure}

\begin{subfigure}[t]{0.192\textwidth}
\centering
ABFS \cite{dong2022abandoning}
\end{subfigure}
\begin{subfigure}[t]{0.192\textwidth}
\centering
RNS \cite{zhang2021rethinking}
\end{subfigure}
\begin{subfigure}[t]{0.192\textwidth}
\centering
Camera JPG
\end{subfigure}
\begin{subfigure}[t]{0.192\textwidth}
\centering
Ours
\end{subfigure}

\begin{subfigure}[t]{1\textwidth}
\centering
\caption{Denoising results for images with $1/_{5k} (s)$ shutter speed.}
\end{subfigure}
\end{figure*}

\begin{figure*}[t]\ContinuedFloat
\centering
\begin{subfigure}[t]{0.8\textwidth}
\centering
    \includegraphics[width=0.24\textwidth]{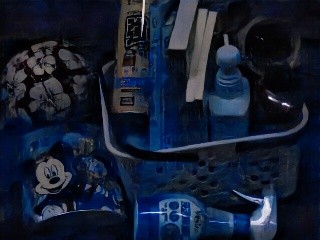}
    \includegraphics[width=0.24\textwidth]{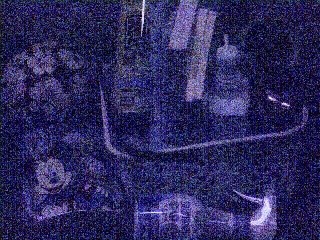}
    \includegraphics[width=0.24\textwidth]{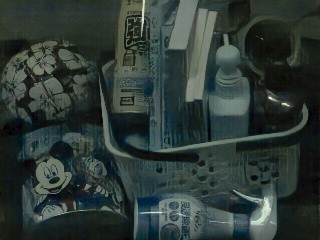}
    \includegraphics[width=0.24\textwidth]{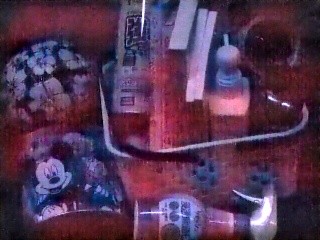}
\end{subfigure}
\\[1ex]
\begin{subfigure}[t]{0.192\textwidth}
\centering
    \includegraphics[width=0.48\textwidth]{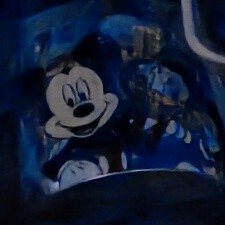}
    \includegraphics[width=0.48\textwidth]{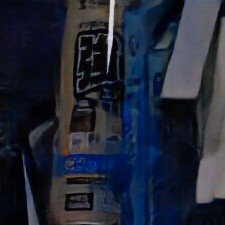}
\end{subfigure}
\begin{subfigure}[t]{0.192\textwidth}
\centering
    \includegraphics[width=0.48\textwidth]{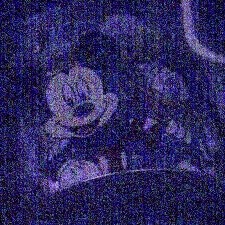}
    \includegraphics[width=0.48\textwidth]{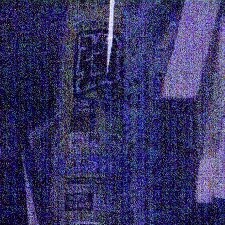}
\end{subfigure}
\begin{subfigure}[t]{0.192\textwidth}
\centering
    \includegraphics[width=0.48\textwidth]{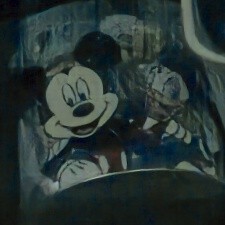}
    \includegraphics[width=0.48\textwidth]{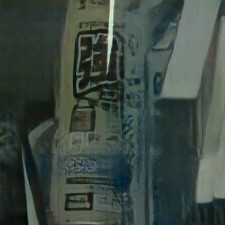}
\end{subfigure}
\begin{subfigure}[t]{0.192\textwidth}
\centering
    \includegraphics[width=0.48\textwidth]{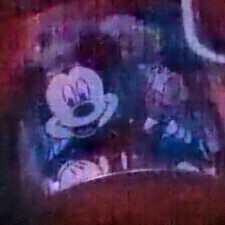}
    \includegraphics[width=0.48\textwidth]{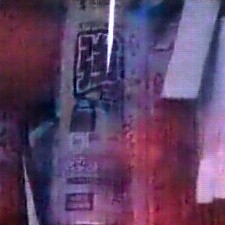}
\end{subfigure}
\\[3ex]

\begin{subfigure}[t]{0.8\textwidth}
\centering
    \includegraphics[width=0.24\textwidth]{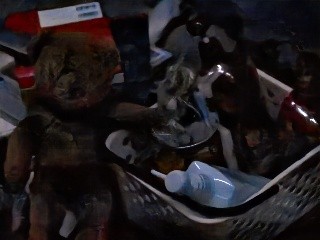}
    \includegraphics[width=0.24\textwidth]{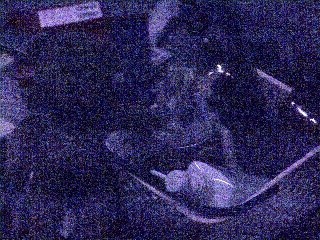}
    \includegraphics[width=0.24\textwidth]{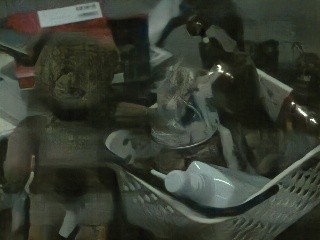}
    \includegraphics[width=0.24\textwidth]{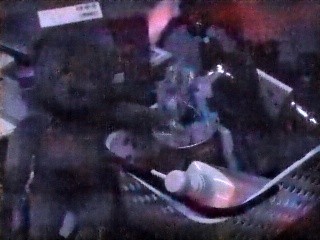}
\end{subfigure}
\\[1ex]
\begin{subfigure}[t]{0.192\textwidth}
\centering
    \includegraphics[width=0.48\textwidth]{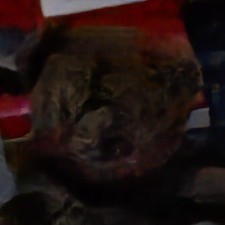}
    \includegraphics[width=0.48\textwidth]{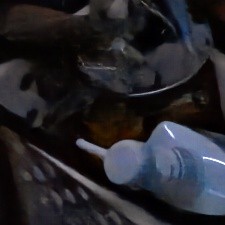}
\end{subfigure}
\begin{subfigure}[t]{0.192\textwidth}
\centering
    \includegraphics[width=0.48\textwidth]{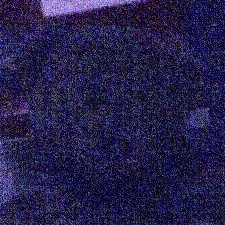}
    \includegraphics[width=0.48\textwidth]{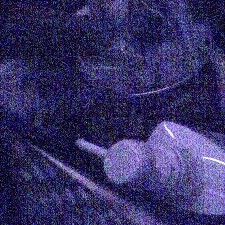}
\end{subfigure}
\begin{subfigure}[t]{0.192\textwidth}
\centering
    \includegraphics[width=0.48\textwidth]{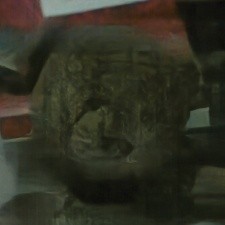}
    \includegraphics[width=0.48\textwidth]{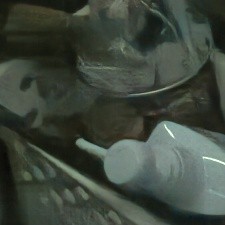}
\end{subfigure}
\begin{subfigure}[t]{0.192\textwidth}
\centering
    \includegraphics[width=0.48\textwidth]{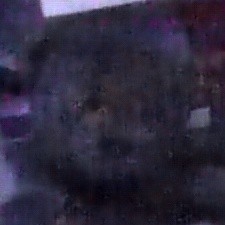}
    \includegraphics[width=0.48\textwidth]{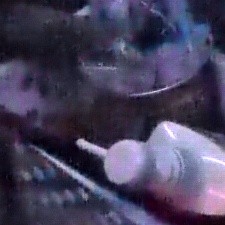}
\end{subfigure}

\begin{subfigure}[t]{0.192\textwidth}
\centering
SID \cite{chen2018learning}
\end{subfigure}
\begin{subfigure}[t]{0.192\textwidth}
\centering
SMID \cite{chen2019seeing}
\end{subfigure}
\begin{subfigure}[t]{0.192\textwidth}
\centering
ELD \cite{wei2020physics}
\end{subfigure}
\begin{subfigure}[t]{0.192\textwidth}
\centering
RED \cite{lamba2021restoring}
\end{subfigure}
\\[3ex]

\begin{subfigure}[t]{0.8\textwidth}
\centering
    \includegraphics[width=0.24\textwidth]{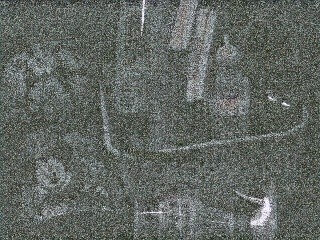}
    \includegraphics[width=0.24\textwidth]{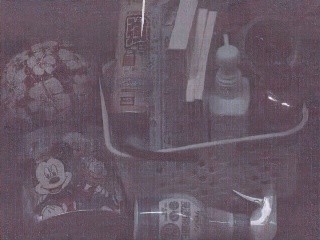}
    \includegraphics[width=0.24\textwidth]{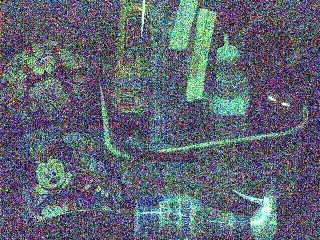}
    \includegraphics[width=0.24\textwidth]{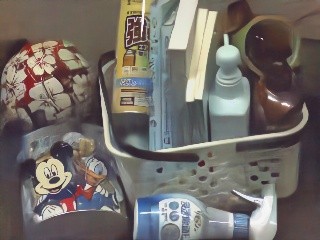}
\end{subfigure}
\\[1ex]
\begin{subfigure}[t]{0.192\textwidth}
\centering
    \includegraphics[width=0.48\textwidth]{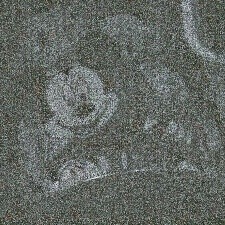}
    \includegraphics[width=0.48\textwidth]{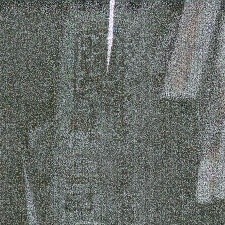}
\end{subfigure}
\begin{subfigure}[t]{0.192\textwidth}
\centering
    \includegraphics[width=0.48\textwidth]{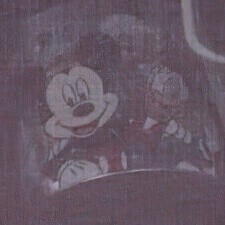}
    \includegraphics[width=0.48\textwidth]{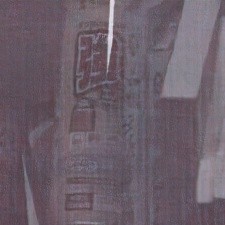}
\end{subfigure}
\begin{subfigure}[t]{0.192\textwidth}
\centering
    \includegraphics[width=0.48\textwidth]{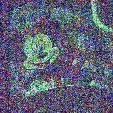}
    \includegraphics[width=0.48\textwidth]{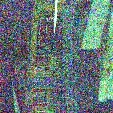}
\end{subfigure}
\begin{subfigure}[t]{0.192\textwidth}
\centering
    \includegraphics[width=0.48\textwidth]{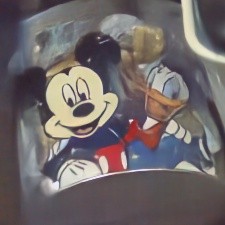}
    \includegraphics[width=0.48\textwidth]{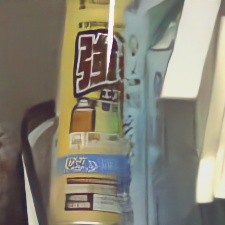}
\end{subfigure}
\\[3ex]

\begin{subfigure}[t]{0.8\textwidth}
\centering
    \includegraphics[width=0.24\textwidth]{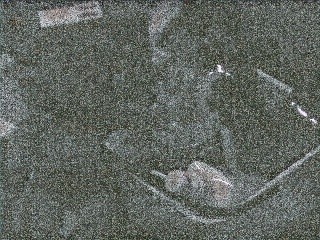}
    \includegraphics[width=0.24\textwidth]{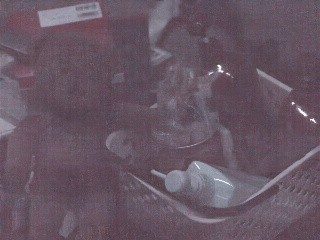}
    \includegraphics[width=0.24\textwidth]{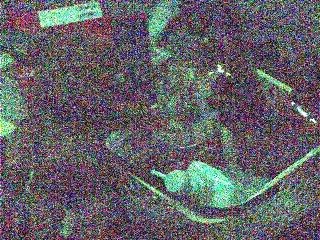}
    \includegraphics[width=0.24\textwidth]{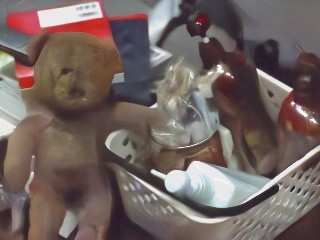}
\end{subfigure}
\\[1ex]
\begin{subfigure}[t]{0.192\textwidth}
\centering
    \includegraphics[width=0.48\textwidth]{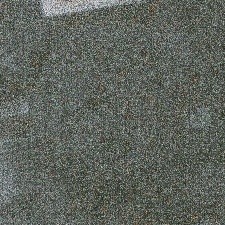}
    \includegraphics[width=0.48\textwidth]{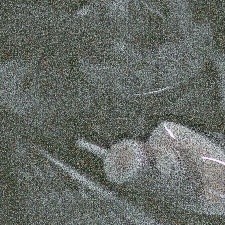}
\end{subfigure}
\begin{subfigure}[t]{0.192\textwidth}
\centering
    \includegraphics[width=0.48\textwidth]{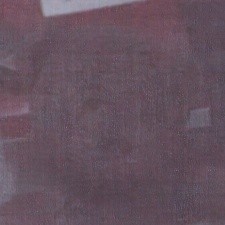}
    \includegraphics[width=0.48\textwidth]{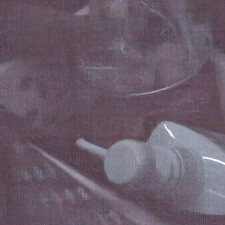}
\end{subfigure}
\begin{subfigure}[t]{0.192\textwidth}
\centering
    \includegraphics[width=0.48\textwidth]{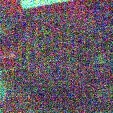}
    \includegraphics[width=0.48\textwidth]{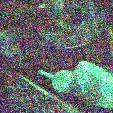}
\end{subfigure}
\begin{subfigure}[t]{0.192\textwidth}
\centering
    \includegraphics[width=0.48\textwidth]{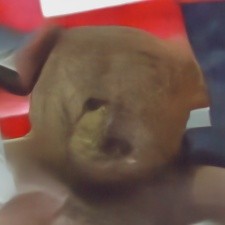}
    \includegraphics[width=0.48\textwidth]{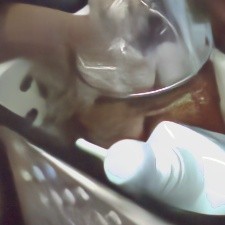}
\end{subfigure}

\begin{subfigure}[t]{0.192\textwidth}
\centering
ABFS \cite{dong2022abandoning}
\end{subfigure}
\begin{subfigure}[t]{0.192\textwidth}
\centering
RNS \cite{zhang2021rethinking}
\end{subfigure}
\begin{subfigure}[t]{0.192\textwidth}
\centering
Camera JPG
\end{subfigure}
\begin{subfigure}[t]{0.192\textwidth}
\centering
Ours
\end{subfigure}

\begin{subfigure}[t]{1\textwidth}
\centering
\caption{Denoising results for images with $1/_{10k} (s)$ shutter speed.}
\end{subfigure}
\caption{Qualitative experimental results with existing models.}
\label{fig_existing}
\end{figure*}

\begin{figure*}[t]
\centering
\begin{subfigure}[t]{1\textwidth}
\centering
    \includegraphics[width=0.24\textwidth]{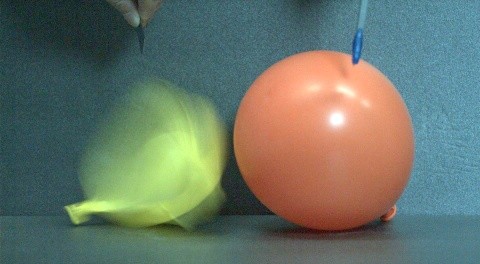}
    \includegraphics[width=0.24\textwidth]{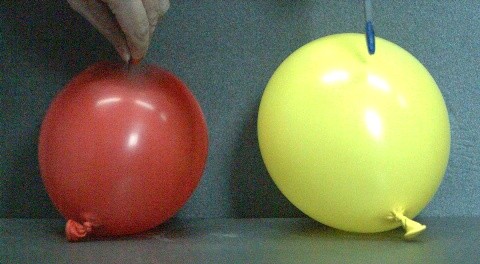}
    \includegraphics[width=0.24\textwidth]{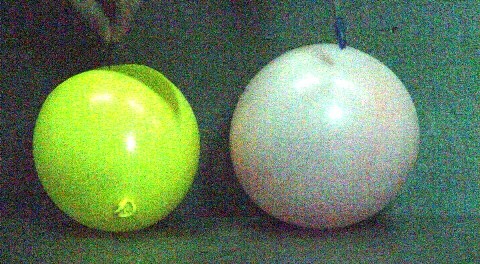}
    \includegraphics[width=0.24\textwidth]{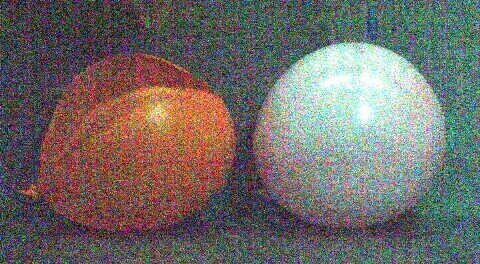}
\end{subfigure}
\\[1ex]
\begin{subfigure}[t]{0.24\textwidth}
\centering
    \includegraphics[width=0.49\textwidth]{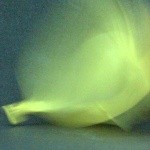}
    \includegraphics[width=0.49\textwidth]{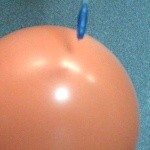}
\end{subfigure}
\begin{subfigure}[t]{0.24\textwidth}
\centering
    \includegraphics[width=0.49\textwidth]{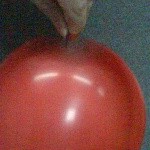}
    \includegraphics[width=0.49\textwidth]{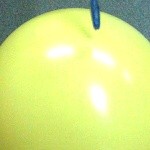}
\end{subfigure}
\begin{subfigure}[t]{0.24\textwidth}
\centering
    \includegraphics[width=0.49\textwidth]{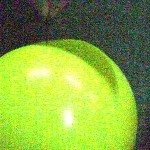}
    \includegraphics[width=0.49\textwidth]{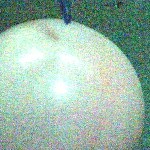}
\end{subfigure}
\begin{subfigure}[t]{0.24\textwidth}
\centering
    \includegraphics[width=0.49\textwidth]{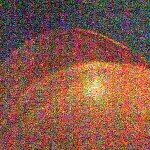}
    \includegraphics[width=0.49\textwidth]{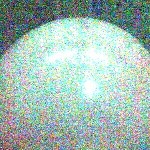}
\end{subfigure}
\\[3ex]

\begin{subfigure}[t]{1\textwidth}
\centering
    \includegraphics[width=0.24\textwidth]{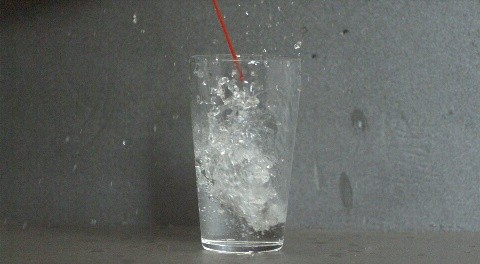}
    \includegraphics[width=0.24\textwidth]{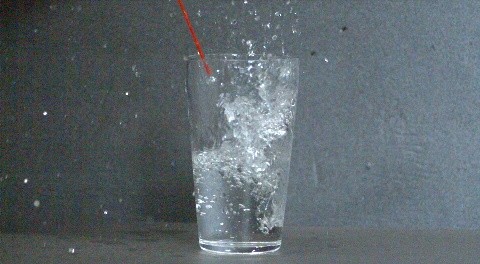}
    \includegraphics[width=0.24\textwidth]{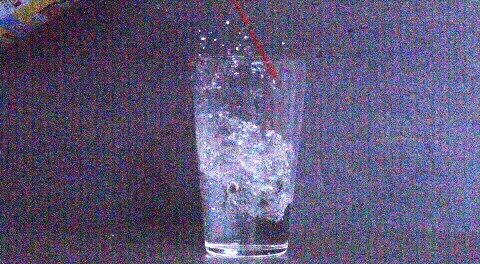}
    \includegraphics[width=0.24\textwidth]{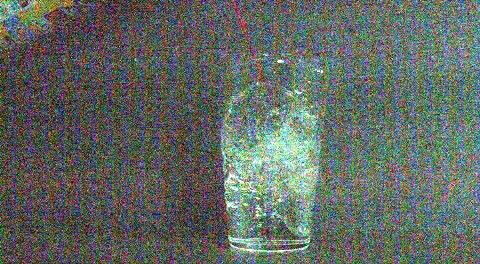}
\end{subfigure}
\\[1ex]
\begin{subfigure}[t]{0.24\textwidth}
\centering
    \includegraphics[width=0.49\textwidth]{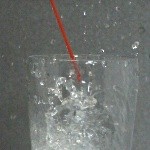}
    \includegraphics[width=0.49\textwidth]{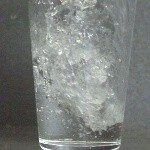}
\end{subfigure}
\begin{subfigure}[t]{0.24\textwidth}
\centering
    \includegraphics[width=0.49\textwidth]{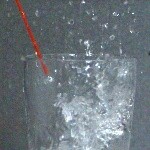}
    \includegraphics[width=0.49\textwidth]{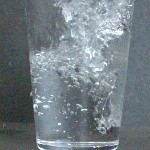}
\end{subfigure}
\begin{subfigure}[t]{0.24\textwidth}
\centering
    \includegraphics[width=0.49\textwidth]{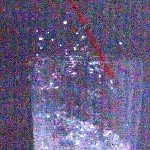}
    \includegraphics[width=0.49\textwidth]{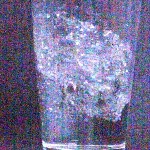}
\end{subfigure}
\begin{subfigure}[t]{0.24\textwidth}
\centering
    \includegraphics[width=0.49\textwidth]{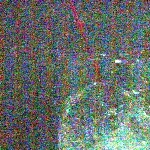}
    \includegraphics[width=0.49\textwidth]{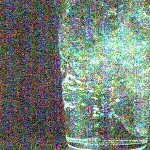}
\end{subfigure}
\\[3ex]

\begin{subfigure}[t]{1\textwidth}
\centering
    \includegraphics[width=0.24\textwidth]{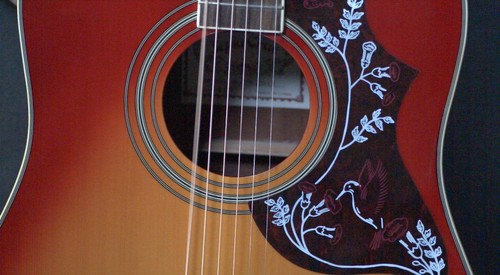}
    \includegraphics[width=0.24\textwidth]{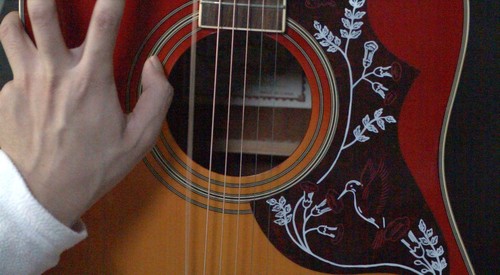}
    \includegraphics[width=0.24\textwidth]{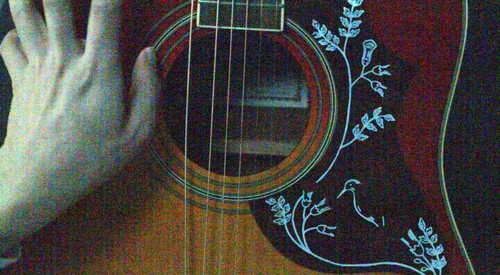}
    \includegraphics[width=0.24\textwidth]{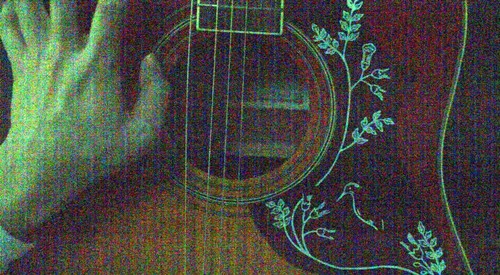}
\end{subfigure}
\\[1ex]
\begin{subfigure}[t]{0.24\textwidth}
\centering
    \includegraphics[width=0.49\textwidth]{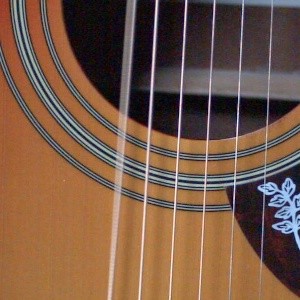}
    \includegraphics[width=0.49\textwidth]{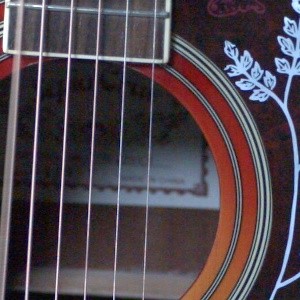}
\end{subfigure}
\begin{subfigure}[t]{0.24\textwidth}
\centering
    \includegraphics[width=0.49\textwidth]{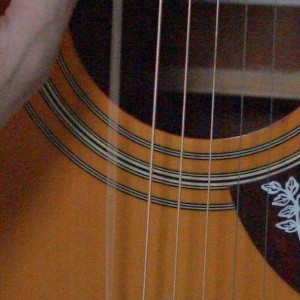}
    \includegraphics[width=0.49\textwidth]{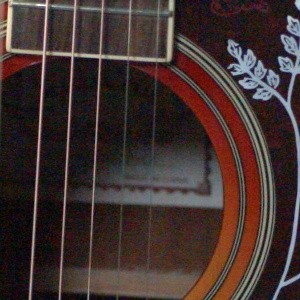}
\end{subfigure}
\begin{subfigure}[t]{0.24\textwidth}
\centering
    \includegraphics[width=0.49\textwidth]{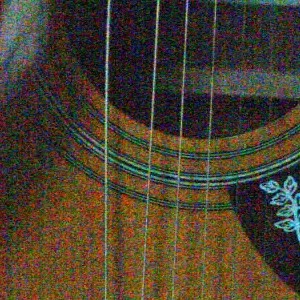}
    \includegraphics[width=0.49\textwidth]{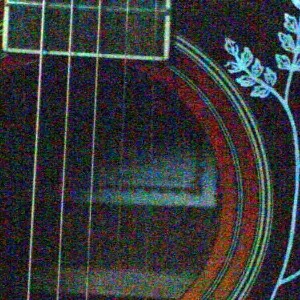}
\end{subfigure}
\begin{subfigure}[t]{0.24\textwidth}
\centering
    \includegraphics[width=0.49\textwidth]{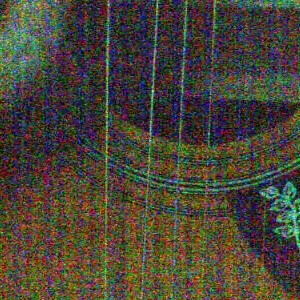}
    \includegraphics[width=0.49\textwidth]{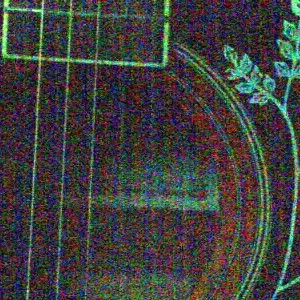}
\end{subfigure}
\\[3ex]

\begin{subfigure}[t]{1\textwidth}
\centering
    \includegraphics[width=0.24\textwidth]{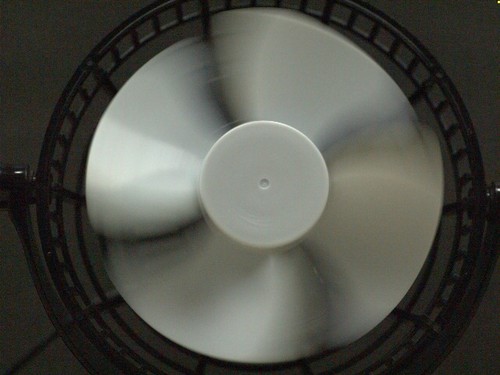}
    \includegraphics[width=0.24\textwidth]{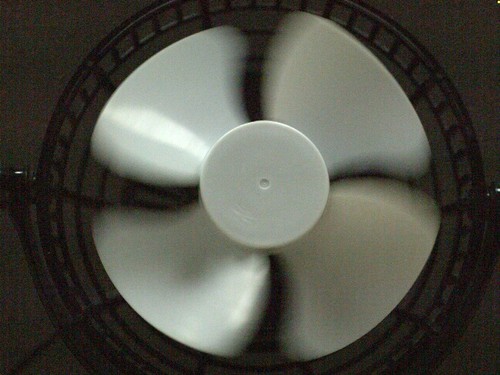}
    \includegraphics[width=0.24\textwidth]{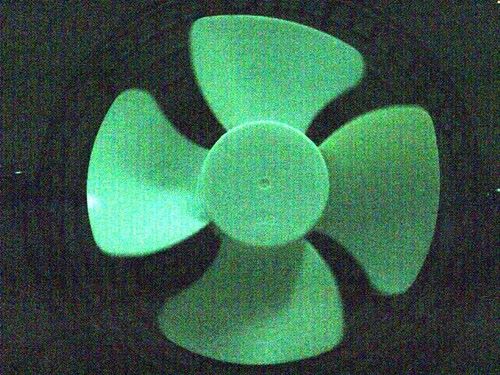}
    \includegraphics[width=0.24\textwidth]{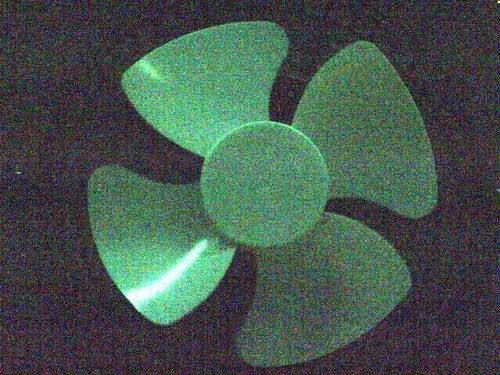}
\end{subfigure}
\\[1ex]
\begin{subfigure}[t]{0.24\textwidth}
\centering
    \includegraphics[width=0.49\textwidth]{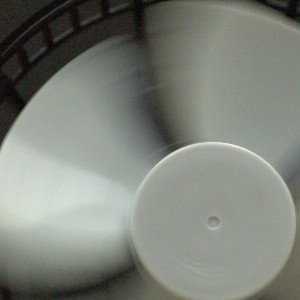}
    \includegraphics[width=0.49\textwidth]{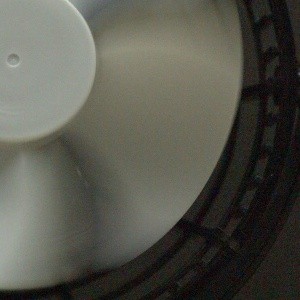}
\end{subfigure}
\begin{subfigure}[t]{0.24\textwidth}
\centering
    \includegraphics[width=0.49\textwidth]{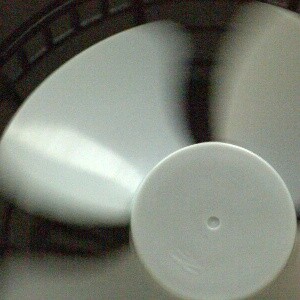}
    \includegraphics[width=0.49\textwidth]{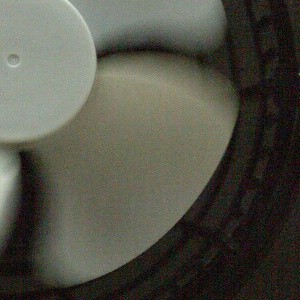}
\end{subfigure}
\begin{subfigure}[t]{0.24\textwidth}
\centering
    \includegraphics[width=0.49\textwidth]{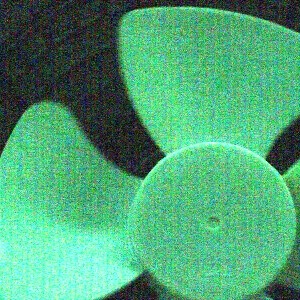}
    \includegraphics[width=0.49\textwidth]{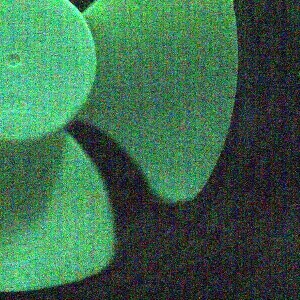}
\end{subfigure}
\begin{subfigure}[t]{0.24\textwidth}
\centering
    \includegraphics[width=0.49\textwidth]{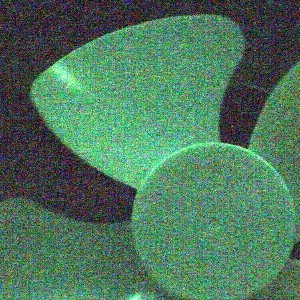}
    \includegraphics[width=0.49\textwidth]{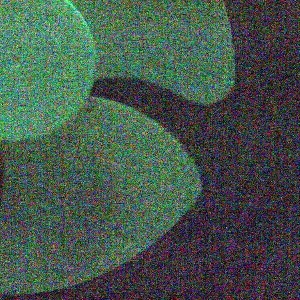}
\end{subfigure}

\begin{subfigure}[t]{0.24\textwidth}
\centering
$1/_{500} (s)$
\end{subfigure}
\begin{subfigure}[t]{0.24\textwidth}
\centering
$1/_{1k} (s)$
\end{subfigure}
\begin{subfigure}[t]{0.24\textwidth}
\centering
$1/_{5k} (s)$
\end{subfigure}
\begin{subfigure}[t]{0.24\textwidth}
\centering
$1/_{10k} (s)$
\end{subfigure}

\begin{subfigure}[t]{0.24\textwidth}
\centering
\caption{Camera JPG with Gain}
\end{subfigure}
\end{figure*}

\begin{figure*}[t]\ContinuedFloat
\centering
\begin{subfigure}[t]{1\textwidth}
\centering
    \includegraphics[width=0.24\textwidth]{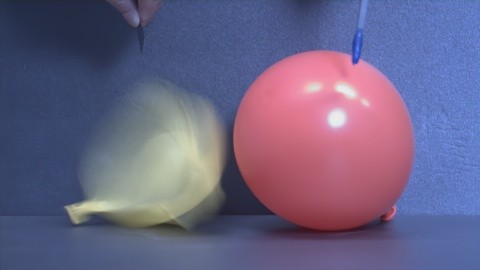}
    \includegraphics[width=0.24\textwidth]{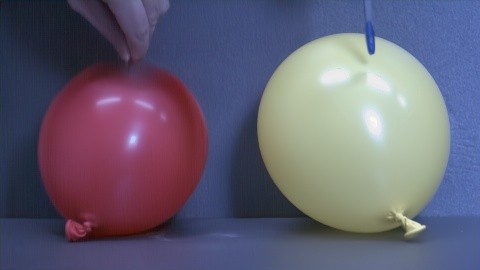}
    \includegraphics[width=0.24\textwidth]{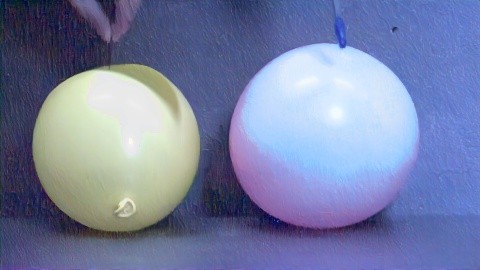}
    \includegraphics[width=0.24\textwidth]{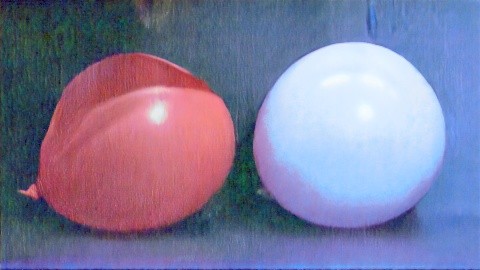}
\end{subfigure}
\\[1ex]
\begin{subfigure}[t]{0.24\textwidth}
\centering
    \includegraphics[width=0.49\textwidth]{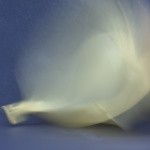}
    \includegraphics[width=0.49\textwidth]{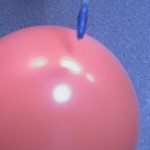}
\end{subfigure}
\begin{subfigure}[t]{0.24\textwidth}
\centering
    \includegraphics[width=0.49\textwidth]{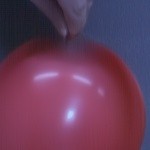}
    \includegraphics[width=0.49\textwidth]{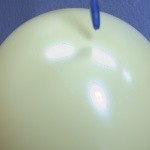}
\end{subfigure}
\begin{subfigure}[t]{0.24\textwidth}
\centering
    \includegraphics[width=0.49\textwidth]{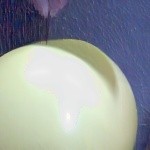}
    \includegraphics[width=0.49\textwidth]{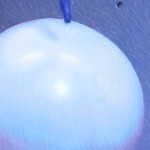}
\end{subfigure}
\begin{subfigure}[t]{0.24\textwidth}
\centering
    \includegraphics[width=0.49\textwidth]{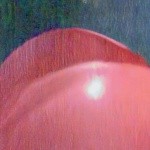}
    \includegraphics[width=0.49\textwidth]{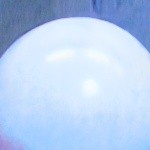}
\end{subfigure}
\\[3ex]

\begin{subfigure}[t]{1\textwidth}
\centering
    \includegraphics[width=0.24\textwidth]{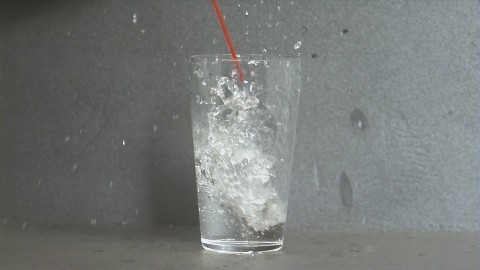}
    \includegraphics[width=0.24\textwidth]{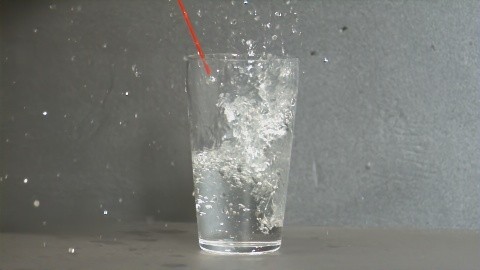}
    \includegraphics[width=0.24\textwidth]{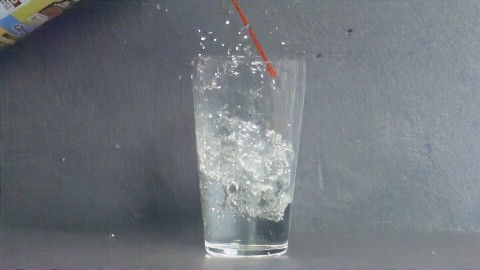}
    \includegraphics[width=0.24\textwidth]{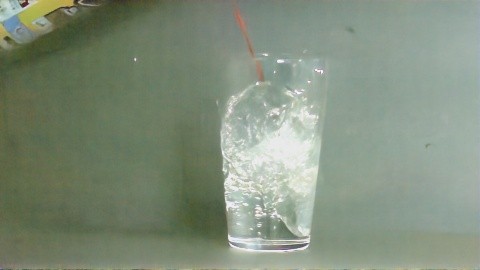}
\end{subfigure}
\\[1ex]
\begin{subfigure}[t]{0.24\textwidth}
\centering
    \includegraphics[width=0.49\textwidth]{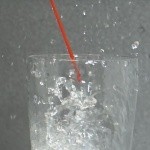}
    \includegraphics[width=0.49\textwidth]{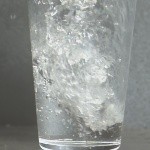}
\end{subfigure}
\begin{subfigure}[t]{0.24\textwidth}
\centering
    \includegraphics[width=0.49\textwidth]{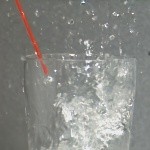}
    \includegraphics[width=0.49\textwidth]{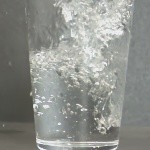}
\end{subfigure}
\begin{subfigure}[t]{0.24\textwidth}
\centering
    \includegraphics[width=0.49\textwidth]{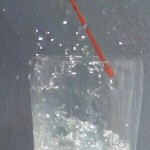}
    \includegraphics[width=0.49\textwidth]{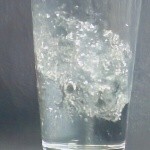}
\end{subfigure}
\begin{subfigure}[t]{0.24\textwidth}
\centering
    \includegraphics[width=0.49\textwidth]{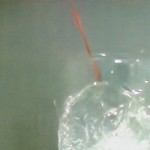}
    \includegraphics[width=0.49\textwidth]{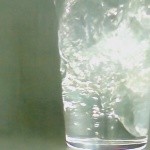}
\end{subfigure}
\\[3ex]

\begin{subfigure}[t]{1\textwidth}
\centering
    \includegraphics[width=0.24\textwidth]{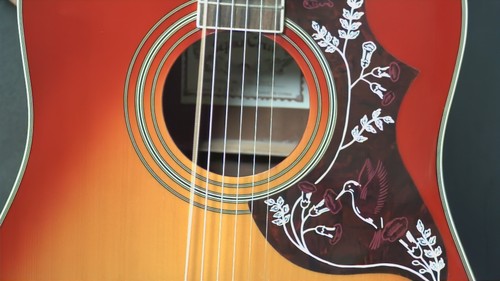}
    \includegraphics[width=0.24\textwidth]{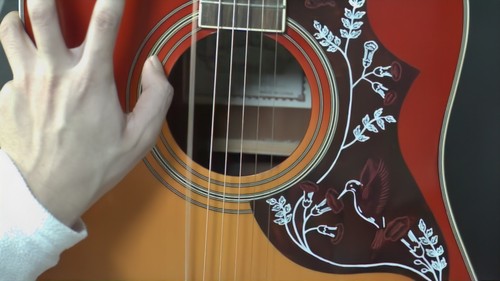}
    \includegraphics[width=0.24\textwidth]{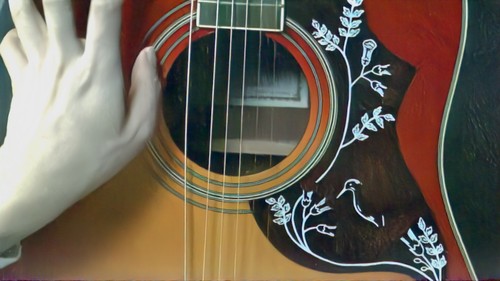}
    \includegraphics[width=0.24\textwidth]{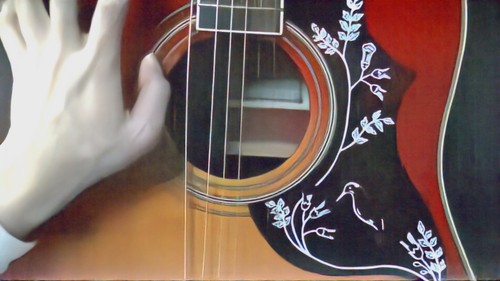}
\end{subfigure}
\\[1ex]
\begin{subfigure}[t]{0.24\textwidth}
\centering
    \includegraphics[width=0.49\textwidth]{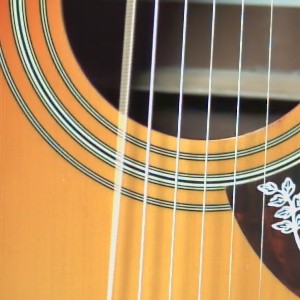}
    \includegraphics[width=0.49\textwidth]{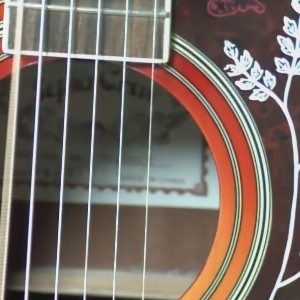}
\end{subfigure}
\begin{subfigure}[t]{0.24\textwidth}
\centering
    \includegraphics[width=0.49\textwidth]{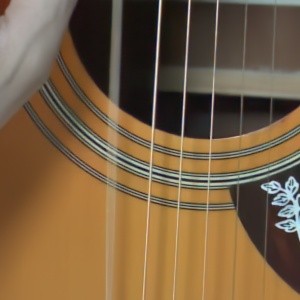}
    \includegraphics[width=0.49\textwidth]{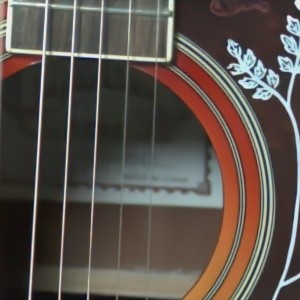}
\end{subfigure}
\begin{subfigure}[t]{0.24\textwidth}
\centering
    \includegraphics[width=0.49\textwidth]{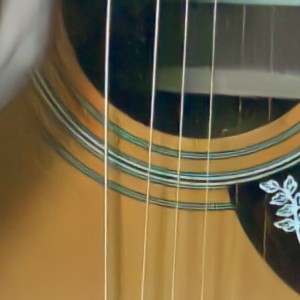}
    \includegraphics[width=0.49\textwidth]{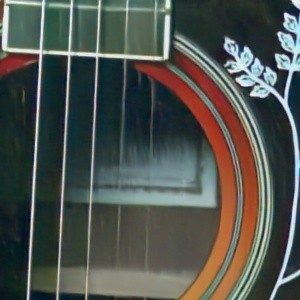}
\end{subfigure}
\begin{subfigure}[t]{0.24\textwidth}
\centering
    \includegraphics[width=0.49\textwidth]{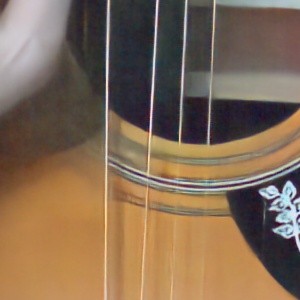}
    \includegraphics[width=0.49\textwidth]{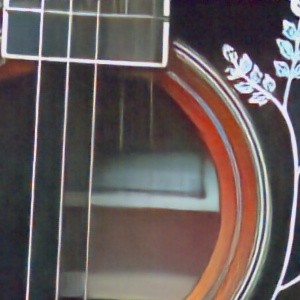}
\end{subfigure}
\\[3ex]

\begin{subfigure}[t]{1\textwidth}
\centering
    \includegraphics[width=0.24\textwidth]{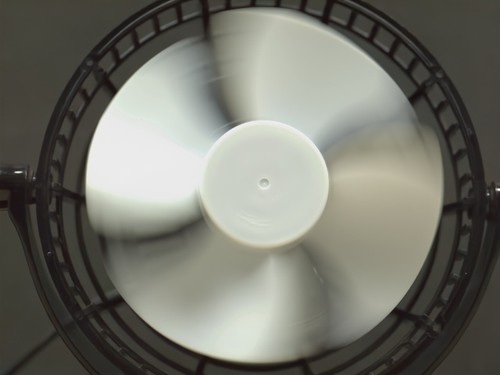}
    \includegraphics[width=0.24\textwidth]{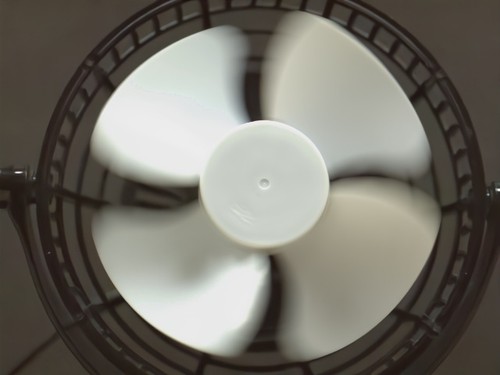}
    \includegraphics[width=0.24\textwidth]{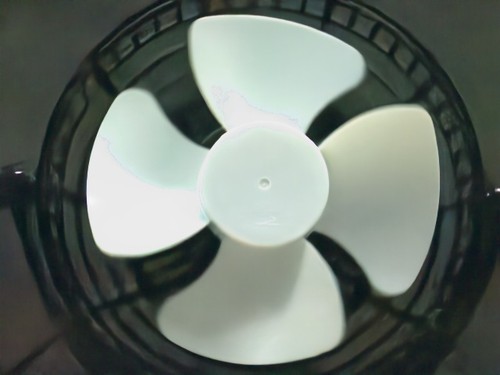}
    \includegraphics[width=0.24\textwidth]{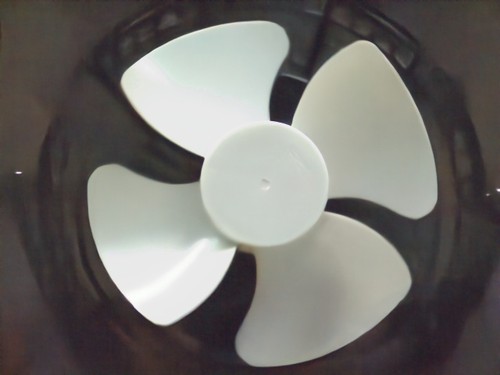}
\end{subfigure}
\\[1ex]
\begin{subfigure}[t]{0.24\textwidth}
\centering
    \includegraphics[width=0.49\textwidth]{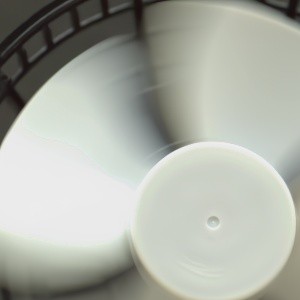}
    \includegraphics[width=0.49\textwidth]{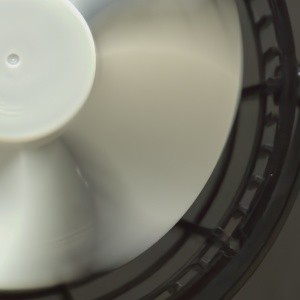}
\end{subfigure}
\begin{subfigure}[t]{0.24\textwidth}
\centering
    \includegraphics[width=0.49\textwidth]{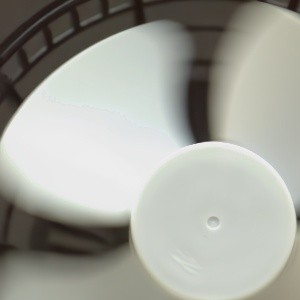}
    \includegraphics[width=0.49\textwidth]{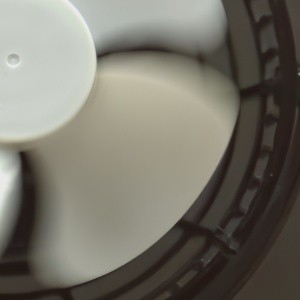}
\end{subfigure}
\begin{subfigure}[t]{0.24\textwidth}
\centering
    \includegraphics[width=0.49\textwidth]{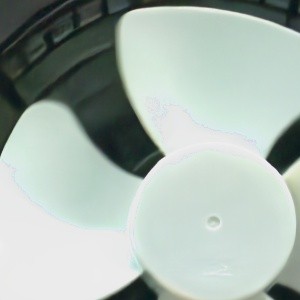}
    \includegraphics[width=0.49\textwidth]{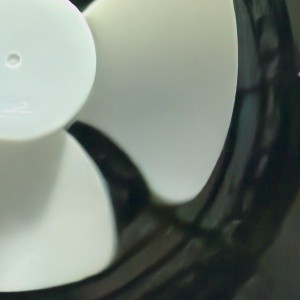}
\end{subfigure}
\begin{subfigure}[t]{0.24\textwidth}
\centering
    \includegraphics[width=0.49\textwidth]{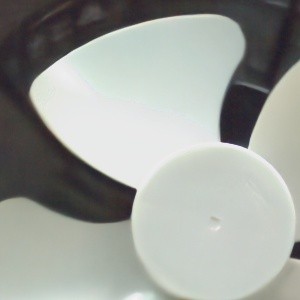}
    \includegraphics[width=0.49\textwidth]{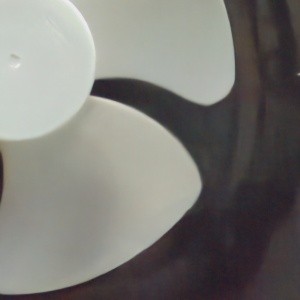}
\end{subfigure}

\begin{subfigure}[t]{0.24\textwidth}
\centering
$1/_{500} (s)$
\end{subfigure}
\begin{subfigure}[t]{0.24\textwidth}
\centering
$1/_{1k} (s)$
\end{subfigure}
\begin{subfigure}[t]{0.24\textwidth}
\centering
$1/_{5k} (s)$
\end{subfigure}
\begin{subfigure}[t]{0.24\textwidth}
\centering
$1/_{10k} (s)$
\end{subfigure}

\begin{subfigure}[t]{0.24\textwidth}
\centering
\caption{Ours}
\end{subfigure}
\caption{Dynamic videos.}
\label{fig_video}
\end{figure*}

\end{document}